\documentclass[lettersize,journal,9pt]{IEEEtran} 
\usepackage{amsmath,amsfonts,siunitx}
\usepackage{times}
\usepackage{etoolbox}
\usepackage{adjustbox}
\usepackage{graphicx}
\usepackage{multirow}
\usepackage{pgfplots} 
\usepackage{amssymb}
\usepackage{mathtools}
\usepackage{physics}
\pgfplotsset{compat=newest}
\usetikzlibrary{plotmarks}
\usepackage{tikz}
\usepackage{tikzscale}
\usetikzlibrary{pgfplots.groupplots}
\usetikzlibrary{backgrounds}
\usepackage{bm}
\usepackage[nolist]{acronym}		
\usepackage{verbatim}
\usepackage{import}
\usepackage{tkz-euclide,subfigure}
\usepackage{arydshln}  
\usetikzlibrary{patterns}
\usepackage{orcidlink}
\usepackage{verbatim}
\usepackage{cite}
\hypersetup{hidelinks}
\usepackage{blindtext}
\usepackage{hyperref}

\begin{document}
\begin{acronym}[]	
\acro{vpa}[VPA]{velocity profile analysis}
\acro{ransac}[RANSAC]{random sample consensus algorithm}
\acro{ros2}[ROS2]{Robot Operating System 2}
\acro{fmcw}[FMCW]{Frequency Modulated Continuous Wave}
\acro{if}[IF]{intermediate frequency}
\acro{awgn}[AWGN]{additive white Gaussian noise}
\acro{adas}[ADAS]{Advanced Driver Assistance Systems}
\acro{ad}[AD]{Autonomous Driving}
\acro{dbscan}[DB-SCAN]{Density-Based Spatial Clustering of Applications with Noise}
\acro{FoV}[FoV]{Field-of-View}
\acro{IPS}[IPS]{Indoor Position System}
\acro{CPI}[CPI]{coherent processing interval}
\acro{EOT}[EOT]{extended object tracking}
\acro{rsme}[RSME]{Root-Mean-Square Error}
\acro{ula}[ULA]{uniform linear antenna}
\acro{RCS}[RCS]{Radar Cross Section}
\acro{CNN}[CNN]{convolutional neural network}
\acro{GNN}[GNN]{Graph Neural Network}
\acro{dbf}[DBF]{digital beamforming}
\acro{if}[IF]{intermediate frequency}
\acro{ula}[ULA]{uniform linear array}
\acro{fft}[FFT]{Fast Fourier Transformation}
\acro{cfar}[CFAR]{Constant False Alarm Rate}
\acro{snr}[SNR]{Signal-to-Noise Ratio}
\acro{rmse}[RMSE]{Root Mean Square Error}
\acro{std}[STD]{Standard Deviation}
\acro{lstm}[LSTM]{long short-term memory}
\acro{ol}[OL]{overlap}
\acro{IMM-EOT}[IMM-EOT]{Interacting Multiple Model with Extended Object Tracking}
\acro{RMM-EOT}[RMM-EOT]{Random Matrix Method with Extended Object Tracking}
\acro{HTG}[HTG]{Hierarchical Truncated Gaussian}
\acro{em}[ST]{Standard Tracking}
\acro{immeot}[IMM-EOT]{interacting multiple model with extended object tracking}
\acro{fn}[FN]{False-Negatives}
\acro{LSTM} [LSTM] {Long Short-Term Memory Networks}
\acro{BEV} [BEV] {Bird's-Eye-View}
\acro{MLP} [MLP] {Multilayer Perceptron}
\acro{CS} [CS] {Chirp-Sequence}
\acro{cso} [cso] {coordinate system origin}
\acro{SE} [SE] {Structuring Element}
\acro{PoI} [PoI] {pixel of interest}
\acro{E-RPN}[E-RPN]{Euler-Region-Proposal}
\acro{CSP-Darknet53}[CSP-Darknet53]{Cross-Stage Partial Networks with Darknet53}
\acro{PaNet}[PANet]{Path Aggregation Network}
\acro{YOLO}[YOLO]{You Only Look Once}
\acro{SPP}[SPP]{Spatial Pyramid Pooling}
\acro{md}[\textmu{D}]{micro-Doppler}
\acro{RMM}[RMM]{Random Matrix Method}
\acro{IMM}[IMM]{Interacting Multiple Model}
\acro{CoG}[CoG]{Center of Gravity}
\acro{CuT}[CuT]{Cell under Test}
\acro{roi}[RoI]{Region of Interest}
\acro{adam}[ADAM]{Adaptive Moment Estimation}
\acro{ai}[AI]{artificial intelligence}
\acro{sprb}[SPRB]{set of points on a rigid body}
\acro{WHO}[WHO]{World Health Organization}
\end{acronym}
\title{Anchor-Based AI Approach for Pre-Crash Object Detection Utilizing Micro-Doppler Signatures in Automotive Radar}

\author{\vspace{-1mm}Patrick Zaumseil\raisebox{1mm}{\orcidlink{0009-0000-9981-8301}}, Rainer Engert\raisebox{1mm}{\orcidlink{0009-0008-6058-7540}}, Dagmar Steinhauser\raisebox{1mm}{\orcidlink{0000-0003-2735-2703}}, Jonathan Wache\raisebox{1mm}{\orcidlink{0009-0000-2552-8731}}, Soumya Dewangan\raisebox{1mm}{\orcidlink{0009-0007-8553-785X}}, and \\ Thomas Brandmeier\raisebox{1mm}{\orcidlink{0009-0006-1294-3905}}
\thanks{This work has been submitted to the IEEE for possible publication. Copyright may be transferred without notice, after which this version may no longer be accessible. This work was supported by the Bavarian State Ministry for Economics, Regional Development, 
and Energy (StMWi) within the funding program BayVFP Förderlinie Digitalisierung under 
contract number DIK0187/06. The first two authors contributed equally to this work. Corresponding authors: Patrick Zaumseil and Rainer Engert.

Patrick Zaumseil, Rainer Engert, Dagmar Steinhauser and Thomas Brandmeier are with the
Center of Automotive Research on Integrated Safety Systems and Measurement
Area (CARISSMA), Technische Hochschule Ingolstadt, 85049
Ingolstadt, Germany (e-mail: patrick.zaumseil@carissma.eu).

Jonathan Wache is with the Continental ADC Automotive Distance Control Systems GmbH, 88131 Lindau, Germany (e-mail: jonathan.wache@continental-corporation.com)

Soumya Dewangan is with Continental Autonomous Mobility India Private Limited, 560099 Bangalore, India (e-mail: soumya.dewangan@continental-corporation.com)
}}
\maketitle

\begin{abstract}
Advanced automated driving presents significant potential to improve modern automotive safety systems, but it depends highly on the reliable activation of restraint systems. Innovative solutions and the integration of forward-looking sensors are crucial for immediate and precise object detection. Recent developments in automotive radar technology enable detailed environment detection and the recognition of high-resolution features, such as micro-Doppler signatures. Combined with advanced AI techniques, these features significantly enhance object detection and improve the accuracy of kinematic parameter estimation. This is essential for the early and reliable activation of irreversible safety systems, such as smart airbags and adaptive seat belts. Therefore, an anchor-based AI model is presented, designed to process high-resolution radar data with an explicit focus on micro-Doppler signatures to improve pre-crash object detection. Furthermore, these signatures can improve the accuracy of kinematic object parameter estimation and reduce false negatives, especially in the critical near-field. To address the challenges of sparse and fluctuating radar point clouds, an innovative radar-image dilation technique on the feature input channels was developed to amplify local radar patterns, like micro-Doppler features. Therefore, this approach increases the system's reliability and increases its ability to detect objects in pre-crash scenarios despite radar multipath reflections and ghost objects.
In order to investigate the applicability and compare the model's performance with advanced automotive radar tracking methods, a radar data set using series sensors and pre-crash relevant scenarios was recorded. The results demonstrate the advantages of the anchor-based AI model over established tracking approaches. It excels at estimating object parameters in dynamic scenarios and underscores its ability to process different data sets effectively.
\end{abstract}

\begin{IEEEkeywords}
vehicle safety, radar feature analysis, micro-Doppler signature, AI vehicle detection, pre-crash detection
\end{IEEEkeywords}

\section{Introduction}
\label{Intro}
\IEEEpeerreviewmaketitle

\IEEEPARstart{A}{nnually}, about 1.2 million lives are lost in traffic-related accidents, making road traffic injuries a significant cause of death. The \ac{WHO} has established an ambitious goal for road safety and sustainable mobility, aiming to reduce road traffic injuries by \SI{50}{\percent} by 2030 \cite{Reference:WHO2023}. Furthermore, the primary goal of the Vision Zero safety strategy is to completely prevent fatalities and serious injuries from traffic accidents \cite{Reference:EuroNCAP}. Despite the long-standing and persistent prevalence of these injuries, most road accidents can be predicted and prevented using advanced vehicle sensor technology integrated into modern pre-crash safety systems. To greatly reduce the number of people injured and killed on the road, current series sensor technologies need to be further developed, and state-of-the-art real-time algorithms and \ac{ai} approaches optimized for more accurate object parameter estimations.

\vspace{-2mm}
\subsection{Motivation}
Innovative trends in automated driving present promising opportunities and significant challenges for predictive vehicle safety systems. Although it opens doors to improved safety features, it also introduces new risks, such as changes in driving behavior and complex scenarios in mixed traffic, alongside the persisting human error factor. To address these challenges and use the opportunities presented by automated driving, it is essential to meet existing vehicle safety standards. In addition, \ac{adas} must be systematically improved with the integration of innovative research advances to achieve Vision Zero \cite{Innenraumkonzepte-La}. Innovative vehicle safety systems are essential for new driverless interior concepts. For example, rotated seat positions can be adjusted just before a collision occurs, and adaptive smart airbags and seat belts automatically adjust to these changed positions, thus maximizing occupant protection. To enable these proactive safety systems, advanced, forward-looking sensors are crucial for accurately detecting potential crash objects in the near-field, predicting the trajectories, and evaluating crash severity. These factors are critical for the early and effective activation of vehicle safety systems, such as seat adjustment and restraint systems \cite{Innenraumkonzepte-La,Innenraum3}.

One critical weakness of current vehicle safety systems is their reactive nature, which means that they detect collisions only during an accident. This limits their ability to react quickly, which is essential for the implementation of advanced vehicle safety concepts. The timing demands associated with activating advanced safety systems underscore the need for forward-looking sensor technology coupled with early and reliable algorithmic collision detection to ensure timely intervention to mitigate the severity of the crash. Advanced \ac{ai} concepts introduce novel approaches to real-time object detection within a single sensor cycle, providing vital object information. This is important, especially in high-dynamic driving scenarios, particularly in moments leading to potential collisions \cite{Insassenbewegung-Ol, Ganzheitliche-Ko}.

Radar sensors are crucial for ensuring the safety of autonomous driving alongside optical sensors such as LiDAR and camera. Recent advances in radar technologies enable the detection of distinctive high-resolution radar signatures of objects in close range \cite{Engels_Advances_in_Automotive_Radar_A_framework_on_computationally_efficient_high_resolution_frequency_estimation}. This predictive sensor technology has become essential for its robust performance in various weather conditions and its ability to measure vehicle Doppler velocities. Furthermore, vehicle movement with rotating wheels induces specific frequency modulations in low-level radar data, referred to as the \ac{md} effect, further enhancing the radar capabilities \cite{microdoppler-Ch,Kamann_Extended_Object_Tracking_Using_Spatially_Resolved_Micro-Doppler_Signatures}.

Addressing the challenges of automated driving and improving vehicle safety requires researching the potential of \ac{md} features to enhance the model performance of \ac{ai}-driven techniques. This paper examines whether unique \ac{md} signatures can effectively be used to improve \ac{ai}-based object detection in vehicle safety applications.

\subsection{Related Works}
\label{Related Works}
Ongoing advancements in radar technology enable the detection of multiple reflective points within an enlarged object. This breakthrough has expanded the capabilities of object detection and tracking, making traditional assumptions about point targets obsolete. Addressing the challenges posed by multiple reflection points and extended objects has aggravated the difficulty of detection and tracking tasks.
Consequently, improved tracking techniques and new \ac{ai} methods have been developed to deal with this increased complexity \cite{Held_A_Novel_Approach_for_Model_Based_Pedestrian_Tracking_Using_Automotive_Radar}.

\subsubsection{Public Database}
\label{subsubsec:public_db}
This work focuses on an \ac{ai}-based method for processing and analyzing high-resolution radar data, specifically for radar point clouds in pre-crash scenarios. The public database KITTI is not suitable because it only includes camera and LiDAR data \cite{geiger_kitti}. In contrast, the NuScenes database includes radar point clouds and camera as well as LiDAR data \cite{Caesar_nuScenes}. The study in \cite{Reference:Nobis_Kernel_Point_Convolution_LSTM_Networks_for_Radar_Point_Cloud_Segmentation} by F. Nobis et al. shows that the density of the radar point clouds influences the feature extraction of point-based segmentation models on different subsets of NuScenes. This effect can be attributed to the low resolution of the used series radar, which leads to a small radar point cloud and a limited number of radar features per frame \cite{Reference:Scheiner_Object_detection_for_automotive_radar_point_clouds_a_comparison}.
Public radar databases often have a so-called blind spot in the near-field. This is caused by the positioning of the LiDAR sensor, which serves as a ground truth for the radar frames on the roof of the ego vehicle. Due to this arrangement, the lack of labeled radar frames poses a significant problem \cite{early_fusion}, especially since these distances are crucial to achieving the objectives of this work. While public databases cover many scenarios, they typically do not contain relevant pre-crash scenarios. Furthermore, the low resolution of series radar limits the availability of \ac{md} signatures of rotating vehicle wheels, which are essential for this research.

\subsubsection{Object Detection using Deep Learning Networks}
\label{subsubsec:ai_object_det}
Integrating automotive radar data into deep learning algorithms is in its early stages and requires several signal processing steps to make it possible. Numerous methods in this domain rely on radar detections that comprise individual range, azimuth, Doppler, and \ac{RCS} values caused by objects\cite{Reference:Patel_Deep_Learning_based_Object_Classification_on_Automotive_Radar_Spectra}.

Various publications have explored object classification and bounding box regression in automotive radar data. Traditionally, these tasks are treated separately, as shown in studies such as Lombacher et al.'s work on semantic radar grids \cite{Reference:Lombacher_Semantic_radar_grids} and Roos et al.'s investigation into the orientation estimation of vehicles through bounding box regression \cite{Reference:Roos_Estimation_of_the_orientation_of_vehicles_in_high_resolution_radar_images}.
For object classification, Schumann et al. compare the performances of \ac{lstm} networks utilizing radar data \cite{Reference:Schumann_Comparison_of_random_forest_and_long_short_term_memory_network_performances_in_classification_tasks_using_radar}. Ulrich et al. introduce DeepReflecs, a deep learning approach for classification of automotive objects with radar reflections \cite{Reference:Ulrich_DeepReflecs_Deep_Learning_for_Automotive_Object_Classification_with_Radar_Reflections,Reference:Schumann_Supervised_Clustering_for_Radar_Applications_On_the_Way_to_Radar_Instance_Segmentation}. 
Additionally, Schumann et al. delve into supervised clustering techniques for radar applications, paving the way for radar instance segmentation \cite{Reference:Schumann_Comparison_of_random_forest_and_long_short_term_memory_network_performances_in_classification_tasks_using_radar}. Scheiner et al. address the crucial question of resolution in automotive radar classification by comparing standard sensors with experimental radar setups \cite{Reference:Scheiner_Off-the-shelf_sensor_vs_experimental_radar_How_much_resolution_is_necessary_in_automotive_radar_classification?}.

The approach outlined in \cite{Reference:Danzer_2D_Car_Detection_in_Radar_Data_with_PointNets} represents a significant step in integrating multiple tasks, allowing object detection exclusively through automotive radar detections. These enhancements can be classified according to the input data provided to the \ac{CNN}. Drawing inspiration from Frustum PointNets \cite{Reference:Qi_Frustum_PointNets_for_3D_Object_Detection_from_RGB-D_Data}, this innovative method employs a two-stage network architecture, which yields promising results for radar-based object detection.
An alternative approach is spectrum-based models that use radar data incorporating range, Doppler velocity, and azimuth angle. The input data can be formatted in a three-dimensional structure \cite{Reference:Palffy_CNN_Based_Road_User_Detection_Using_the_3D_Radar_Cube} or in a reduced two-dimensional structure \cite{Reference:Meyer_Graph_Convolutional_Networks_for_3D_Object_Detection_on_Radar_Data,Reference:Wang_RODNet_Radar_Object_Detection_using_Cross_Modal_Supervision, Reference:Brodeski_Deep_Radar_Detector}.

Grid-based architectures involve discretizing the radar point cloud consisting of x, y, radial velocity, and \ac{RCS} onto a 2D \ac{BEV} or even a 3D voxel grid. This process can be achieved through manually designed feature extraction methods, as highlighted in works such as \cite{Reference:AI_Object_Detection,Reference:Meyer_Deep_Learning_Based_3D_Object_Detection_for_Automotive_Radar_and_Camera, Reference:Scheiner_Object_detection_for_automotive_radar_point_clouds_a_comparison,Reference:Lee_Deep_Learning_on_Radar_Centric_3D_Object_Detection,Referemce:Nobis_Radar_Voxel_Fusion_for_3D_Object_Detection}, or through the application of learned feature encoders using \ac{CNN}, as demonstrated in studies such as \cite{Reference:Scheiner_Object_detection_for_automotive_radar_point_clouds_a_comparison,Reference:Xu_RPFA-Net_a_4D_RaDAR_Pillar_Feature_Attention_Network_for_3D_Object_Detection,Reference:Lang_PointPillars_Fast_Encoders_for_Object_Detection_From_Point_Clouds}.

In the work by Niederlöhner et al. \cite{Reference:Niederloehner_Self-Supervised_Velocity_Estimation_for_Automotive_Radar_Object_Detection_Networks}, a novel approach was introduced to learn the cartesian velocity of objects using a grid-based object detection network applied to automotive radar data. A key aspect of their method is its self-supervised nature, where the model generates its training signal for velocity estimation. The need for labels is limited to single-frame-oriented bounding boxes (OBBs).
A related study by Köhler et al. \cite{Reference:Koehler_Improved_Multi_Scale_Grid_Rendering_of_Point_Clouds_for_Radar_Object_Detection_Networks} proposed an innovative technique for grid rendering. This method utilizes kernel point convolutions to enhance the encoding of local point cloud contexts. In addition, the authors present a comprehensive multiscale grid rendering approach designed to integrate multiscale feature maps into the convolutional frameworks of detection networks.

Recent advances in \ac{ai} for automotive applications hold great promise in enhancing state-of-the-art vehicle safety systems. Despite this progress, there are challenges to ensure robustness, achieve highly accurate object detection, and improve processing speeds to meet the reliability requirements of these critical safety systems.

\subsubsection{Data Perturbations in Camera and Radar Images}
\label{subsubsec:ai_robust}
Prior research has examined the vulnerability of deep learning models to deceptive input, which can cause incorrect classification and reduce robustness \cite{Reference:Ian_Explaining_and_Harnessing_Adversarial_Examples, Reference:Szegedy_Intriguing_properties_of_neural_networks}. This approach modifies the input data by introducing worst-case perturbations to the test frames of the data set. Hendrycks et al. \cite{Reference:Hendrycks_Benchmarking_Neural_Network_Robustness_to_Common_Corruptions_and_Perturbations} have introduced a benchmark specifically designed to assess the performance of neural networks in the face of typical corruptions and perturbations, rather than in worst-case scenarios. Their findings highlight the effectiveness of augmenting the training dataset with noisy perturbations, demonstrating increased resilience to common perturbations.
Furthermore, the work presented in \cite{Reference:Shuangzhi_Common_Corruption_Robustness_of_Point_Cloud_Detectors_Benchmark_and_Enhancement} introduces a physics-based simulation approach to generate degraded point clouds under common perturbations. 

The reliability of radar point cloud data is crucial in safety-critical applications. However, adverse weather, multipath reflections, and ghost targets can significantly affect data quality and degrade \ac{snr} \cite{Reference:Infuence_Rain, Reference:Multipath}. Detecting vehicles with high-resolution radar sensors involves modeling them as extended objects characterized by geometric extent, \ac{RCS}, and velocity \cite{Held_A_Novel_Approach_for_Model_Based_Pedestrian_Tracking_Using_Automotive_Radar}. \ac{RCS} fluctuations, caused by the phase sensitivity of radar signals, lead to variations in reflected signal strength, posing challenges for reliable object detection. In contrast, the Doppler and \ac{md} from moving components, such as rotating wheels, provide dynamic information, allowing for a more accurate object parameter estimation through detailed motion features \cite{Kamann_Extended_Object_Tracking_Using_Spatially_Resolved_Micro-Doppler_Signatures}. However, due to the nature of radial measurements, both the Doppler and \ac{md} amplitudes depend on orientation and decrease significantly in entirely lateral situations \cite{Held_A_Novel_Approach_for_Model_Based_Pedestrian_Tracking_Using_Automotive_Radar}. These features are crucial to improve object detection, highlighting the need for \ac{CNN}s to utilize them to handle data corruption effectively \cite{Reference:Increasing_the_Robustness_of_Semantic_Segmentation_Models_with_Painting_by_Numbers, Reference:Chamseddine_Ghost_Target_Detection_in_3D_Radar_Data_using_Point_Cloud_based_Deep_Neural_Network}.

\subsubsection{Extended Object Tracking}
A recent comprehensive review of state-of-the-art techniques for tracking extended objects classified various measurement probability methods into several groups \cite{EOT_Granststroem_17}. The first group uses a \ac{sprb}, presuming a fixed count of reflection points on a rigid body shape \cite{Reference:Buhren_SimulationofAutomotiveRadar,Reference:Hammerstrand_AdaptiveRadarSensorModel,Reference:Hammerstrand_Extended_Object_Tracking_using_a_Radar_Resolution_Model,Reference:Gunnarsson_Tracking_vehicles_using_radar_detections}.
An alternative technique for monitoring extended objects involves employing spatial distribution modeling. Initially introduced in \cite{Reference:Gilholm_Poisson_models_for_extended_target_and_group_tracking}, this approach works assuming that the frequency of detections is in accordance with a Poisson distribution. A more precise estimation of the object's shape and dimensions can be attained by meticulously modeling the spatial arrangement of target detections surrounding the object.
In \cite{HTG-XIA}, an innovative surface-volume model closely mirrors the spatial distribution characteristics observed in vehicle radar measurements. At its core is the incorporation of a hierarchical truncated Gaussian measurement model. This model effectively utilizes unobserved measurement sources to capture a distinctive feature frequently observed in radar data, specifically the tendency of radar detection points to cluster around the edges of objects within a designated volume.
Another approach of extended object tracking focuses on modeling the measurement probability by incorporating the intrinsic physical characteristics of the tracked objects. In \cite{Reference:Knill_A_direct_scattering_model_for_tracking_vehicles_with_high-resolution_radars}, a direct distribution extended object model is presented, including range and Doppler shift, through multiple distributions.

As highlighted in the review papers \cite{EOT_Granststroem_17,HTG-XIA}, these diverse methods offer valuable techniques to handle the complexities of advanced extended object tracking methods, presenting practical applications explicitly tailored for automotive radar data.

\vspace{-1mm}
\subsection{Research Contribution}
Section \ref{Related Works} provides an overview of the current state of radar-based object detection in the automotive field, highlighting that \ac{ai}-based approaches which are still relatively new to radar object detection, represent a rapidly growing field of research. This paper investigates the potential of high-resolution radar data, particularly the distinctive \ac{md} components of vehicles, to improve the accuracy of \ac{ai}-based methods for object parameter estimation and object detection within a single measurement cycle.

Publicly available databases such as NuScenes (referred to in section \ref{subsubsec:public_db}) are not suitable for this purpose, as they are neither designed for pre-crash scenarios nor include high-resolution radar features, particularly \ac{md} information from approaching vehicles. Furthermore, the radar frames in these databases are characterized by sparse radar point clouds and include blind spots, resulting in a shortage of radar frames relevant to pre-crash analysis. Therefore, a specialized pre-crash radar database was generated, featuring an increased point density, extended detection characteristics, and a detailed representation of \ac{md} components. This database serves as the foundation for a systematic investigation of the impact of these components on the accuracy and precision of object parameter estimation in pre-crash scenarios.

An anchor-based deep learning approach is employed due to its efficient in processing data and suitable for quick analysis of 3D point clouds \cite{Reference:Simon_Complex-YOLO_Real_time_3D_Object_Detection_on_Point_Clouds}. Initially designed for LiDAR data processing, the \ac{ai} architecture's input channels were specifically adapted to handle automotive radar data. Notably, Doppler information was integrated into the architecture to capture the vehicle's body velocity and the unique micro-Doppler signatures from rotating wheels, an area that has seen limited research. In addition, a dilation method was implemented directly on the input channels to improve radar data object detection and amplify the \ac{md} features. This approach mitigates interference from radar multipath reflections and the influence of superimposed ghost objects in the near-field. As a result, it significantly reduces false negatives, particularly in close-range scenarios, ensuring reliable and precise performance. A comparison of the implemented AI method with advanced radar-based tracking techniques using series radar data demonstrates promising results. The AI method achieves comparable accuracy in various scenarios and frequently outperforms established tracking methods. This validates the applicability of the developed approach to current sensor systems and confirms its practical feasibility. In contrast to traditional methods typically rely on sequential measurements and require a settling phase, the \ac{ai}-based solution enables accurate estimation of object parameters from a single frame, which is crucial for objects that appear suddenly.

To the authors' best knowledge, the work represents the first integration and investigation of high-resolution \ac{md} characteristics and their significance in pre-crash object detection combined with innovative radar-image processing. A comparison between state-of-the-art tracking methods and the \ac{ai} approach shows the possibility of using AI algorithms for pre-crash object detection to increase vehicle safety.

\subsection{Outline}
Section II delves into the radar signal processing principle, detailing methods for extracting \ac{md} components from high-resolution radar databases and describing an advanced tracking method.
Section III focuses on the \ac{ai} system architecture, containing radar data pre-processing and feature projection onto 2D \ac{BEV} input images. This section also covers techniques such as morphological radar-image dilation and the concept of a system model.
Section IV provides an overview of the research and series radar databases and a comparative analysis.
Section V presents results, examines critical pre-crash scenarios, and showcases the results of the radar-image dilation method. In addition, it compares the performance of an anchor-based \ac{ai} method against classical series object tracking algorithms utilizing series radar data.
Section VI contends with a discussion of potential challenges and offers corresponding solutions.
Lastly, Section VII summarizes the results obtained and outlines future research directions.
\section{Methods}
\label{sec:methods}
\subsection{Signal Processing}
Modern automotive radar sensors utilize a technique that involves the transmission of a sequence of rapidly modulated chirps \cite{Reference:Xu_FMCW_Chirps} to efficiently isolate the range-dependent component of the \ac{if} signal from the Doppler effect induced by relative radial velocity ($v_d$). This method renders the Doppler contribution insignificant. Instead, the $v_d$ is determined by analyzing the phase shift between successive chirps, which arises solely from the target's velocity \cite{costa1984signal}, \cite{Stove1992LinearFR}.
To determine the azimuth angles of the detected objects, the radar system uses \ac{dbf} to map the range-Doppler spectra in various directions. This mapping is achieved by evaluating the phase progression across the receiver antennas (RX) of a \ac{ula} caused by the incident wave. Typically, a \ac{fft} is employed to estimate the 3-dimensional spectral components of multiple targets within the field of view, as expressed in \cite{Engels_Advances_in_Automotive_Radar_A_framework_on_computationally_efficient_high_resolution_frequency_estimation, Diss_Heidenreich}
\begin{align}
    y(\lambda, \mu, \rho) = \sum_{l_s = 0}^{L_s - 1} \sum_{m_s = 0}^{M_s - 1} \sum_{n_s = 0}^{N_s - 1} w_\lambda(l_s) w_\mu(m_s) w_\rho(n_s) \notag \\ \cdot x(l_s, m_s, n_s) \cdot e^{-j(\lambda l_s + \mu m_s + \rho n_s)}.
    \label{eq:ffts}
\end{align}
In this context, $x(l_s, m_s, n_s)$ represents the \acf{if} signal, where the indices $l_s, m_s$ and $n_s$ are defined within specific ranges: $l_s$ varies from 0 to $L_s - 1$, $m_s$ ranges from 0 to $M_s - 1$ and $n_s$ spans from 0 to $N_s - 1$. The values $L_s$, $M_s$ and $N_s$ correspond to the quantities of range samples, chirps, and azimuth antennas, respectively. 
Additionally, $w_\lambda(l_s)$, $w_\mu(m_s)$ and $w_\rho(n_s)$ denote window functions that have been normalized for different dimensions, specifically for range, Doppler, and azimuth. The frequency parameters associated with the range ($r$), radial relative velocity ($v_d$), and azimuth angle ($\delta$) are expressed as follows:
\begin{figure}[!t]
  \centering
  \subfigure[\ac{dbscan} cluster concept]{\includegraphics[width=0.48\columnwidth]{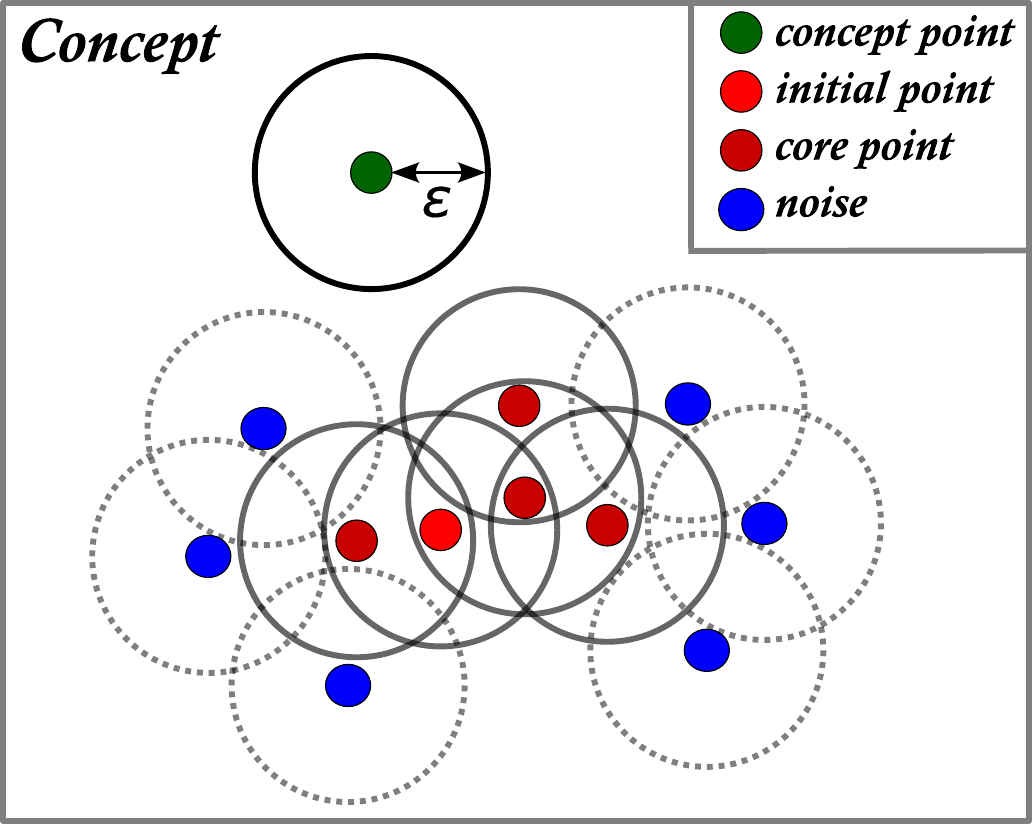}}\label{fig:we_ransac}
  \hfill
  \subfigure[Altered cluster concept]{\includegraphics[width=0.48\columnwidth]{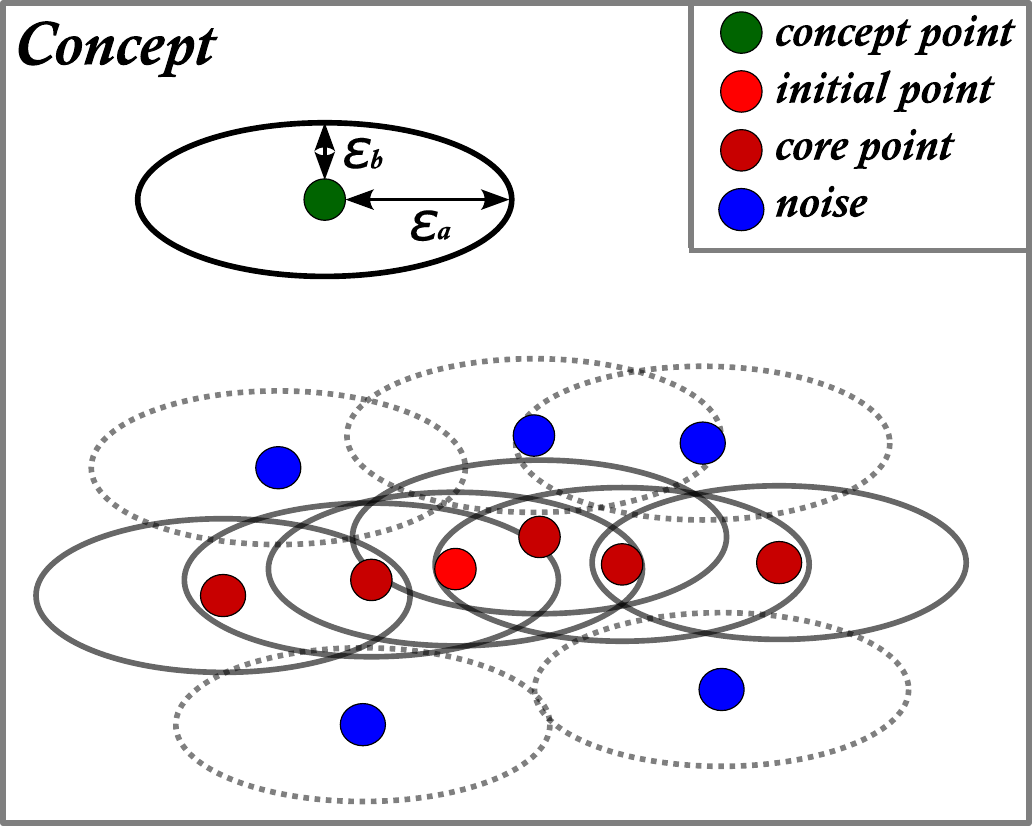}}
  \label{fig:we_md}
  \caption{Comparison of two cluster concepts with exemplary data points clustered. Results are indicated by colors: initial point (light red), core and border points (dark red), noise points (blue) and the concept points (green)}
  \vspace{-4mm}
  \label{fig: cluster}
\end{figure}
$\lambda = (2\pi/L_s) \cdot l_s$, $\mu = (2\pi/M_s) \cdot l_s,$ and $\rho = (2\pi/N_s)\cdot n_s$. This representation facilitates the translation between discrete indices and continuous frequency parameters, allowing effective signal processing in systems like radar, where multiple dimensions play a crucial role in data analysis.

Instead of employing a \ac{fft} to estimate the power spectral densities in azimuth, we opt for the Burg algorithm \cite{Reference:Burg}. It offers superior spectral resolution compared to windowed periodograms generated using a \ac{fft}, as it mitigates sidelobe phenomena, as mentioned in reference \cite{Reference:Burg2}.

\subsubsection{Constant False Alarm Rate (CFAR)}
In the context of power detection, the process involves averaging the magnitudes of complex range-Doppler spectra that are individually computed on all $N_s$ antennas. This averaging is performed through non-coherent integration, as described in \cite{signal_Richards_05}.
\vspace{-2mm}
\begin{align}
    \abs{\overline{X}(\lambda,\mu)}_{N_s} = \frac{1}{N_s} \sum^{N_s}_{n_s=1} \abs{X(\lambda,\mu)}_{n_s}
    \label{eq:power_detection}
\end{align}
The enhancement of \ac{snr} achieved through this integration significantly improves the ability to distinguish between noise and genuine target detections. Consequently, this improvement greatly improves the overall detection performance. A \ac{cfar} filter is a fundamental technique used in radar and signal processing to detect and differentiate targets from background clutter within a noisy and interference-prone environment. This process involves the identification of detections through the search for peaks in the averaged magnitude-squared range-Doppler spectrum.

\subsubsection{Density-Based Spatial Clustering of Applications with Noise (DB-SCAN)}
Upon completing the \ac{cfar} procedure on the radar data cube, the next step involves the application of a clustering technique to group the detections that emanate from a unified extended target, exemplified by a vehicle. The \acs{dbscan} is an algorithm that identifies clusters within a dataset based on the density of data points. It works by first selecting an initial data point, then examines the density of data points within a specified radius, referred to as $\epsilon$. If this radius contains at least a minimum number of data points, often referred to as MinPts, the selected point is designated as the core point. Data points that are within the $\epsilon$ radius of a core point but do not meet the criteria to be core points themselves - i.e., they have fewer than MinPts neighboring points - are labeled as density-reachable points. These points are considered part of a cluster but are not core points. Any data points that do not qualify as core points or density-reachable points are categorized as noise points and do not belong to any cluster. 
Enhancing the performance of the clustering algorithm involves a comprehensive rewrite and 
\begin{figure}[!t]
  \centering
  \subfigure[An exemplary model of the \ac{ransac} algorithm, where outliers are represented in red, inliers in blue, and the model with its associated threshold as dotted lines.]{\includegraphics[width=0.48\columnwidth]{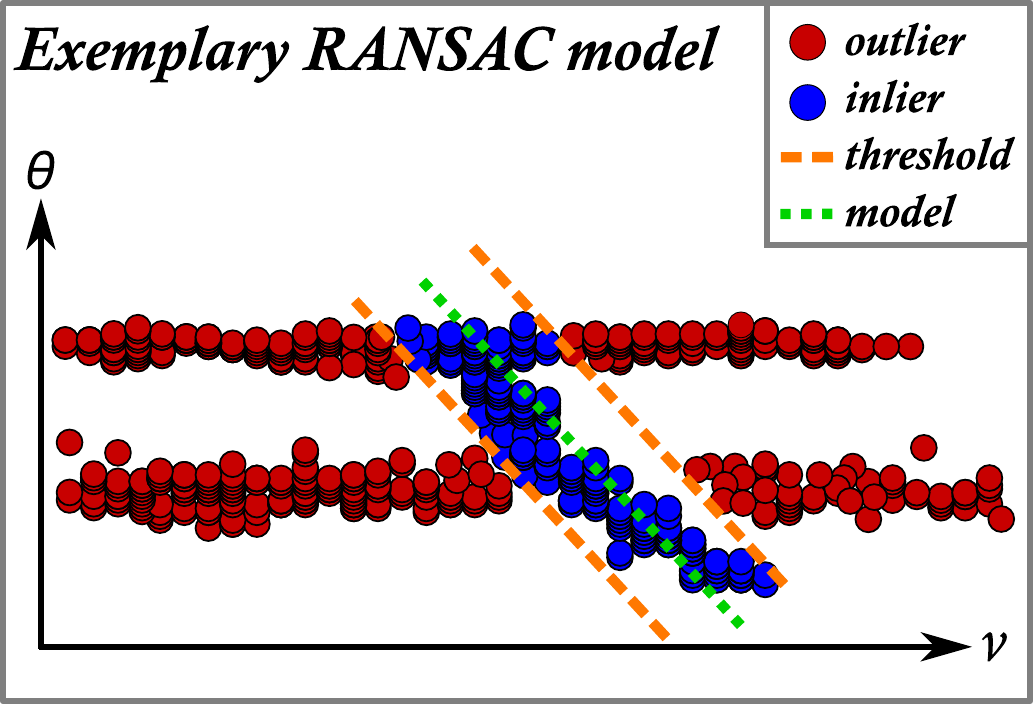}}
  \label{fig:ransac}
  \hfill
  \subfigure[Overview of the wheel velocity $v_\mathrm{total}$ (blue arrow) consists of both a rotational $\overrightarrow{v}_{\omega}$ (green arrow) and a translational velocity component $v_\mathrm{ego}$ (red arrow). \textcopyright 2023 IEEE \cite{Reference:Zaumsiel_Radar_Signature_of_a_Micro_Doppler_generating-Soft_Target_for_Automotive_Pre_Crash_Systems}]{\includegraphics[width=0.48\columnwidth]{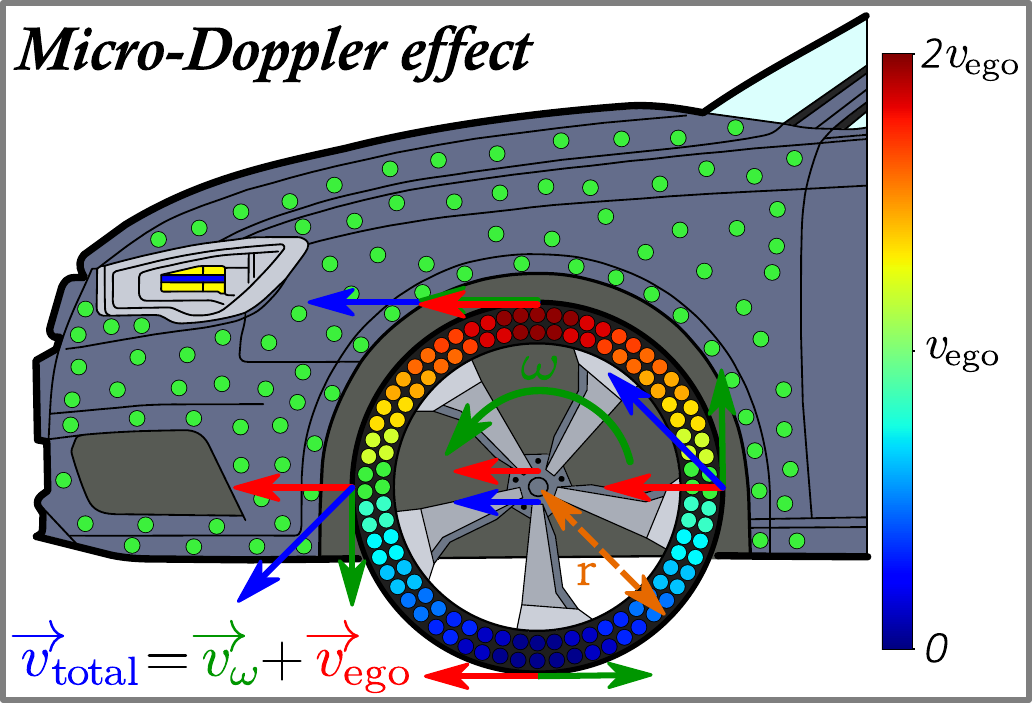}}
  \label{fig:md}
  \caption{Visualization of two key components in the wheel extraction process: the \ac{ransac} model and the micro-Doppler effect caused by a rotating wheel.}
  \vspace{-3.5mm}
  \label{fig:ransac_md}
\end{figure}
optimization, introducing several significant changes. One of the most notable alterations is to replace the single epsilon value $\epsilon$ used in the original algorithm with two distinct $\epsilon_a$ and $\epsilon_b$ for each dimension in the input data. This modification enables the algorithm to cluster data points asymmetrically, improving the clustering outcomes. A comparison of the two algorithms is depicted in Figure \ref{fig: cluster}.

\subsection{Wheel Extraction Algorithm}
\label{sec:wheel_ex}
The wheel extraction algorithm is a method used to detect vehicle wheels in radar data. The process begins by forming clusters in both the x-y and range-Doppler planes from the raw radar data to approximate the vehicle's rigid body shape. To identify detections associated with the bulk velocity of the vehicle, a \ac{ransac} technique was used. This approach utilizes a velocity profile in the velocity-azimuth plane as the underlying signal model, depicted as a dashed green line in Figure \ref{fig:ransac_md}(a) \cite{Ransac_Fischler}.

The number of iterations, $n$, is determined using Equation (\ref{eq:iterations}). This calculation considers the total number of radar detections $s$, the proportion of outliers $\epsilon$ that do not fit the expected model, and the desired probability $\rho$ of obtaining at least one outlier-free calculation.
{
\begin{align}
    n = \frac{\log (1-\rho)}{\log (1-(1-\epsilon)^s)}
    \label{eq:iterations}
\end{align}
}

In each iteration $n$, the cumulative error $C$ is determined by summing the individual error values corresponding to the number of outliers. The cumulative error is then compared to that of the previous model. If $C$ is lower, the current model is considered a better fit and replaces the previous one. If $C$ is higher, the current model is discarded. Once the optimal model has been determined, as shown in Figure \ref{fig:ransac_md}(a), the next step is to calculate the distance between the velocity signal model and each detection point. This distance provides the value $v_\theta$, which represents the velocity at a specific angle. These values are then used to exclude detection points that, due to physical constraints, do not qualify as wheel detection.
These dependencies result from combining two key velocities: the chassis velocity $v_{\text{ego}}$ and the wheel's rotational velocity $v_\omega$. The velocity $v_{\text{total}}$ varies based on the position on the wheel. At the highest point of the wheel, $v_{\text{ego}}$ and $v_\omega$ are aligned, producing a total velocity of 2 $v_{\text{ego}}$. At the contact point with the road, the two velocities oppose each other, effectively canceling out and resulting in a total velocity of 0. It is illustrated in Figure \ref{fig:ransac_md}(b).
The \ac{md} parameter is subsequently calculated using Equation (\ref{eq:we_md_para}). It is derived from the normalized squared deviation $v_{i,\text{diff}}^2$ between the detected velocity $v_{\text{i,det}}$ and the model-based velocity $v_\theta$ for each detection point $i$, as defined in Equation (\ref{eq:we_vdiff}). Finally, the individual $v_{i,\text{diff}}^2$ values are averaged to obtain the \ac{md} parameter. This parameter is a threshold for further filtering, excluding the remaining non-wheel detections.
\begin{align}
    & \text{Difference velocity:} && v_{i,\text{diff}}^2 = (\frac{v_{\text{i,det}} - v_\theta}{v_\theta})^2 \label{eq:we_vdiff}\\
    & \text{\ac{md} parameter:} && \ac{md} = \sqrt{\sum_{i=0}^{N-1}{v^2_{i,\text{diff}}}} \label{eq:we_md_para}
\end{align}
The remaining detection points correspond to all detected wheels. A custom clustering method is applied in two stages to differentiate them. First, clustering is performed in the velocity-range and angle-range planes. Then, a second RANSAC-based approach is applied in the angle-velocity plane to assign each remaining radar detection point to a specific wheel.

\vspace{-1mm}
\subsection{Chassis Velocity Algorithm}
\label{sec:chassis_vel}
The chassis velocity algorithm is used to modify \ac{md} velocities of the wheel obtained from radar data, ensuring that the velocities of all detected points align with a specific part of the vehicle’s chassis. The algorithm depends on the data of the previously described wheel extraction algorithm in section \ref{sec:wheel_ex}, specifically the details regarding inliers (representing the chassis detection points) and separated wheel detections.

In the initial stage, the entire azimuthal region of the identified object is divided into smaller segments called bins. The value assigned to each bin is determined by calculating the mean velocity of all inliers within that specific bin.

The mapping process assigns each wheel's detection points to an appropriate bin based on their azimuth angle. The predefined bin value is assigned as the updated velocity value for the corresponding detection point. This iteration continues until all velocities have been re-evaluated.

\vspace{-1mm}
\subsection{EOT with HTG integration}
\label{methodes:EOT}
As stated in \cite{Reference:EOT}, the integration of \ac{RMM} based \ac{EOT} with \ac{IMM} improved tracking accuracy for dynamically challenging scenarios such as pre-crash. The \acs{IMM-EOT} combination provides enhanced position and velocity estimates compared to \ac{IMM} and \ac{EOT} alone. The RMM-based EOT (RMM-EOT) represents the states of the target object as the kinematic state $[x,y,v_x,v_y,a_x,a_y,\omega]$ and the extent matrix $X_k \in R^{2 \times 2}$, which is a definite symmetric and positive matrix. The kinematic states of the object include position, velocity, acceleration in lateral and longitudinal directions, and yaw rate.

To further enhance the estimation of the extent and the accuracy of the kinematics, a realistic spatial distribution of the automotive radar is recreated by introducing the \ac{HTG} measurement model. \ac{HTG} measurement model presents surface-volume distribution in which detections surround the boundary of the target vehicle with some volume around its edges \cite{HTG-XIA}. The introduction of the \ac{HTG} model in the state update step of \acs{RMM-EOT} provides better extent (length, width, and orientation) estimation. Truncated bounds in \ac{HTG} model are modeled as variables, which are calculated in every timestamp; detections from previous timestamps are also considered to estimate the truncation bound of the target vehicle. The measurements from the current and last two timestamps are converted from sensor to object coordinate. The expectation maximization algorithm divides the object coordinates converted measurements into four clusters. The update of the truncation bound for each cluster is done according to \cite{HTG-XIA-WANG}, using Halley’s numerical approximation method. \ac{HTG} model is combined with \acs{RMM-EOT} by introducing pseudo-measurements, which integrates the surface-volume model with the \acs{RMM-EOT} assumption of a uniform distribution over the ellipsoidal extent of the target.

The observed and pseudo measurements are collectively called augmented measurements, and the center and measurement spread of augmented measurements are calculated as described in \cite{HTG-XIA-Journal}. The further state update steps are according to \acs{RMM-EOT} with \ac{IMM} integration. The half-length \textit{l} and half-width \textit{w} are calculated from the estimated extent matrix \cite{Ellipse-Li}. The orientation of the target vehicle is calculated from the tracked velocities.
\section{System Model}
\label{sec:SA}
The \ac{YOLO} approach by Bochkovskiy et al. \cite{Reference:Alexey_Bochkovskiy_YOLOv4_Optimal_Speed_and_Accuracy_of_Object_Detection} is the foundation for object detection in the extensive MS COCO image dataset. Inspired by Simon et al.'s work \cite{Reference:Simon_Complex-YOLO_Real_time_3D_Object_Detection_on_Point_Clouds} this paper extends the lidar-based object detection and regression model alongside the PyTorch implementation of Yolov4 heads, including the \ac{E-RPN}. This integration establishes the foundation for investigating the innovative high-resolution radar-based object detection and regression model denoted in this paper as ComplexYOLO.

\begin{figure*}[htbp]
    \centering
    \includegraphics[width=\linewidth]{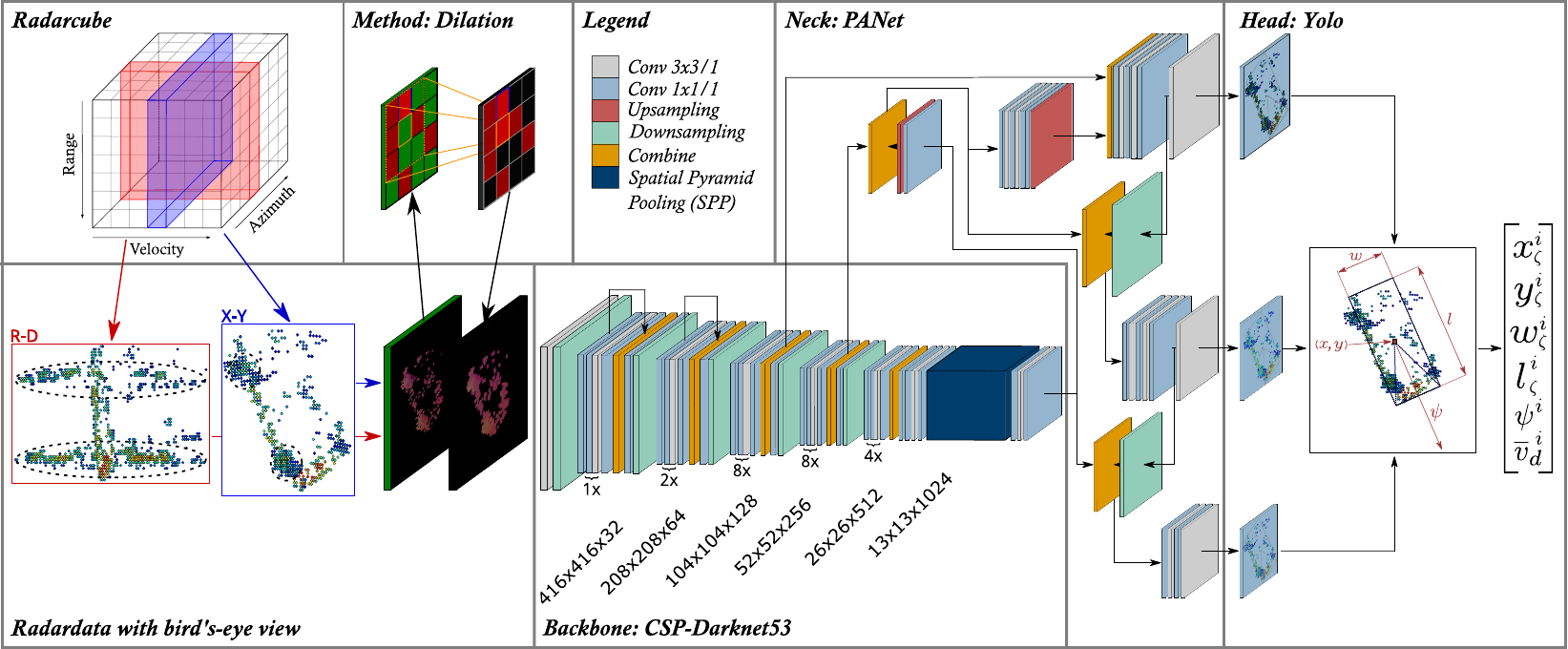}
\caption{The modified ComplexYOLO model is specifically designed for object detection and regression using automotive radar data. An exemplary high-resolution radar snapshot is visualized in Range-Doppler domain (\textcolor{red}{\rule[0.5ex]{0.8em}{.55pt}}) with velocities and extended \ac{md} velocities as well as X-Y in plane (\textcolor{blue}{\rule[0.5ex]{0.8em}{.55pt}}). An innovative radar-image dilation technique supports the projection of high-resolution radar features onto the RGB map. Following subsequent image processing steps done by the \ac{CSP-Darknet53} backbone, an \ac{SPP} technique enlarges the receptive field and extracts global features. Feature merging from different stages of the network is reached by the \ac{PaNet} in the neck. Finally, YOLO heads, operating at different image resolutions, detect the object and predict kinematic parameters with the mean Doppler velocity supplementing the output vector.}
    \label{fig:ComplexYolo_Architecture}
    \vspace{-4mm}
\end{figure*}

A modified architecture of the ComplexYOLO model designed for radar-based pre-crash object detection and regression, is visualized in Figure \ref{fig:ComplexYolo_Architecture}. This architecture was specifically developed to meet the demands of critical pre-crash scenarios using radar data with a focus on unique \ac{md} components as input. It incorporates a \ac{CSP-Darknet53} backbone to boost detection rates by integrating features from various stages, an \ac{SPP} technique for enlarging receptive fields and extracting global features, a \ac{PaNet} for merging features, and \ac{YOLO} heads for object detection and regression across different image resolutions.
Furthermore, the framework integrates an innovative radar-image dilation method to address challenges associated with sparse radar point clouds and to extract and expand local radar features, such as the \ac{md} signatures \cite{Reference:Simon_Complex-YOLO_Real_time_3D_Object_Detection_on_Point_Clouds, Reference:Alexey_Bochkovskiy_YOLOv4_Optimal_Speed_and_Accuracy_of_Object_Detection}. Furthermore, a \ac{MLP} extends the overall system to allow estimation of the target velocity.

\vspace{-1mm}
\subsection{Radar Point Cloud Pre-Processing}
\label{subsec:Point-Cloud_Pre_Processing}
\subsubsection{Radar Point Cloud Representation}
Figure \ref{fig:ComplexYolo_Architecture} depicts the general structure of the modified ComplexYOLO model. As input, a 3D radar cube is generated through signal processing, detailed in Section \ref{sec:methods}. Each radar frame at time index \(i\) is represented as $\bm{P}_{\text{in}}^i \in \mathbb{R}^{m\times n}$, where \(m\) is the number of detections and \(n\) is the number of radar features. The radar snapshot in the time frame \(i\) is displayed in the Range-Doppler domain (highlighted by the red box) and the X-Y plane (highlighted by the blue box) represented in the cube in Figure \ref{fig:ComplexYolo_Architecture}. The different velocities discussed in Section \ref{sec:wheel_ex} are observable in the Range-Doppler plot, with \ac{md} velocities highlighted by dashed ellipses in Figure \ref{fig:ComplexYolo_Architecture}. Each radar frame encapsulates distinct detections characterized by unique radar attributes, expressed as 
\begin{equation}
\bm{P}_{\text{in}}^{i}=
  \begin{pmatrix}
    \bm{r}^{i}, \bm{\varphi}^{i}, \bm{v_{\text{d}}}^{i}, \bm{\sigma}^{i}
  \end{pmatrix}^\mathbf{T}
\end{equation}
with range \(\mathbf{r}^i = [r_{1}^i, ..., r_{k}^i]\) spanning from \( r_{k}^i \in [r_\text{min},..., r_\text{max}]\), azimuth angles \(\bm{\varphi}^i = [\varphi_{1}^i, ..., \varphi_{k}^i]\) varying between \(\varphi_{k}^i \in [\varphi_{min},\varphi_{max}]\), doppler velocity \(\mathbf{v_{\text{d}}}^{i} = [v_{\text{d},1}^i, ..., v_{\text{d},k}^i] \) ranging from \(v_{\text{d},k}^i \in [v_{\text{d,min}},...,v_{\text{d,max}}]\) and \ac{RCS} values \(\bm{\sigma}^{i}=[\sigma_{1}^i, ..., \sigma_{k}^i]\) with values between \(\sigma_{k}^i \in [\sigma_{\text{min}},...\sigma_{\text{max}}]\) based on the shape and reflectivity of the object. The index $k$, ranging from 1 to $n$, corresponds to each detection within the input radar point cloud $\mathbf{P}_{\text{in}}^{i}$. Following a coordinate transformation from polar to image coordinates, and relocating the coordinate origin to the center of the \ac{BEV} image ($\mathbf{\hat{x}}^{i}$, $\mathbf{\hat{y}}^{i}$), and computing the Euclidean Distance $\mathbf{d}^{i}$ from the origin to each detection $k$, the new point cloud vector is defined as
\begin{equation}
\bm{\hat{P}}_{\text{in}}^{i}=
  \begin{pmatrix}
    \bm{\hat{x}}^{i}, \bm{\hat{y}}^{i}, \bm{d}^{i}, \bm{v_{\text{d}}}^{i}, \bm{\sigma}^{i}
  \end{pmatrix}^\mathbf{T}.
  \label{eq:description_pc}
\end{equation}
Building upon the methodologies introduced by Simon et al. \cite{Reference:Simon_Complex-YOLO_Real_time_3D_Object_Detection_on_Point_Clouds} and Chen et al. \cite{Reference:Chen_Multi_View_3D_Object_Detection_Network_for_Autonomous_Driving}, handcrafted features were derived from a 3D point cloud and subsequently integrated into 2D input maps. Following this approach, significant radar features ($\bm{d}^{i}$, $\bm{v_{\text{d}}}^{i}$, $\bm{\sigma}^{i}$) were extracted for each radar frame at each time index \(i\) and mapped on the feature input channels.

\subsubsection{Radar Point Cloud Projection}
High-resolution radar features (${d}^{i},{v_d}{^i},{\sigma}^{i}$)  are projected onto the input channels ($C_R$, $C_G$, $C_B$) in \ac{BEV} format, covering a spatial range of $[r_\text{min},..., r_\text{max}]$ in both the x and y directions. Each channel has a defined width \textit{W} and height \textit{H} denoted as \(C_{W \times H}^{i}\), with \textit{W}, \textit{H} $\in \mathbb{N}$. After adjusting the coordinate systems, the mathematical origin lies in the center of the input channels.
The normalized radar point cloud \(\bm{\hat{P}_\text{in}}^{i}\) maintains the same coordinates, with \(\bm{\hat{x}}^{i} \in [1,..., W]\) and \(\bm{\hat{y}}^{i} \in [1,..., H]\). These points are assigned to the corresponding pixels in the image plane using \(\bm{m_{j \times l}}\), where \(j\) ranges from 1 to \(W\) and \(l\) ranges from 1 to \(H\). The mapping process is described as
\begin{equation}
m_{j \times l} = 
    \begin{cases}
        \bm{\hat{P}}_{\text{in}}^{i} (\bm{d}^{i},\bm{v_d}{^i},\bm{\sigma}^{i})\  \text{if } m_{j \times l} = \emptyset \land \ [(j < x \leq j + 1), \\ \quad \quad \quad \quad \quad \quad \quad \quad \quad \quad \quad \quad \quad \  (l < y \leq l + 1)] \\
            m_{j \times l}  \ \ \quad \quad \quad \quad \text{else}.
    \end{cases}
    \label{eq:2D_Discretization_radar_point_cloud}
\end{equation}
Due to the inherent resolution of the radar system, situations occur where detections are assigned to the exact radar grid cell, leading to identical input image coordinates. When this happens, the radar features from the initial detection are projected onto all input channels, while subsequent detection points with the exact image coordinates are ignored. Following this process, the input channels are merged to create a normalized RGB image, illustrated in Fig. \ref{fig:ComplexYolo_Architecture}, which serves as the input for ComplexYOLO.

\subsubsection{Radar-Image Dilation on Short-Range Radar Data}
\label{subsec:morphological_operations}
Despite high-resolution radar point clouds, the data is limited compared to conventional camera images, as numerous pixel areas in the RGB image remain unoccupied \cite{early_fusion}. Furthermore, \ac{ai}-based object detection methods face a substantial challenge due to frame-wise fluctuations in radar detections, compounded by interference of radar multipath reflections and ghost objects \cite{Reference:Multipath, Held_A_Novel_Approach_for_Model_Based_Pedestrian_Tracking_Using_Automotive_Radar}. These factors can lead to unfavorable pixel-bin occupancy in the RGB image, diminishing the model's performance.
Especially in near-field scenarios, achieving a high level of robustness in object detection is crucial. Various methods, such as morphological techniques, have been analyzed in the literature to improve object detection accuracy. An innovative approach builds upon morphological dilation $\delta$, but instead of operating directly on images, it focuses on individual feature input channels. This unique method allows for the augmentation of the vehicle's pixel contour, enhancing the quality of radar features, and emphasizing distinct radar signatures. By utilizing a specified \ac{SE} with dimensions represented by \ensuremath{\alpha\mspace{-6mu}\times\mspace{-6mu}\alpha}, $\Lambda = \mathrm{floor}\bigr(\frac{\alpha}{2}\bigl)$ and the relative position of the \ac{PoI} $x$, $y$ as the center, specific subregions of the image can be targeted for improved extraction of local radar patterns \cite{Computer-Ha, Morphological-So}.
The final dilated image $C^{*}_\gamma (x, y)$ is represented by
\begin{align}
    C^{*}_\gamma (x, y) = \delta_\Lambda(C_\gamma, x, y) = \max_{a, b} |C_\gamma (x + a, y + b)|,  \\
    \text{ for integer } a, b = \{-\Lambda,...,\Lambda\} \notag.
\end{align}
The subindex $\gamma$ refers to the red, green, and blue input channels and is expressed as $\gamma \in [R, G, B]$. The maximum absolute value, including the negative values as well, is extracted in the defined neighborhood around the corresponding \ac{PoI} is extracted and assigned to the output image \cite{Computer-Ha, Morphological-So}. This process is repeated for all the images of the individual input channels. Depending on the shape of the \ac{SE}, the target's contour, including distinctive radar patterns, especially \ac{md} signatures, can be accentuated, as exemplified in the RGB output image in Fig. \ref{fig:ComplexYolo_Architecture}.

\vspace{-1.5mm}
\subsection{Kinematic Object State Estimation}
\label{subsec:Object_Parameter_Estimation}
ComplexYOLO utilizes three separate \ac{YOLO} heads that work on different image resolutions to estimate the state of the object, as illustrated in Figure \ref{fig:ComplexYolo_Architecture}. An extension incorporating complex angle regression through an \ac{E-RPN}, explained in \cite{Reference:Simon_Complex-YOLO_Real_time_3D_Object_Detection_on_Point_Clouds}, facilitates the estimation of the object's center of mass ($x_\zeta$, $y_\zeta$), dimensions ($w_\zeta$, $l_\zeta$) and the object's heading $\psi$. Initially, a method for estimating the heading is integrated. However, an adjustment was necessary given that conventional camera images usually have the coordinate origin in the upper left corner, which is not true in this scenario \cite{cam_img_Gao2021}. Here, the coordinate origin is positioned in the center of the image, requiring an additional adaptation of the \ac{E-RPN} for accurate object parameter estimation. The resulting target state vector $\tau^i$ can be formulated as
\vspace{2mm}
\begin{align}
    \tau^i = [x_{\zeta}^i,\ y_{\zeta}^i, \ w_{\zeta}^i,\ l_{\zeta}^i, \ \psi^i]^T.
\label{eq:ComplexYOLO_output_vector}
\end{align}
A radar sensor can measure the radial velocities of an object but not its absolute velocity, which depends on the object's heading. To estimate the target velocity, an \ac{MLP} network is integrated into the ComplexYOLO framework, extending the target state vector of the output. 
\section{Database}
\label{DB}
\subsection{Data acquisition}
Public radar databases like NuScenes face notable limitations, including sparse radar point clouds per frame, the lack of high-resolution \ac{md} signatures, and no relevant near-field pre-crash radar frames \cite{Reference:Scheiner_Object_detection_for_automotive_radar_point_clouds_a_comparison, early_fusion}. Various methods address the issue of sparse point clouds by integrating multiple scans to create a more comprehensive representation of objects \cite{Reference:Niederloehner_Self-Supervised_Velocity_Estimation_for_Automotive_Radar_Object_Detection_Networks,Reference:Koehler_Improved_Multi_Scale_Grid_Rendering_of_Point_Clouds_for_Radar_Object_Detection_Networks}. However, these approaches undermine the \ac{CNN}s' ability to operate effectively frame-by-frame. In time-sensitive pre-crash situations, high-resolution radars provide dense radar point clouds, delivering detailed insights into an object's shape and high-resolution features, which significantly improve the accuracy of target state estimation. This level of precision is critical for ensuring the timely activation of safety systems. Furthermore, public databases typically focus on standard road traffic scenarios in rural and urban environments, neglecting relevant pre-crash scenarios. For this reason, dedicated radar databases were explicitly created for \ac{ai}-based pre-crash object detection.

\subsubsection{Series radar database}
\begin{figure}[h!]
\vspace{-3mm}
  \centering
  \subfigure[Series Radar Database]{\includegraphics[height=3.4cm]{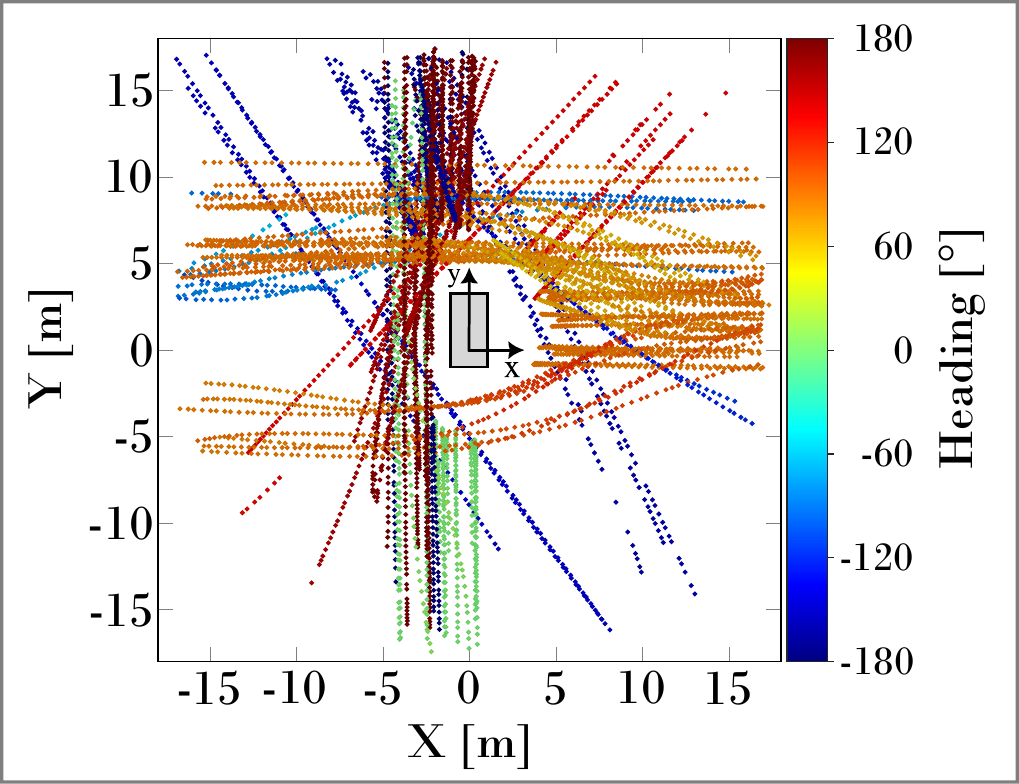}}
  \label{fig:trj-conti}
  \hfill
  \subfigure[High-Resolution Database]{\includegraphics[height=3.4cm]{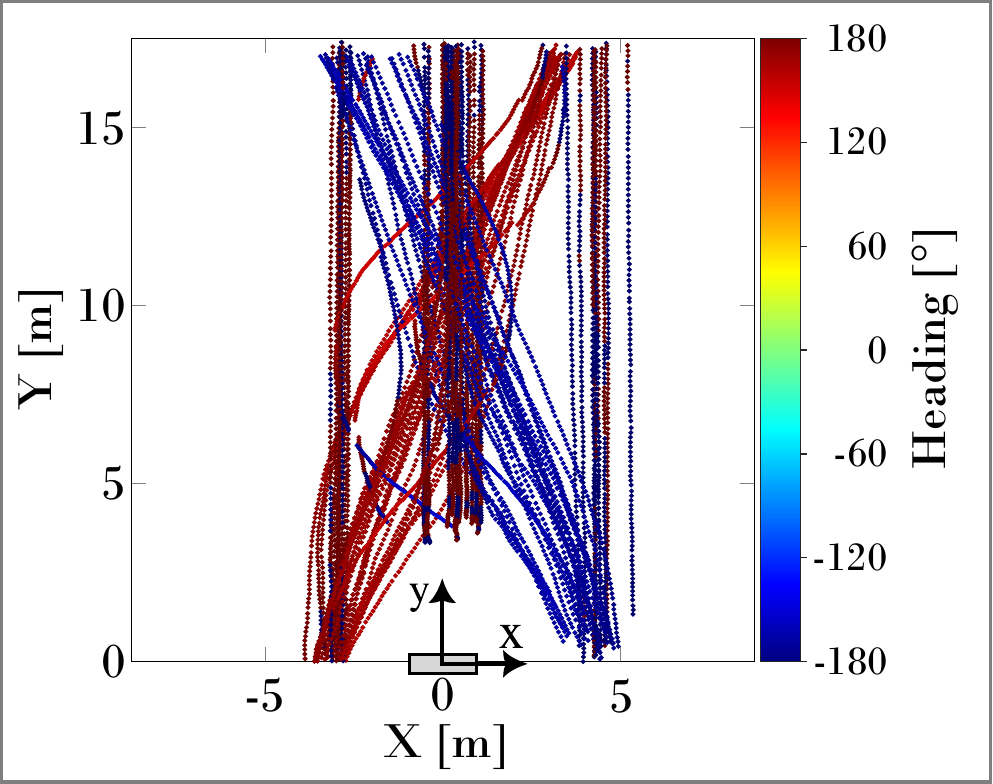}}
  \label{fig:trj-rdl}
  \vspace{-2mm}
  \caption{Visualization of the trajectories in combination with the ground truth and the colored heading information.}
  \label{fig:database}
\end{figure}

A collaborative effort with Continental Lindau created a novel series radar database without the limitations of the public database. Data were collected in an indoor and an outdoor research facility, incorporating various environmental conditions. A complete measurement catalog containing various driving scenarios, including overtaking, overlapping, diagonal approaches, and velocities ranging from \SI{10}{\kilo\meter/\hour} to \SI{30}{\kilo\meter/\hour}. These scenarios are visualized in Figure \ref{fig:database}(a), with the test vehicle positioned at the origin. Each data point represents the target's reference center point in different driving scenarios. The colors indicate the target's heading ground truth. The primary focus was on acquiring data in pre-crash scenarios to enable detailed analyses of the model's performance.

\noindent The Continental test vehicle was equipped with four short-range radars at each corner and remained stationary during data collection. An RTK referencing system was installed on the target vehicle to annotate the radar data, recording position, velocity, orientation, and timestamps throughout the data collection.
The radar points from all four corner sensors were adjusted to align with the center of the vehicle's rear axle. This adjustment aimed to reduce the sparsity often seen in single-series radar sensors, thereby increasing the point cloud density in each frame. The time synchronization for each radar frame is achieved by aligning the timestamps from each radar sensor cycle. Consequently, every radar frame is synchronized with the corresponding timestamp of the reference system for labeling purposes. The database consists of 8646 labeled radar frames.

Continental's short-range radar operates at \SI{77}{\giga\hertz} and has a cycle time around \SI{50}{\milli\second}. It provides a maximum range of about \SI{100}{\meter} and a range resolution of \SI{0.63}{\meter}. The maximum speed range is \SI{\pm 300}{\kilo\meter/\hour} with a resolution of \SI{0.34}{\kilo\meter/\hour} and a maximum azimuth \ac{FoV} of \SI{\pm 90}{\degree} with a \ac{FoV} accuracy of \SI{\pm 1.3}{\degree}.

\subsubsection{High resolution radar database}
\label{subsec:High_res_DB}
Current limitations in automotive radar databases are primarily due to constraints in the number of frames and the quality and density of radar point clouds. These constraints present significant challenges for radar-based deep learning approaches, especially in pre-crash scenarios where precise and accurate object detection is critical. The sparsity of point clouds greatly restricts performance in these situations \cite{Reference:Schumann_RadarScenes_A_Real_World_Radar_Point_Cloud-Data_Set_for_Automotive_Applications}. With recent advances in radar technologies, high-resolution radars can generate denser point clouds, offering new potential for analyzing the impact of \ac{md} on object detection in pre-crash scenarios. Multiple custom databases were created to support this analysis. These databases allow for a controlled assessment by focusing on a single vehicle, laying the groundwork for more comprehensive research.

\begin{figure}[t!]
  \centering
  \subfigure[SRR-NativeDB]{\includegraphics[width=0.48\columnwidth]{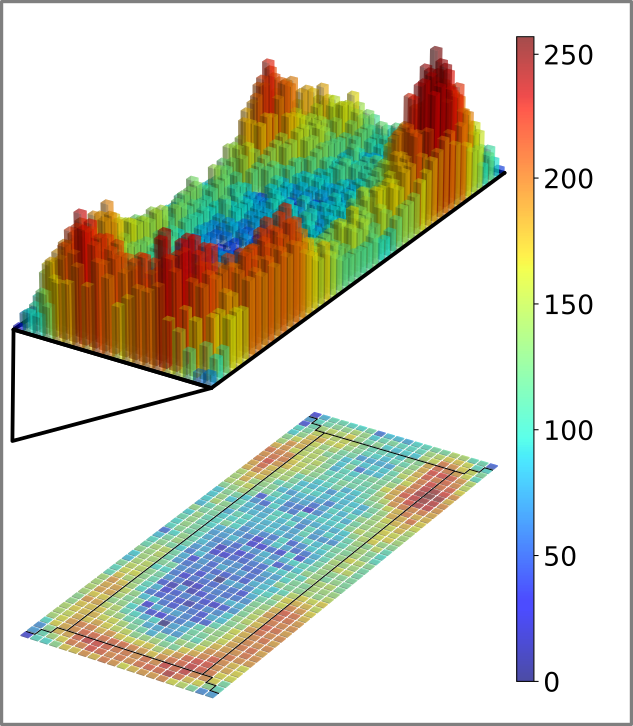}}
  \label{fig:conti-all}
  \hfill
  \subfigure[RDL-NativeDB]{\includegraphics[width=0.48\columnwidth]{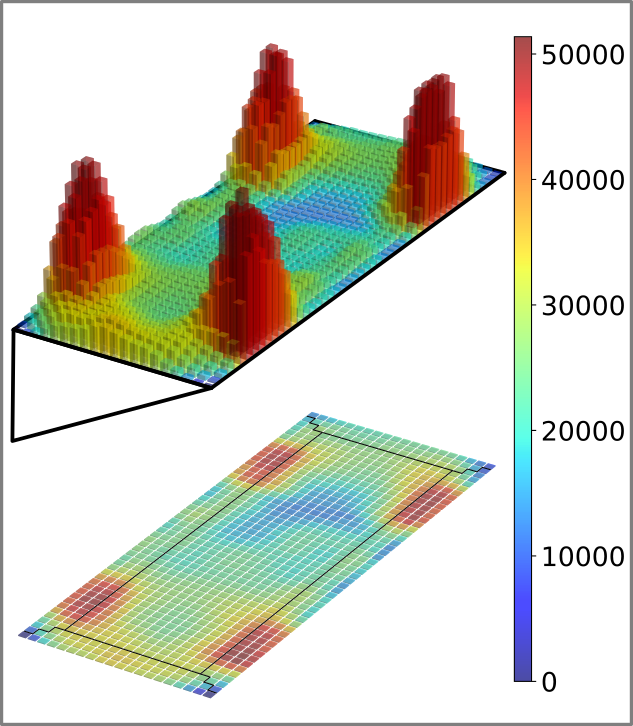}}
  \label{fig:rl-all}
  \vspace{-1mm}
  \subfigure[RDL-WheelExtDB]{\includegraphics[width=0.48\columnwidth]{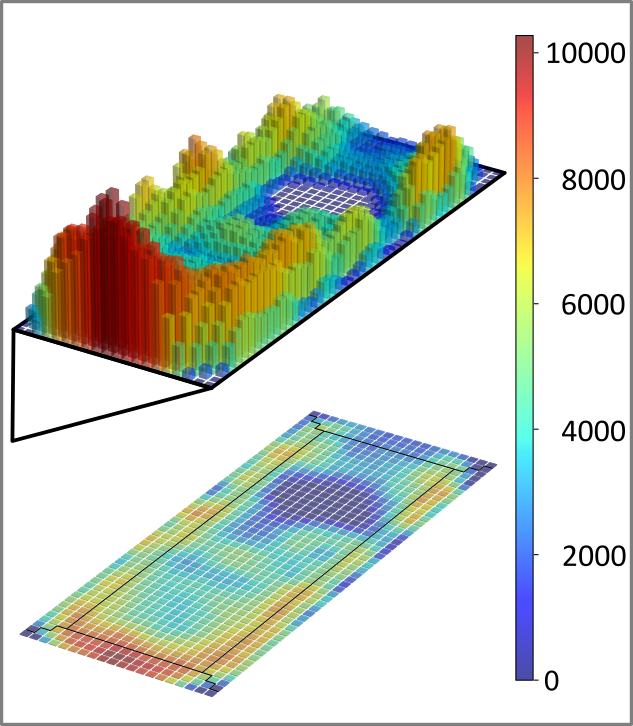}}
  \label{fig:rl-chassis}
  \hspace{3pt}
  \subfigure[RDL-WheelDB]{\includegraphics[width=0.48\columnwidth]{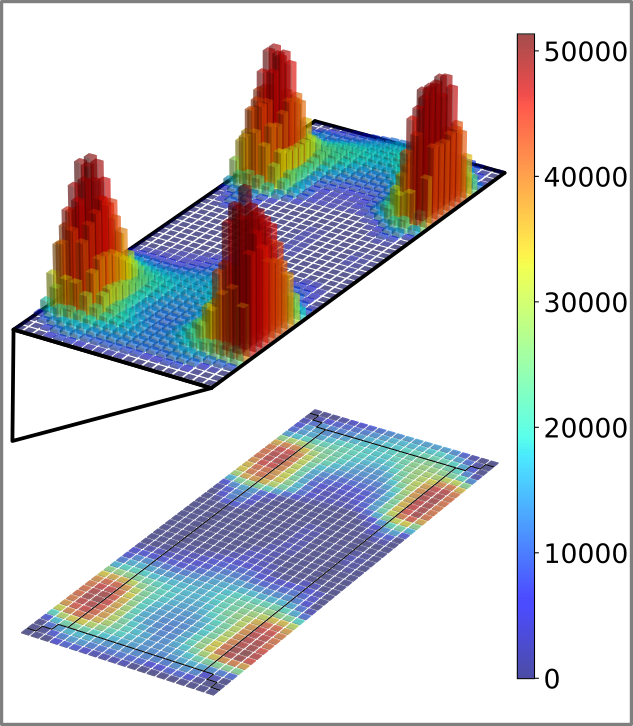}}
  \label{fig:rl-wheels}  
  \vspace{-2mm}
  \caption{Visualization of the cumulative series and research radar databases as histograms and heatmaps. The color of the bars represent the detection count in a \num{10}×\SI{10}{\centi\meter} bin. The black lines in the heatmaps represent the chassis edges.}
  \label{fig:data-conti-rl}
  \vspace{-4mm}
\end{figure}

Data collection for the high-resolution radar database occurred in a controlled testing environment, ensuring consistent and reproducible conditions. The test setup consists of well-defined trajectories with close linear and dynamic vehicle passes, featuring variations in overlap, braking maneuvers, and starting angles, as depicted in Figure \ref{fig:database}(b). All measurements were performed consistently at \SI{20}{\kilo\meter/\hour} to ensure uniformity.

The radar system remained stationary throughout the data collection process, positioned at a height of \SI{0.3}{\meter} in the center of the testing facility. An INRAS Radarlog is used, and the system operates at \SI{76}{\giga\hertz} \ac{fmcw} signals with chirp sequence modulation. The key parameters included a configurable bandwidth of \SI{1.5}{\giga\hertz} (\SIrange{76}{77.5}{\giga\hertz}), a chirp duration of \SI{51.2}{\micro\second}, and a chirp repetition rate of \SI{60}{\micro\second}, encompassing 512 chirps within each coherent processing interval. This configuration resulted in a range gate length of \SI{0.1}{\meter} and a maximum unambiguous measurable distance of \SI{25}{\meter}. Achieving a Doppler resolution of \SI{0.06}{\meter/\second}, the radar front-end featured a \ac{ula} with one transmit and 16 receive antennas, providing an angular azimuth accuracy of \SI{0.12}{\degree}.

An \ac{IPS} was used to continuously track the position, speed, and timestamps of the target during all measurements, ensuring precise reference data. In post-processing, each radar frame was synchronized with the timestamps from the reference system, significantly streamlining the labeling process.
The recorded RadarLog pre-crash is named RDL-NativeDB and shown in Figure \ref{fig:data-conti-rl}(b), includes 11,304 radar frames.
The wheel extraction database called RDL-WheelExtDB, depicted in Figure \ref{fig:data-conti-rl}(c), is a modified version of the RDL-NativeDB, with wheel detections removed using the wheel extraction algorithm described in Section \ref{sec:wheel_ex}. The extracted wheel data is stored separately in a dedicated RDL-WheelDB database for comparison purposes.
The RDL-ChassisDB is an adjusted version of the native database, presented in \ref{fig:data-conti-rl}(d). In this database, the \ac{md} velocities of the wheels have been changed to align with the vehicle chassis velocity. This adjustment was achieved using the Chassis Velocity Algorithm detailed in Section \ref{sec:chassis_vel}. The RDL-ChassisDB was created to better analyze the impact of the \ac{md}, as removing the wheels in the RDL-WheelExtDB also alters the number of detections.

These modifications to the radar database serve as an essential component of an in-depth analysis to evaluate the impact of the detection count and \ac{md} on the network's performance.

\subsection{Database - radar detection distribution}
\label{subsec:db-dist}
In the following section, the databases of the previous sections are analyzed and compared. It is crucial to note the disparity in the number of measurements among these databases, resulting in distinct frame counts. The series radar sensor database (SRR) contains 8646 frames, 7454 of which are devoted to training and 1192 for testing. In contrast, the research radar sensor database RadarLog (RDL) contains 11,304 frames, 8062 of which are assigned for training and 3242 for testing.

\begin{table}[h!]
    \centering
    \caption{comparison of series and high-resolution radar databases, highlighting detection counts and distributions.}
    \label{tab:Distribution}
    \setlength{\tabcolsep}{3pt}
    \renewcommand{\arraystretch}{1.15}
    \adjustbox{width=\columnwidth}{
    \begin{tabular}{|c|c|c|c|c|}
    \hline
        \multirow{2}{*}{\textbf{Areas}} & \rule{0pt}{2ex}\textbf{SRR} &  \multicolumn{3}{c|}{\rule{0pt}{2ex}\textbf{RadarLog (RDL)}}  \\
        \cline{2-5}
            & \rule{0pt}{2ex}NativeDB & \rule{0pt}{2ex}NativeDB & \rule{0pt}{2ex}WheelDB & \rule{0pt}{2ex}WheelExtDB \\
        \hline
        \multirow{1}{*}{\rule{0pt}{1.5ex}All Detections} & \rule{0pt}{2ex}\num{81395} & \rule{0pt}{2ex}\num{8039326} & \rule{0pt}{2ex}\num{5391276} (\SI{67}{\percent}) & \rule{0pt}{2ex}\num{2620614} (\SI{33}{\percent}) \\
        \hline
        \multirow{1}{*}{\rule{0pt}{1.5ex}All Edges} & \rule{0pt}{2ex}\num{40591} (\SI{50}{\percent}) & \rule{0pt}{2ex}\num{4450437} (\SI{55}{\percent}) & \rule{0pt}{2ex}\num{3213995} (\SI{60}{\percent}) & \rule{0pt}{2ex}\num{1219715} (\SI{47}{\percent}) \\
        \hline
        \multirow{1}{*}{\rule{0pt}{1.5ex}Front} & \rule{0pt}{2ex}\num{7459} (\SI{19}{\percent}) & \rule{0pt}{2ex}\num{379383} (\SI{9}{\percent}) & \rule{0pt}{2ex}\num{79518} (\SI{3}{\percent}) & \rule{0pt}{2ex}\num{299429} (\SI{25}{\percent}) \\
        \hline
        \multirow{1}{*}{\rule{0pt}{1.5ex}Back} & \rule{0pt}{2ex}\num{3829} (\SI{10}{\percent}) & \rule{0pt}{2ex}\num{208638} (\SI{5}{\percent}) & \rule{0pt}{2ex}\num{122306} (\SI{4}{\percent}) & \rule{0pt}{2ex}\num{85659} (\SI{7}{\percent}) \\
        \hline
        \multirow{1}{*}{\rule{0pt}{1.5ex}Left} & \rule{0pt}{2ex}\num{16853} (\SI{41}{\percent}) & \rule{0pt}{2ex}\num{2044870} (\SI{46}{\percent}) & \rule{0pt}{2ex}\num{1629655} (\SI{50}{\percent}) & \rule{0pt}{2ex}\num{404433} (\SI{33}{\percent}) \\
        \hline
        \multirow{1}{*}{\rule{0pt}{1.5ex}Right} & \rule{0pt}{2ex}\num{12450} (\SI{30}{\percent}) & \rule{0pt}{2ex}\num{1817546} (\SI{40}{\percent}) & \rule{0pt}{2ex}\num{1382516} (\SI{43}{\percent}) & \rule{0pt}{2ex}\num{430194} (\SI{35}{\percent}) \\
        \hline
    \end{tabular}
    }
    \vspace{-1mm}
\end{table}

Table \ref{tab:Distribution} and Figure \ref{fig:data-conti-rl} compare the detection count and distribution of the SRR and RDL sensor databases. In addition, the RadarLog section includes not only the native database but also two additional databases: RDL-WheelDB and RDL-WheelExtDB. Figure \ref{fig:data-conti-rl}(a) shows the cumulative detection distribution for series radar sensors, while Figure \ref{fig:data-conti-rl}(b) presents the distribution for research radar sensors. The series radar displays a lower point cloud density and fewer detections along the vehicle's contour compared to the research radar. Table \ref{tab:Distribution} provides a detailed analysis of the detection count, distribution, and edge distribution for both radar types.

The histograms in Figure \ref{fig:data-conti-rl} represent the general distribution of the detection points in the database. The vehicle bounding box is divided into \num{10}×\SI{10}{\centi\meter} areas. The color and height of the bars indicate the number of detection points in each grid. The black lines on the heatmap beneath the histogram indicate the delineation of vehicle edges, determined by the radar's ability to penetrate approximately \SI{30}{\centi\meter}. Figure \ref{fig:data-conti-rl}(b) visualizes the summation of the entire radar data base with pronounced wheel detections and \ac{md} signatures.

Table \ref{tab:Distribution} provides a detailed analysis of the point distribution. The category "All Detections" shows the database's total number of detection points. "All Edges" represents the combined number of detections across all edges, with the percentage in parentheses indicating the proportion relative to the total detection points. The table also includes specific edge categories, front, back, right, and left, along with their respective detection point counts and percentages, relative to "All Edges", shown in parentheses.

\subsection{Database - radar range-dependent detection distribution}
\label{subsec:db-range-dist}

\begin{figure}[t!]
  \centering
  \subfigure[RDL-NativeDB $>$\SI{12}{\meter} ]{\includegraphics[width=0.48\columnwidth]{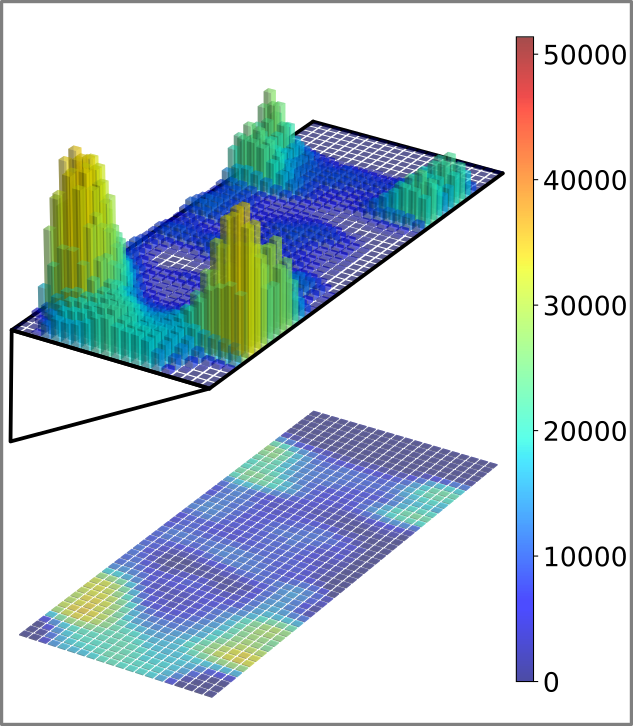}\label{fig:data-range-5m}}
  \hfill
  \subfigure[RDL-NativeDB \SIrange{5}{12}{\meter}]{\includegraphics[width=0.48\columnwidth]{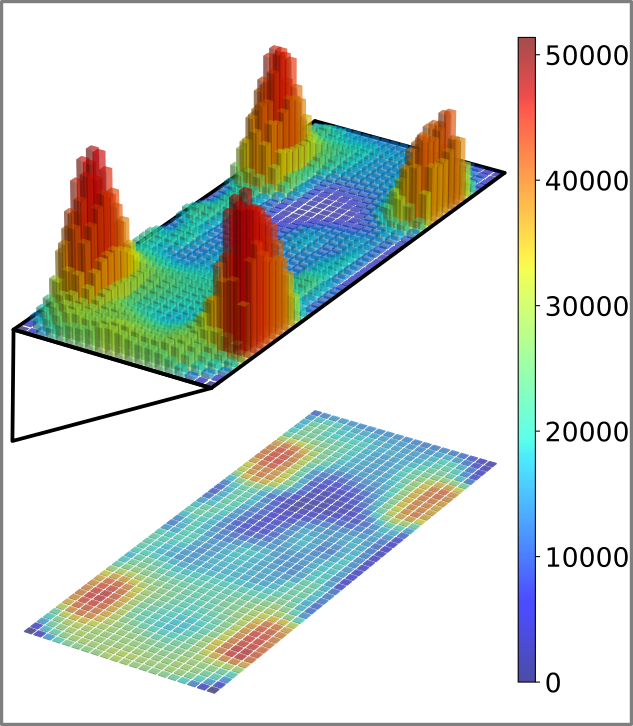}}
  \label{fig:data-range-12m}
  \vspace{-1mm}
  \subfigure[RDL-NativeDB $<$\SI{5}{\meter}]{\includegraphics[width=0.48\columnwidth]{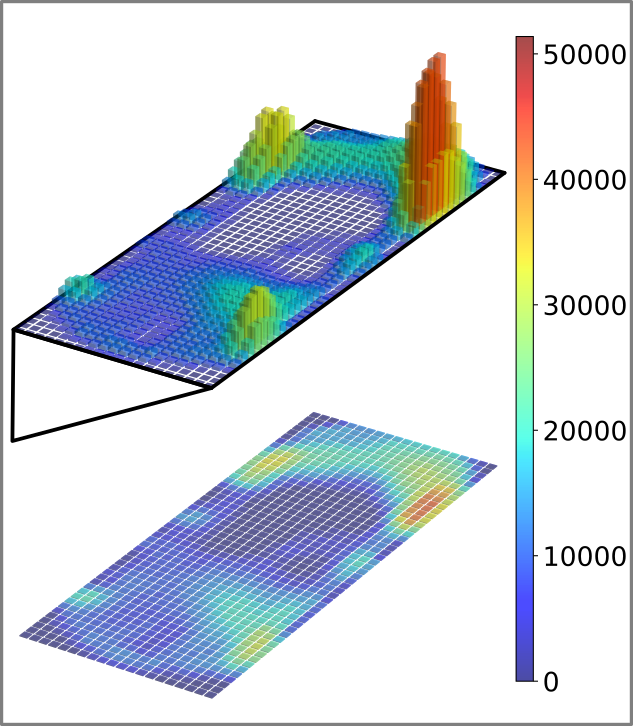}}
  \label{fig:data-range-max}
  \hspace{3pt}
  \subfigure[RDL-NativeDB - RangeGraph]{\includegraphics[width=0.48\columnwidth]{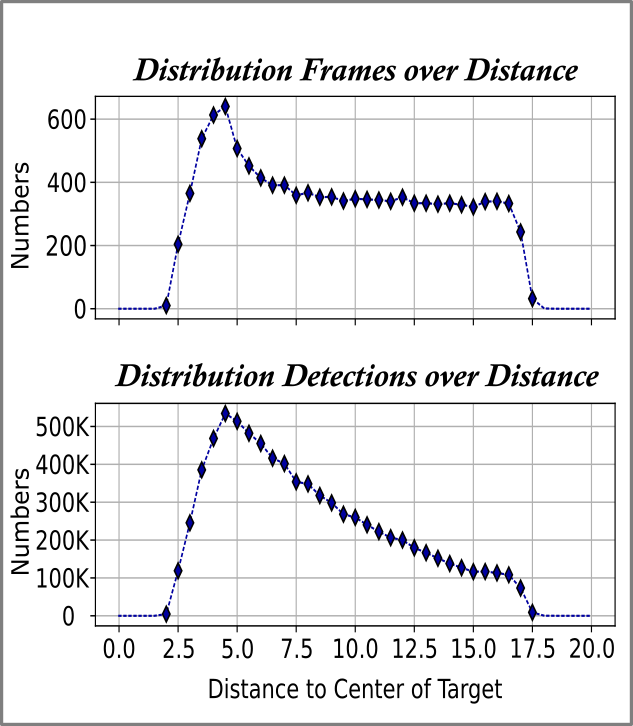}}
  \label{fig:data-range-graph}
  \vspace{-1mm}
  \caption{Visualization of the RDL-NativeDB with three distinct range sections: below 5 meters, 5-12 meters, and above 12 meters. The color of the bars represent the detection count in a \num{10}×\SI{10}{\centi\meter} bin. The black lines in the heatmaps represent the chassis edges. The two graphs in subfigure (d) represent the distribution of detections and frames across the whole distance.}
  \label{fig:data-range-dist}
  \vspace{-3mm}
\end{figure}

The following section introduces the distance-based data distribution, improving the understanding of the database. For this purpose, the native RadarLog database has been divided into three distance ranges: $>$\SI{12}{\meter}, \SIrange{5}{12}{\meter} and $<$\SI{5}{\meter}.
The range above $>$\SI{12}{\meter} reflects distances most relevant to maneuvering and situational awareness in order to prevent accidents by enabling early object detection. The \SIrange{5}{12}{\meter} range captures detailed contours and radar features, like \ac{md} signatures of the approaching vehicle. The range less than $<$\SI{5}{\meter} corresponds to the critical near-field region, where an accurate and fast detection of close objects is essential for vehicle safety systems. The point distribution for each range is represented as a histogram in the Figures \ref{fig:data-range-dist}(a-c) comparable to those of the previous Section \ref{subsec:db-dist}. 

In Figure \ref{fig:data-range-dist}(a), frames beyond \SI{12}{\meter} display a higher concentration of detections towards the front of the target, with less pronounced visibility toward the sides. The \ac{md} signatures of the front wheels are observable, occasionally accompanied by the visibility of the rear wheels as well. In the range from \SI{5}{\meter} to \SI{12}{\meter}, Figure \ref{fig:data-range-dist}(b) exhibits a detailed detection distribution along all edges, accurately representing the contour of the car with the \ac{md} signatures of all four wheels visible. Visualization \ref{fig:data-range-dist}(c) outlines the relevant pre-crash near-field within \SI{5}{\meter}, capturing detections mainly on the rear and side of the target vehicle. When observing the wheels during passing and avoidance maneuvers, the \ac{md} signatures become vastly blurred within the chassis. In \ac{ol} scenarios, they are primarily undetectable due to concealment.

To provide a detailed overview of the detection point and frame distribution, the diagrams in Figure \ref{fig:data-range-dist}(d) illustrate the number of detections and frames throughout the range. It is important to note that the distances are measured from the center of the detected vehicle, which means that distances below \SI{5}{\meter} and beyond \SI{17}{\meter} only capture part of the vehicle. Both number of frames and detections peak between \SI{2.5}{\meter} and \SI{5}{\meter}. Beyond this range, the number of frames remains consistent from 6 to 16 meters before sharply dropping to zero at the maximum range. Similarly, detections gradually decline beyond \SI{5}{\meter}, with a sharp decrease after \SI{16}{\meter}, reflecting the pattern observed in frame counts.
\section{Results}
\label{results}
\subsection{Training}
The anchor-based model was trained using various combinations of RGB image resolutions, radar image dilation configurations, and different \ac{SE} sizes. The \ac{roi} covers a \num{40}×\SI{40}{\meter} area, with the origin centered on the RGB map. Empirical studies have shown that the optimal resolution for high-resolution radar data is \num{416}×\num{416}, corresponding to a grid size of \num{0.1}×\SI{0.1}{\meter}, closely matching the radar resolution of approximately \SI{0.1}{\meter}.
Furthermore, the radar-image dilation method using a \ac{SE} size \num{3}×\num{3} and normalized values between -1 and 1 significantly reduces \ac{fn} detections. The model demonstrated optimal object detection and regression performance using the \ac{adam} optimizer with a learning rate of \num{0.001}. The parameterization of the model is thoroughly examined and analyzed in the context of research and series radar sensors.

Following the hold-out method, the databases were split into \SI{70}{\percent} for training, with frames arranged in random order, and \SI{30}{\percent} for testing, consisting of complete measurement runs excluded from the training dataset.
The model was trained on the native high-resolution radar database (RDL-NativeDB). For testing, both native and modified databases, as described in Section \ref{subsec:High_res_DB}, were used, and in the case of series radar data, training and testing relied solely on the native database.

The supplementary \ac{MLP} network was trained using ground truth data, which included labeled frames with positional coordinates (x, y), heading information and the mean Doppler velocity $\overline{v_d}$ derived from radar detections within the target's ground truth. The output frames of the ComplexYOLO model, containing the predicted information, were used for testing. The database was partitioned into training and testing sets consistent with the ComplexYOLO approach.
\vspace{1mm}
\subsection{Radar Feature Analysis for AI Pre-Crash Object Detection}
\label{subsec:Feature_Analysis}
\begin{figure}[t]
  \centering
  \subfigure[X-Y plane: Passing scenario.]{\includegraphics[width=0.48\columnwidth]{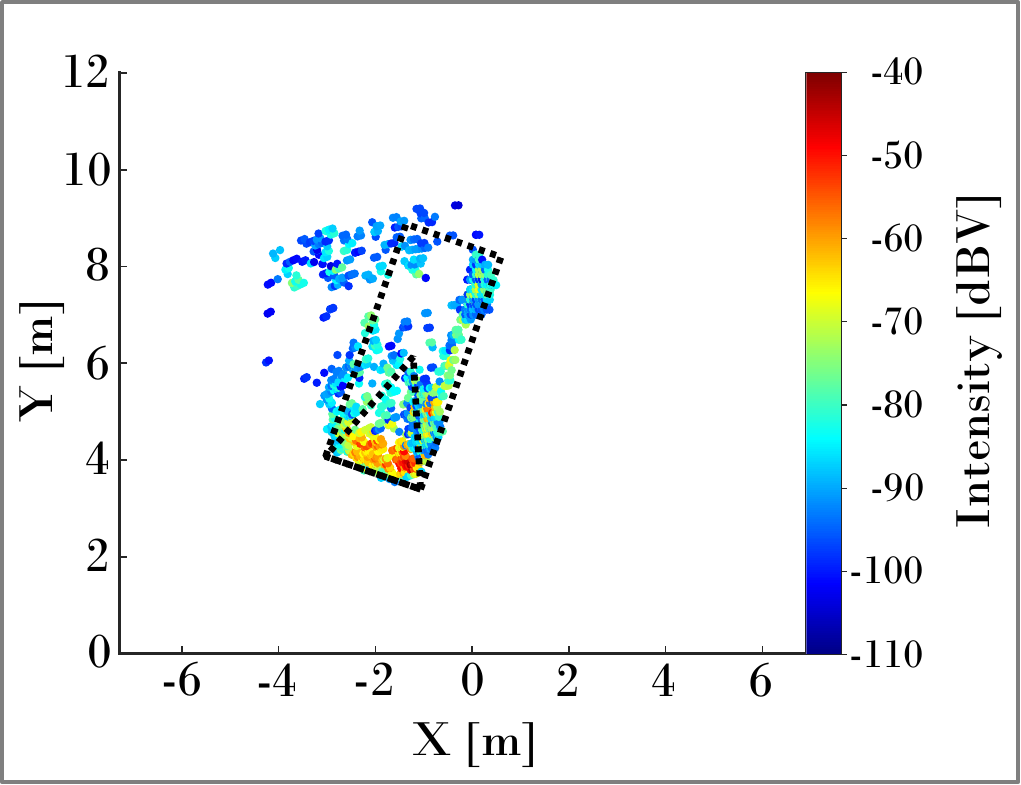}}
  \label{fig:x_y_avoi}
  \subfigure[R-V plane: Passing scenario.]{\includegraphics[width=0.48\columnwidth]{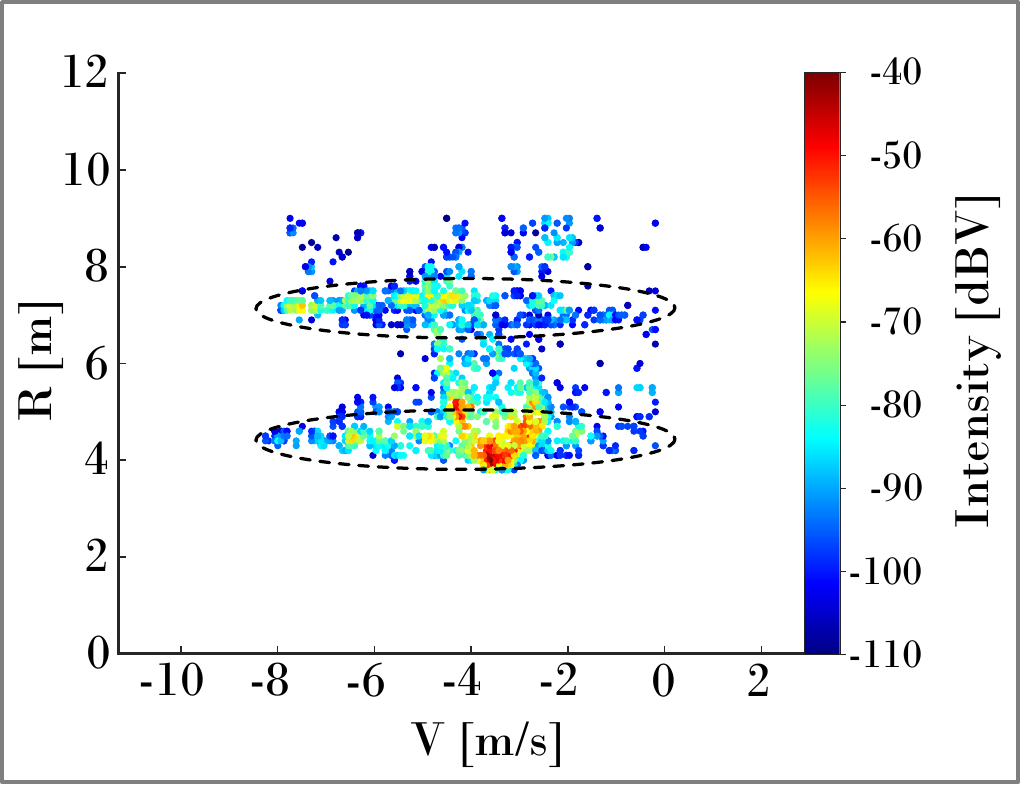}}
  \label{fig:r_v_avoi}
  \vspace{-2mm}
  \subfigure[X-Y plane: Overlap scenario.]{\includegraphics[width=0.48\columnwidth]{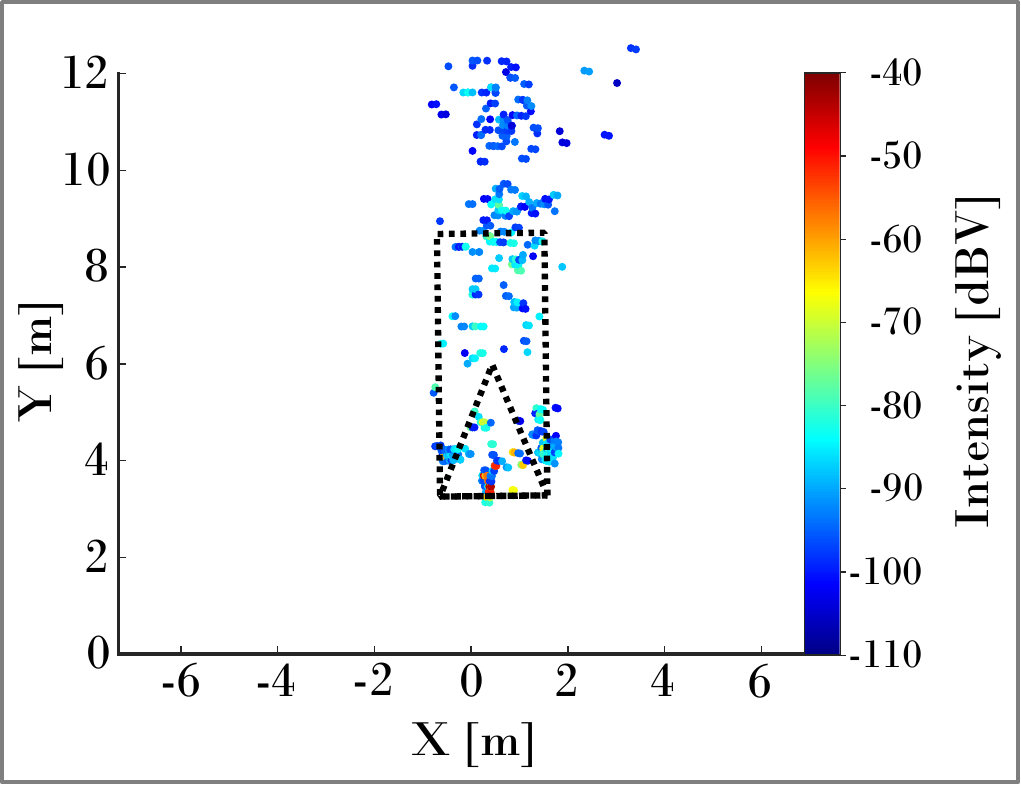}}
  \label{fig:x_y_ol}
  \subfigure[R-V plane: Overlap scenario.]{\includegraphics[width=0.48\columnwidth]{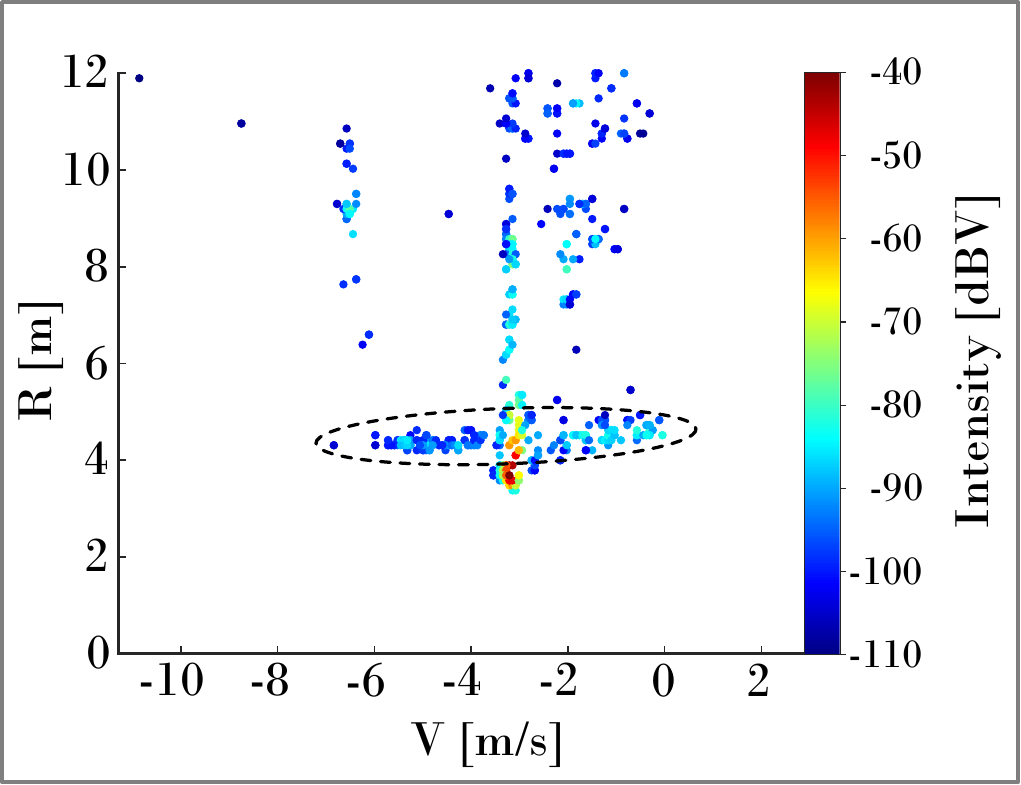}}
  \label{fig:r_v_ol}  
  \vspace{-1mm}
  \caption{Exemplary visualization of radar snapshots in x-y planes with reference in black dotted lines and r-v planes with \ac{md} signatures in black dotted lines. Fig. \ref{fig:range_doppler_xy}(a),(b) illustrates a passing scenario with multiple radar detections along the vehicle's contour forming well-defined L-Shape and \ac{md} signatures. Fig. \ref{fig:range_doppler_xy}(c),(d) demonstrates reduced radar detections and \ac{md} signatures in \ac{ol} scenarios with addition radar multipath reflections.}
  \label{fig:range_doppler_xy}
  \vspace{-4mm}
\end{figure}
The use of a high-resolution research radar sensor offers a significant improvement in detecting vehicle contours and capturing critical kinematic features, such as Doppler velocities from the vehicle chassis and \ac{md} signatures from rotating wheels. Figures \ref{fig:data-conti-rl}(a) and (b) show the cumulative distribution of detections for both series and high-resolution native radar databases. A key observation is the substantially higher detection count provided by the research radar sensor, along with more distinct detections of the wheels compared to the series radar. When extracting wheel detections from the high-resolution database, the resulting point distribution along the vehicle's contour (Figure \ref{fig:data-conti-rl}(c)) closely resembles that of the series radar (Figure \ref{fig:data-conti-rl}(a)), but with a notably higher density of detections. Furthermore, as shown in Figure \ref{fig:data-conti-rl}(d), the wheels detections constitute a majority of the overall detections, underscoring the improved performance of the research radar sensor.
\begin{figure}[b]
\vspace{-7mm}
  \centering
  \subfigure[Passing]{\includegraphics[height=0.55\columnwidth]{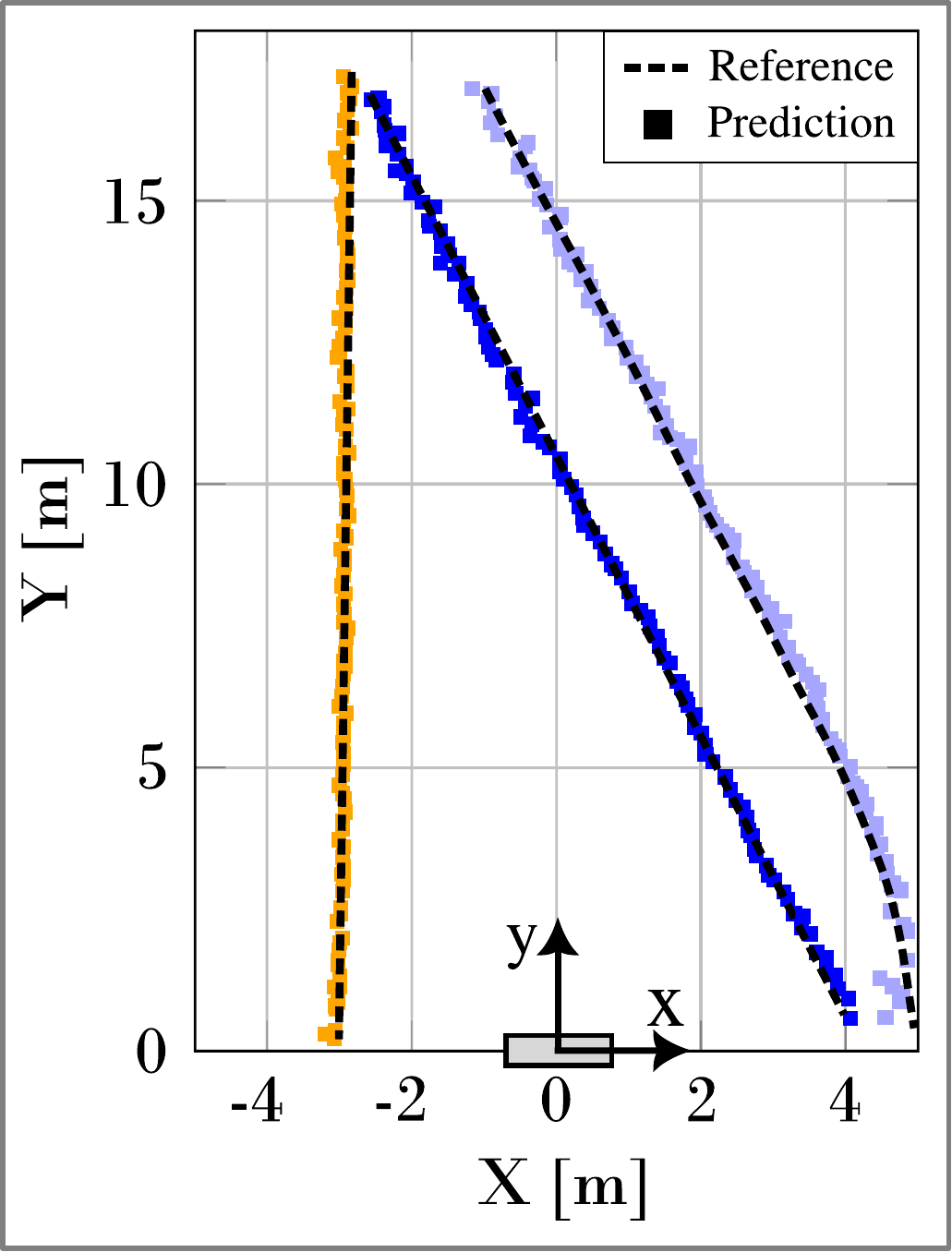}}
  \label{res:overlapscen}
  \subfigure[Overlap]{\includegraphics[height=0.55\columnwidth]{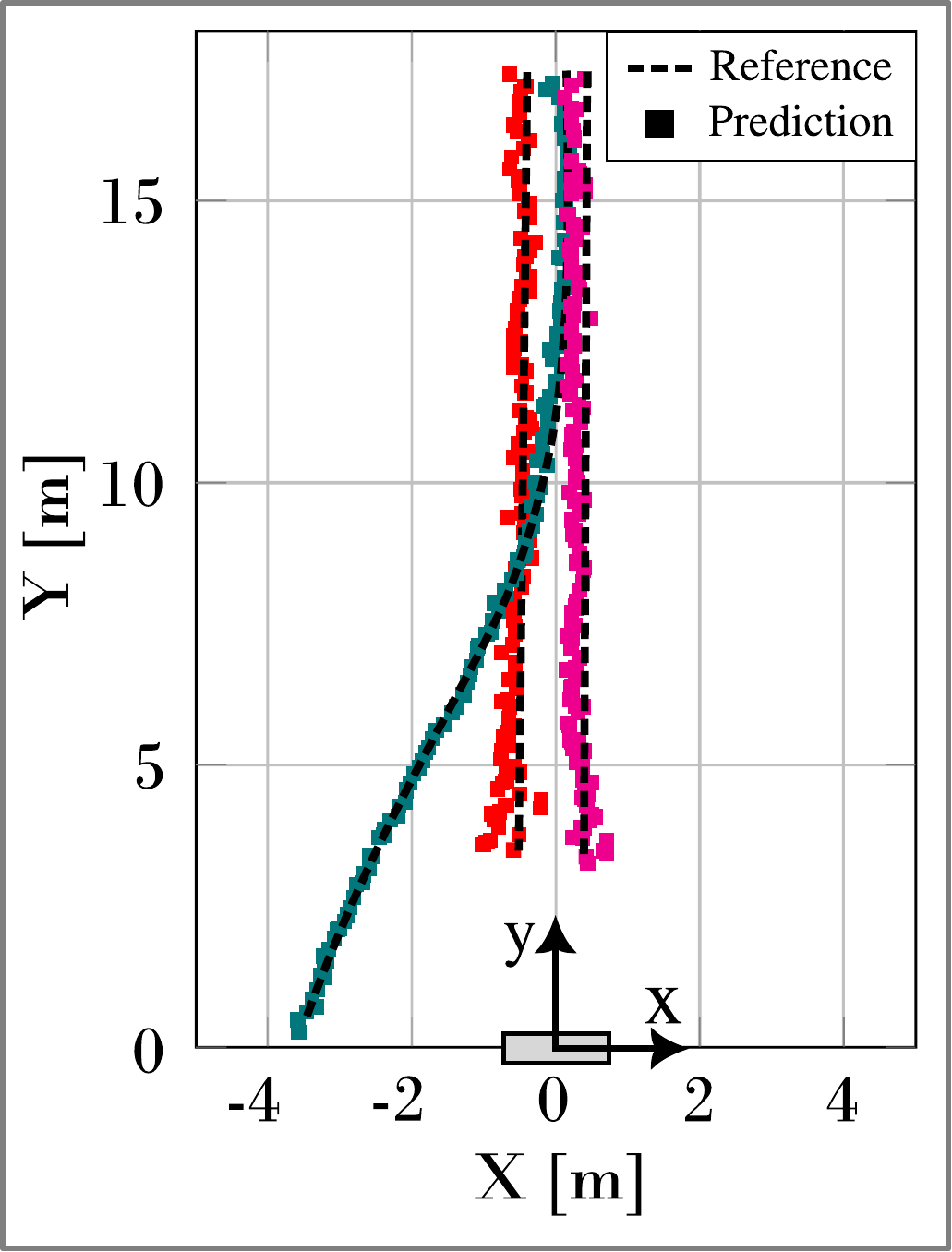}}
  \label{fig:passing}
  \vspace{-2mm}
  \caption{Exemplary illustration of color-coded results of different measurement runs displaying the reference (\textcolor{black}{\rule[0.5ex]{0.3em}{.2pt}\hspace{0.2em}\rule[0.5ex]{0.3em}{.2pt}}\hspace{0.2em}\rule[0.5ex]{0.3em}{.2pt}) and the model predictions (\textcolor{black}{\rule{0.5em}{0.5em}}) with high-resolution radar data.}
  \label{res:overviewTrajectories}
\end{figure}
The density of the point cloud, the number of detections per frame, and the detection of unique wheel characteristics are significantly influenced by the distance of the target, as shown in Figures \ref{fig:data-range-dist} (a)-(c). Furthermore, variations in point cloud density are closely related to driving maneuvers, as shown in Figures \ref{fig:range_doppler_xy}(a) and (c).
Additionally, these maneuvers affect the radar's ability to capture high-resolution features, such as rotating wheels' \ac{md} signatures. The experimental results demonstrate that these signatures can be detected up to a distance of \SI{14}{\meter}.
As a result, the model's performance needs to be evaluated according to the different distance intervals to provide more information on the impact of the wheel detections and the \ac{md} velocities, especially in the pre-crash relevant range below \SI{5}{\meter}.

\setlength{\dashlinedash}{1pt}
\setlength{\dashlinegap}{2pt}
\newcommand{\scaletab}{1.1}
\begin{table}[t]
    \centering
    \caption{rmse metrics on native high-resolution radar data.}
    \label{tab:Overview_RadarLog_Measurment}
    \setlength{\tabcolsep}{3.5pt}
    \adjustbox{width=0.9\columnwidth}{\begin{tabular}{!{\vrule}c!{\vrule}c!{\vrule}cccc!{\vrule}}
        \hline
        \multirow{1}{*}{\textbf{Scenario}} & \multirow{1}{*}{\textbf{Variant}} & \scalebox{\scaletab}{$\rule[-1.3ex]{0pt}{3.ex}\bm{x_{\zeta,\mathrm{e}}}${[\si{\meter}]}} & \scalebox{\scaletab}{$\bm{y_{\zeta,\mathrm{e}}}${[\si{\meter}]}} & \scalebox{\scaletab}{$\bm{\psi_\mathrm{e}}$[\si{\degree}]} & \scalebox{\scaletab}{$\rule[-1.2ex]{0pt}{3.9ex}{\bm{v_{\mathrm{t,e}}}}$[\SI[per-mode=fraction]{}{\kilo\meter\per\hour}]}\\
        \hline
        \multirow{3}{*}{\rule{0pt}{3.5ex}Passing} & Orange &\rule{0pt}{2ex}\hspace{1pt}\num{0.09} & \hspace{4pt}\num{0.08} & \hspace{4pt}\num{1.41} & \hspace{4pt}\num{1.13}\\
        \cdashline{2-6}
        & Light Blue & \rule{0pt}{2.5ex}\hspace{1pt}\num{0.13} & \hspace{4pt}\num{0.07} & \hspace{4pt}\num{3.86} & \hspace{4pt}\num{1.36}\\
        \cdashline{2-6}
        & Dark Blue & \rule{0pt}{2.5ex}\hspace{1pt}\num{0.07} & \hspace{4pt}\num{0.09} & \hspace{4pt}\num{3.03} & \hspace{4pt}\num{1.74}\\
        \hline
        \multirow{3}{*}{\rule{0pt}{3.5ex}Overlap} & Green & \rule{0pt}{2ex}\hspace{1pt}\num{0.07} & \hspace{4pt}\num{0.14} & \hspace{4pt}\num{1.85} & \hspace{4pt}\num{0.61}\\
        \cdashline{2-6}
        & Magenta &\rule{0pt}{2.5ex}\hspace{1pt}\num{0.14} & \hspace{4pt}\num{0.10} & \hspace{4pt}\num{1.98} & \hspace{4pt}\num{1.11}\\
        \cdashline{2-6}
        & Red & \rule{0pt}{2.5ex}\hspace{1pt}\num{0.16} & \hspace{4pt}\num{0.11} & \hspace{4pt}\num{7.02} & \hspace{4pt}\num{1.75}\\
        \hline
    \end{tabular}
    }
    \vspace{-4mm}
\end{table}

\vspace{1mm}
\subsubsection{Analysis of ComplexYOLO's performance in pre-crash object detection in selected scenarios}
Figure \ref{res:overviewTrajectories} shows an overview of the test scenarios, with each maneuver represented by a different color. The scenarios are grouped into passing maneuvers (Figure \ref{res:overviewTrajectories}(a)) and vehicle-to-vehicle \ac{ol} scenarios (Figure \ref{res:overviewTrajectories}(b)). The sensor setup is located at the coordinate origin, where the forward-facing radar is shown as a rectangle. The black dashed lines indicate the reference paths of the target. ComplexYOLO's results, using the native high-resolution radar database, are shown as squares representing the object's \ac{CoG}. The dense point cloud produced in each radar cycle allows for precise object detection and regression.
Overall, adapting the ComplexYOLO model to integrate high-resolution radar data with distinct \ac{md} signatures enables an effective radar-based pre-crash object detection method that functions within a single measurement cycle. This combination of high-resolution radar data and anchor-based \ac{ai} approach delivers a high detection rate with excellent accuracy across various distances, including the critical pre-crash range of less than \SI{5}{\meter}.

ComplexYOLO tends to perform more accurately in passing scenarios, benefiting from better detection distribution along the vehicle's contour and clearer Doppler and \ac{md} signatures, as shown in Figures \ref{fig:range_doppler_xy}(a) and (b).
In contrast, the analysis of the \ac{ol} scenarios - particularly the red and magenta trajectories in Figure \ref{res:overviewTrajectories}(b) - highlights decreasing performance in position estimation. These inaccuracies are due to the weaker visibility of the \ac{md} signatures and lower detection density, which is concentrated at the front edge of the vehicle, as illustrated in Figures \ref{fig:range_doppler_xy}(c) and (d). Additionally, reduced radar feature quality, fluctuations in point cloud density, and the effects of radar multipath reflections in the near-field further contribute to errors in predicting object parameters.

Table \ref{tab:Overview_RadarLog_Measurment} provides a comprehensive summary and categorization of measurement results based on the \ac{rsme} for predicted object parameters, denoted with a subscript $\mathrm{e}$. In particular, the red and magenta scenarios, which represent different \ac{ol} conditions, show increased errors in predicting the object's $x_{\zeta}$ position and, to a lesser extent, the $y_{\zeta}$ position. The critical \SI{50}{\percent} \ac{ol} maneuver, marked in red, exhibits significant errors in position and heading estimation, while the target's velocity $v_t$ remains comparatively less affected.

\subsubsection{Radar Feature Impact on ComplexYOLO's performance - Entire test dataset vs. overlap scenarios}
\setlength{\dashlinedash}{1pt}
\setlength{\dashlinegap}{2pt}
\newcommand{\scalefactor}{1}
\begin{table*}[ht]
    \centering
    \caption{error metrics and false-negatives for predicted object parameters ($\tau$) are analyzed across the native and modified radar databases at various distance intervals accounting all test frames and overlap scenarios only.}
    \label{tab:Feature_Analysis}
    \setlength{\tabcolsep}{2.5pt}
    \adjustbox{width=\textwidth}{\begin{tabular}{!{\vrule}c!{\vrule}ccccc!{\vrule}ccccc!{\vrule}ccccc!{\vrule}ccccc!{\vrule}}
        \hline
       \multicolumn{1}{!{\vrule}c!{\vrule}}{\rule{0pt}{2ex}\textbf{Distance}}
& \multicolumn{5}{c!{\vrule}}{\textbf{Total}}
& \multicolumn{5}{c!{\vrule}}{\textbf{$\mathrm{I_{\text{\fontsize{6}{7}\selectfont \num{1}}}} \geq$} \textbf{\SI{12}{\mathbf{\meter}}}}
& \multicolumn{5}{c!{\vrule}}{\textbf{\SI{12}{\mathbf{\meter}}} \textbf{$> \mathrm{I_{\text{\fontsize{6}{7}\selectfont \num{2}}}} \geq$} \textbf{\SI{5}{\mathbf{\meter}}}}
& \multicolumn{5}{c!{\vrule}}{\textbf{$\mathrm{I_{\text{\fontsize{6}{7}\selectfont \num{3}}}}$ \textless{}} \textbf{\SI{5}{\mathbf{\meter}}}} \\
        \hline
        \raisebox{-0.2ex}{Parameter} & \scalebox{\scalefactor}{$\bm{x_{\zeta,\mathrm{e}}}$} & \scalebox{\scalefactor}{$\bm{y_{\zeta,\mathrm{e}}}$} & \scalebox{\scalefactor}{$\bm{\psi_\mathrm{e}}$} & \scalebox{\scalefactor}{$\rule[-1.3ex]{0pt}{3.ex}\bm{{v_{\mathrm{t,e}}}}$} & \rule{0pt}{2.ex}$\bm{\mathrm{FN}}$ & \scalebox{\scalefactor}{$\bm{x_{\zeta,\mathrm{e}}}$} & \scalebox{\scalefactor}{$\bm{y_{\zeta,\mathrm{e}}}$} & \scalebox{\scalefactor}{$\bm{\psi_\mathrm{e}}$} & \scalebox{\scalefactor}{$\rule[-1.3ex]{0pt}{3.ex}\bm{{v_{\mathrm{t,e}}}}$} & \textbf{FN} & \scalebox{\scalefactor}{$\bm{x_{\zeta,\mathrm{e}}}$} & \scalebox{\scalefactor}{$\bm{y_{\zeta,\mathrm{e}}}$} & \scalebox{\scalefactor}{$\bm{\psi_\mathrm{e}}$} & \scalebox{\scalefactor}{$\rule[-1.3ex]{0pt}{3.ex}\bm{{v_{\mathrm{t,e}}}}$} & $\bm{\mathrm{FN}}$ & \scalebox{\scalefactor}{$\bm{x_{\zeta,\mathrm{e}}}$} & \scalebox{\scalefactor}{$\bm{y_{\zeta,\mathrm{e}}}$} & \scalebox{\scalefactor}{$\bm{\psi_\mathrm{e}}$} & \scalebox{\scalefactor}{$\rule[-1.3ex]{0pt}{3.ex}\bm{{v_{\mathrm{t,e}}}}$} & $\bm{\mathrm{FN}}$ \\ 
        \hdashline
        Unit & $\rule[-1.ex]{0pt}{3.2ex}$[\si{\meter}] & [\si{\meter}] & [\si{\degree}] & [\SI[per-mode=fraction]{}{\kilo\meter\per\hour}] & - & [\si{\meter}]  & [\si{\meter}]  & [\si{\degree}] & [\SI[per-mode=fraction]{}{\kilo\meter\per\hour}] & - & [\si{\meter}] & [\si{\meter}] & [\si{\degree}] & [\SI[per-mode=fraction]{}{\kilo\meter\per\hour}] & - & [\si{\meter}] & [\si{\meter}] & [\si{\degree}] & [\SI[per-mode=fraction]{}{\kilo\meter\per\hour}] & -  \\
        
        \hline   
       \rule{0pt}{2.ex}\multirow{1}{*}{NativeDB} & \num{0.11} & \num{0.13} & \num{4.00} & \num{1.4} & \multirow{1}{*}{\num{34}} & \num{0.11} & \num{0.11} & \num{3.83} & \num{1.26} & \multirow{1}{*}{\num{2}} & \num{0.09} & \num{0.12} & \num{2.63} & \num{1.20} & \multirow{1}{*}{\num{3}} & \num{0.15} & \num{0.16} & \num{6.42} & \num{1.97} & \multirow{1}{*}{\num{29}}\\
           
       \hdashline
       \rule{0pt}{2.ex}\multirow{1}{*}{ChassisDB} & \num{0.11} & \num{0.17} & \num{3.92} & \num{2.59} & \multirow{1}{*}{\textbf{\num{50}}} & \num{0.11} & \num{0.11} & \num{3.78} & \num{2.61} & \multirow{1}{*}{\textbf{\num{7}}} & \num{0.10} & \num{0.12} & \num{2.63} & \num{1.97} & \multirow{1}{*}{\textbf{\num{6}}} & \num{0.15} & \textbf{\num{0.30}} & \num{6.20} & \num{3.71} & \multirow{1}{*}{\textbf{\num{37}}}\\
        
       \hdashline
       \rule{0pt}{2.ex}\multirow{1}{*}{WheelExtDB} & \textbf{\num{0.19}} & \textbf{\num{0.22}} & \num{5.80} & \num{2.40} & \multirow{1}{*}{\textbf{\num{126}}} & \num{0.15} & \num{0.14} & \num{4.87} & \num{2.26} & \multirow{1}{*}{\textbf{\num{9}}} & \textbf{\num{0.20}} & \textbf{\num{0.18}} & \num{5.29} & \num{1.95} & \multirow{1}{*}{\textbf{\num{32}}} & \textbf{\num{0.21}} & \textbf{\num{0.37}} & \num{8.22} & \num{3.52} & \multirow{1}{*}{\textbf{\num{85}}}\\

       \hline
       \rule{0pt}{2.ex}\multirow{1}{*}{NativeDB-OL} & \num{0.14} & \num{0.17} & \num{4.77} & \num{1.46} & \multirow{1}{*}{\num{29}} & \num{0.13} & \num{0.09} & \num{2.97} & \num{0.87} & \multirow{1}{*}{\num{0}} & \num{0.12} & \num{0.20} & \num{2.56} & \num{1.19} & \multirow{1}{*}{\num{0}} & \num{0.21} & \num{0.20} & \num{9.47} & \num{2.52} & \multirow{1}{*}{\num{29}}\\
        
       \hdashline
       \rule{0pt}{2.ex}\multirow{1}{*}{ChassisDB-OL} & \num{0.14} & \textbf{\num{0.26}} & \num{4.60} & \num{2.86} & \multirow{1}{*}{\textbf{\num{37}}} & \num{0.13} & \num{0.09} & \num{2.85} & \num{2.46} & \multirow{1}{*}{\num{0}} & \num{0.12} & \num{0.20} & \num{2.58} & \num{2.45} & \multirow{1}{*}{\textbf{\num{1}}} & \num{0.21} & \textbf{\num{0.49}} & \num{9.16} & \num{4.22} & \multirow{1}{*}{\textbf{\num{36}}}\\
        
       \hdashline
       \rule{0pt}{2.ex}\multirow{1}{*}{WheelExtDB-OL} & \textbf{\num{0.20}} & \textbf{\num{0.31}} & \num{5.84} & \num{2.51} & \multirow{1}{*}{\textbf{\num{91}}} & \num{0.16} & \num{0.14} & \num{4.60} & \num{1.71} & \multirow{1}{*}{\textbf{\num{1}}} & \textbf{\num{0.19}} & \textbf{\num{0.29}} & \num{4.67} & \num{2.06} & \multirow{1}{*}{\textbf{\num{18}}} & \textbf{\num{0.27}} & \textbf{\num{0.56}} & \num{10.22} & \num{4.53} & \multirow{1}{*}{\textbf{\num{72}}}\\
       \hline
    \end{tabular}
    }
    \vspace{-4mm}
\end{table*}

Table \ref{tab:Feature_Analysis} presents a comprehensive analysis of individual radar features and their respective impacts on the accuracy of the object parameter estimation. The evaluation compares the entire test database against critical \ac{ol} collision scenarios, highlighting error metrics and \ac{fn} rates across three different distance intervals $\mathrm{I_1}$ to $\mathrm{I_3}$, organized into ranges above \SI{12}{\meter}, between \SI{12}{\meter} and \SI{5}{\meter}, and below \SI{5}{\meter}. The \textit{Total} section summarizes errors and occurrences of \ac{fn} at all distances. Position errors $x_{\zeta,\mathrm{e}}$ and $y_{\zeta,\mathrm{e}}$ exceeding \SI{5}{\centi\meter}, heading errors $\psi_\mathrm{e}$ exceeding \SI{2}{\degree}, and velocity errors $v_{\mathrm{t,e}}$ exceeding \SI{2}{\kilo\meter/\hour} of the modified databases compared to the native database are highlighted in bold. Furthermore, any instances where modified databases exhibit higher \ac{fn} rates than the native database are also highlighted in bold. 
The RDL-NativeDB test frames demonstrate a notable decrease in the distribution of detection points and \ac{md} signatures within distance intervals $\mathrm{I_1}$ and $\mathrm{I_3}$, in contrast to the interval $\mathrm{I_2}$, as described in Section \ref{subsec:db-range-dist}. This reduction coincides with an increase in errors in both the position and the heading. Furthermore, 29 \ac{fn}s are recorded in the near-field $\mathrm{I_3}$, marking the highest proportion within the RDL-NativeDB. The range distance $\mathrm{I_2}$ shown in Figure \ref{fig:data-range-dist}(b) contains most of the radar frames in the test database and presents minor errors due to the excellent distribution of detections along the vehicle contour and well-pronounced \ac{md} signatures.

Figure \ref{fig:data-range-dist}(d) confirms the error trends and cumulative distributions, illustrating the count of frames and detections over the range. The detections show a prominent peak in the near-field, approximately between \SI{2.5}{\meter} and \SI{6}{\meter}, followed by a gradual decline up to \SI{16}{\meter}, beyond which they drop to zero. The frame count follows a similar pattern, peaking in the near-field, then slightly decreasing before maintaining a steady rate, and finally dropping abruptly to zero around \SI{16}{\meter}.
Similar trends are observed in the RDL-ChassisDB (manipulated \ac{md} velocities) and the RDL-WheelExtDB (removed \ac{md} detections) compared to the RDL-NativeDB.

Furthermore, the influence of \ac{md} velocities on the estimation of kinematic object parameters is minimal when considering all test frames within the RDL-ChassisDB database compared to the origin database. However, in the near-field, $\mathrm{I_3}$ - where radar detection density and feature quality are significantly reduced - the \ac{md} velocities play a more substantial role. Removing the \ac{md} velocities leads to a noticeable increase in \ac{fn}, as shown in Table \ref{tab:Feature_Analysis}.

While the wheels' position information and radar data detections are still included in the RDL-ChassisDB, the complete wheel-related information are extracted in the RDL-WheelExtDB. This significantly decreases the model performance and object detection rate across all test frames. This effect is especially noticeable in the distance intervals $\mathrm{I_2}$ and $\mathrm{I_3}$, where the positional accuracy decreases and the \ac{fn} rate increases.

Table \ref{tab:Feature_Analysis} also shows that critical \ac{ol} scenarios have a significantly greater impact on the model's performance regarding \ac{md} signatures compared to the overall test data. When comparing various distance intervals, increasing position errors emerge across the three databases as the target approaches. Several factors can explain this. In \ac{ol} scenarios, radar sensors detect primarily the front edge of the vehicle, resulting in a concentrated distribution of detections at this edge, as illustrated in Subfig. \ref{fig:range_doppler_xy}(c). This uneven distribution reduces the characteristic L-shape detection pattern typically observed along the vehicle's contour. Furthermore, the \ac{md} signatures are less pronounced in the \ac{ol} scenarios, as shown in Subfig. \ref{fig:range_doppler_xy}(d), compared to passing-by maneuvers. This is mainly due to the reduced visibility of the wheels. Furthermore, the near-field region shows fewer detections, as seen in Figure \ref{fig:data-range-dist}(c-d), resulting in higher position errors. These factors, combined with ghost objects caused by radar multipath reflections in the near-field, significantly contribute to the \ac{fn} within this range segment. The manipulation of the \ac{md} velocities, represented by RDL-ChassisDB, significantly increases the position error $y_{\zeta,\mathrm{e}}$ and leads to a considerable increase in the \ac{fn} rate in the near-field, as shown in Table \ref{tab:Feature_Analysis}. Furthermore, the RDL-WheelExtDB, with all wheel detections completely removed, significantly affects the position error in distance intervals $\mathrm{I_2}$ and $\mathrm{I_3}$, resulting in a higher occurrence of \ac{fn}, especially in the near-field. Within a range of less than \SI{5}{\meter}, the \ac{fn} rates for RDL-NativeDB, RDL-ChassisDB, and RDL-WheelExtDB are \SI{13.94}{\percent}, \SI{17.31}{\percent}, and \SI{34.61}{\percent}, respectively. 

In general, using high-resolution radar features in near-field scenarios significantly improves the accuracy of object parameters while significantly reducing \ac{fn} occurrences. More specifically, using \ac{md} signatures present an opportunity to mitigate positional errors and has a remarkable potential to decrease \ac{fn} rates, particularly for \ac{ai} based pre-crash object detection in the near-field.

\subsection{Radar-Image Dilation for Pre-Crash Object Detection}
Roughly \SI{30}{\percent} of the test database comprises critical vehicle-to-vehicle \ac{ol} scenarios, where most \ac{fn}s occur at close distances, as shown in Table \ref{tab:Feature_Analysis}. These errors are mainly caused by radar multipath reflections. This effect impairs the model's detection capabilities in the near-field, where accurate identification is already challenging.

\begin{figure}[!b]
    \vspace{-3mm}
    \centering
    \definecolor{graphicbackground}{rgb}{0.0667, 0.0667, 0.0667}
\newcommand{\pointMetaMinVal}{-118}
\newcommand{\pointMetaMaxVal}{-60}
\newcommand{\markerSize}{0.5pt}
\newcommand{\Plotheightradar}{1.795cm}
\newcommand{\Plotheight}{1.8cm}

\pgfkeys{/tikz/.cd,
  background color/.initial={rgb,255:red,255; green,255; blue,255},
  background color/.get=\backcol,
  background color/.store in=\backcol,
  background opacity/.initial=1.0,
  background opacity/.get=\backopacity,
  background opacity/.store in=\backopacity,
}

\tikzset{
  background rectangle/.style={
    fill=\backcol,
    draw=black,
    line width=1pt, 
    draw opacity=1,
    draw opacity=0.5,
    fill opacity=\backopacity,
  },
  use background/.style={    
    show background rectangle,
  }
}

\begin{tikzpicture}[use background]
\begin{groupplot}[
        group style={
            group name=my plots,
            group size= 3 by 4,
            horizontal sep=0.1cm,
            vertical sep=0.1cm, 
        },
        scale only axis,
        width=1.\columnwidth,
        xticklabels={},
        yticklabels={},
        grid=both,
        legend style={at={(1,1)},anchor=north east,
        draw=black,
        fill=gray!30,
        fill opacity=1}
      ]
  \nextgroupplot[width=\Plotheightradar,
        height=\Plotheightradar,
        major grid style={draw=none},
        minor grid style={draw=none},
        tick style={draw=none},
        ylabel={\textbf{Input}},
        axis equal,
        ]
    \addplot[
            scatter, 
            only marks,
            scatter, 
            mark=*,
            mark size = \markerSize,
            grid=none,
            point meta=explicit,
            point meta min=\pointMetaMinVal,
            point meta max=\pointMetaMaxVal,
            colormap/jet,] table[x=xcor,y =ycor, point meta={\thisrow{RCS}}]{figures/Results/Dilation/FR_112/RadarLog_112_Dilation_Detections.dat};
 \nextgroupplot[ width=\Plotheightradar,
        height=\Plotheightradar,
        major grid style={draw=none},
        minor grid style={draw=none},
        tick style={draw=none},
        axis equal,
        ]
    \addplot[
            scatter, 
            only marks,
            scatter, 
            mark=*,
            mark size = \markerSize,
            point meta=explicit,
            point meta min=\pointMetaMinVal,
            point meta max=\pointMetaMaxVal,
            colormap/jet,] table[x=xcor,y =ycor, point meta={\thisrow{RCS}}]{figures/Results/Dilation/FR_141/RadarLog_141_Dilation_Detections.dat}; 
\nextgroupplot[ width=\Plotheightradar,
        height=\Plotheightradar,
        major grid style={draw=none},
        minor grid style={draw=none},
        tick style={draw=none},
        axis equal,
        xmin=-4.5,
        xmax=4.5,
        ymin=2,
        ymax=8.5,
        ]
    \addplot[
            scatter, 
            only marks,
            scatter, 
            mark=*,
            mark size = \markerSize,
            point meta=explicit,
            point meta min=\pointMetaMinVal,
            point meta max=\pointMetaMaxVal,
            colormap/jet,] table[x=xcor,y =ycor, point meta={\thisrow{RCS}}]{figures/Results/Dilation/FR_1889/RadarLog_1889_Dilation_Detections.dat};  
 \nextgroupplot[ width=\Plotheight,
        height=\Plotheight,
        major grid style={draw=none},
        minor grid style={draw=none},
        tick style={draw=none},
        axis line style={draw=none},
        ylabel={\textbf{RGB}},
        xmin=0, xmax=15, ymin=0, ymax=15
        ]
    \addplot graphics[xmin=0, xmax=15, ymin=0, ymax=15]{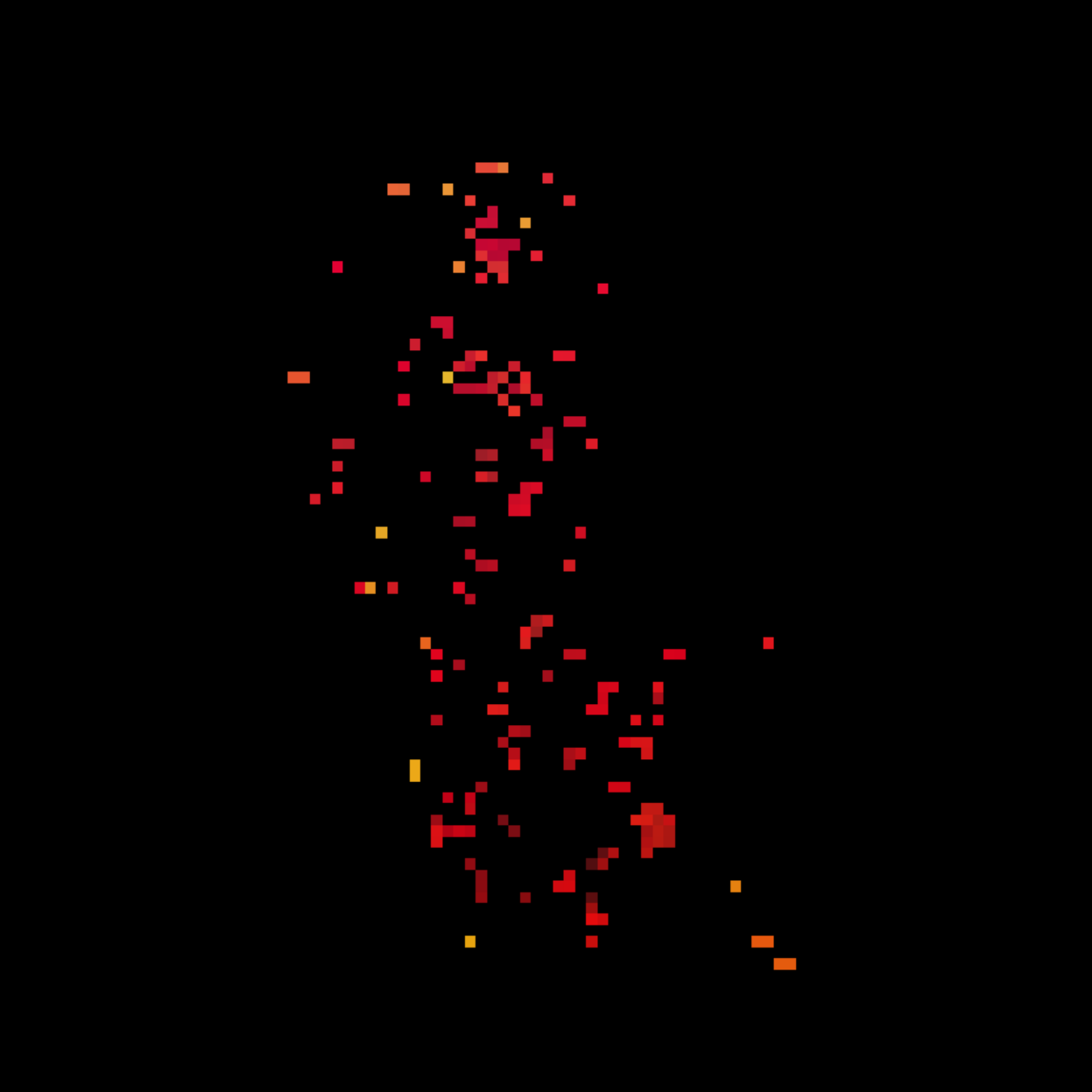};
    
\nextgroupplot[ width=\Plotheight,
        height=\Plotheight,
        major grid style={draw=none},
        minor grid style={draw=none},
        tick style={draw=none},
        axis line style={draw=none},
        xmin=0, xmax=15, ymin=0, ymax=15
        ]
    \addplot graphics[xmin=0, xmax=15, ymin=0, ymax=15]{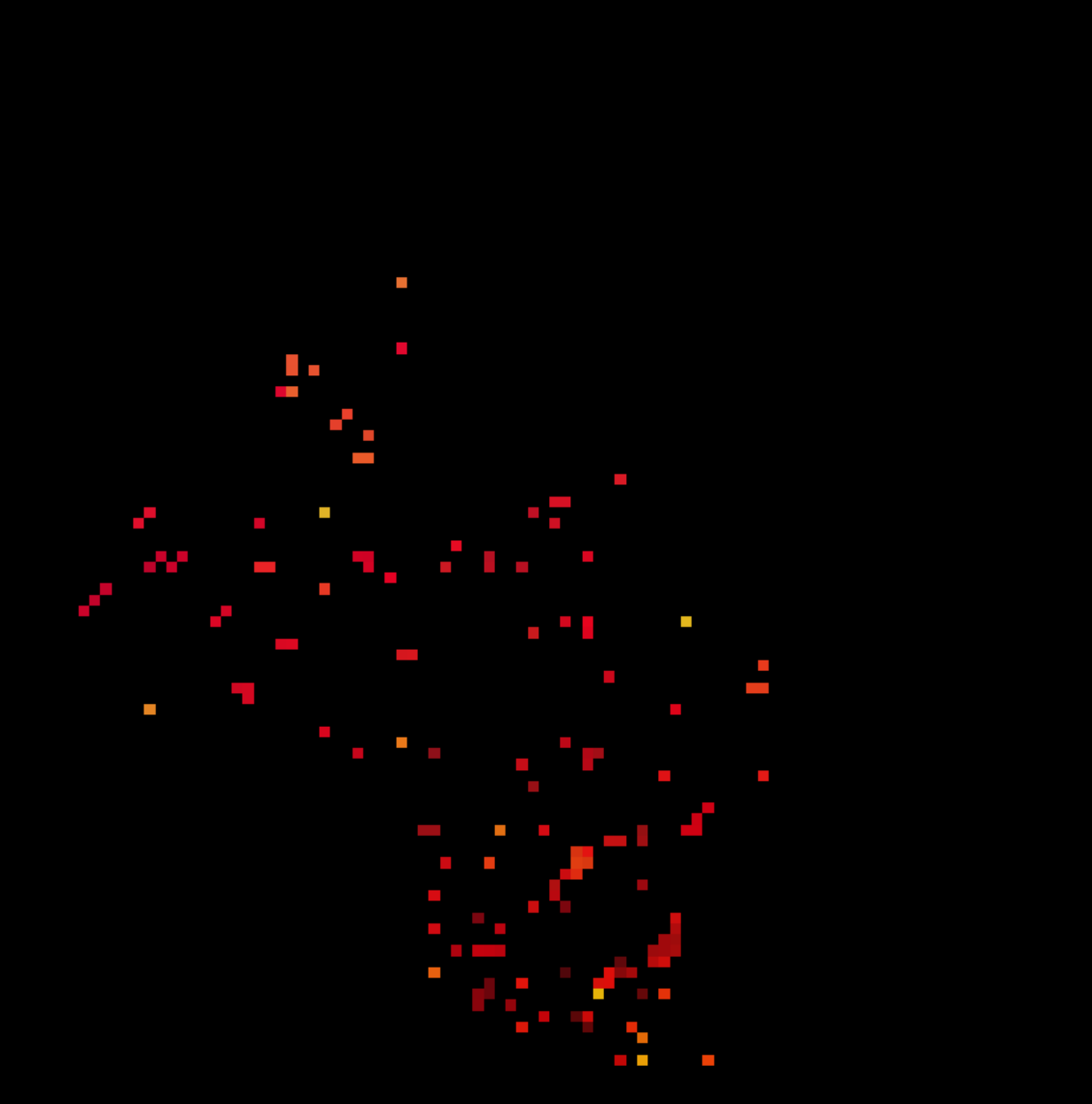};
    
\nextgroupplot[ width=\Plotheight,
        height=\Plotheight,
        major grid style={draw=none},
        minor grid style={draw=none},
        tick style={draw=none},
        axis line style={draw=none},
        xmin=0, xmax=15, ymin=0, ymax=15
        ]
    \addplot graphics[xmin=0, xmax=15, ymin=0, ymax=15]{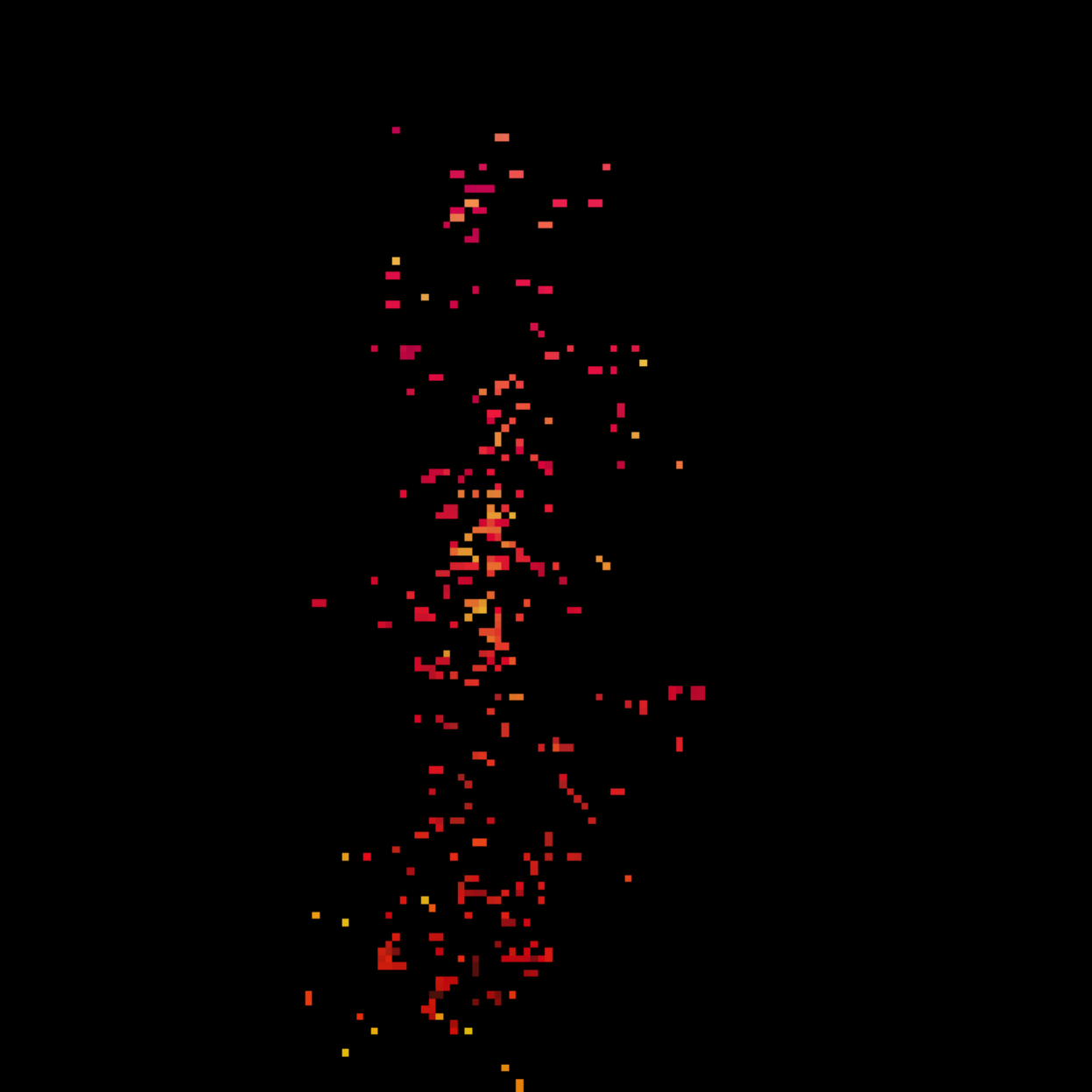};

\nextgroupplot[ width=\Plotheight,
        height=\Plotheight,
        major grid style={draw=none},
        minor grid style={draw=none},
        tick style={draw=none},
        axis line style={draw=none},
        ylabel={\textbf{RGB-DIL}},
        xmin=0, xmax=15, ymin=0, ymax=15
        ]
    \addplot graphics[xmin=0, xmax=15, ymin=0, ymax=15]{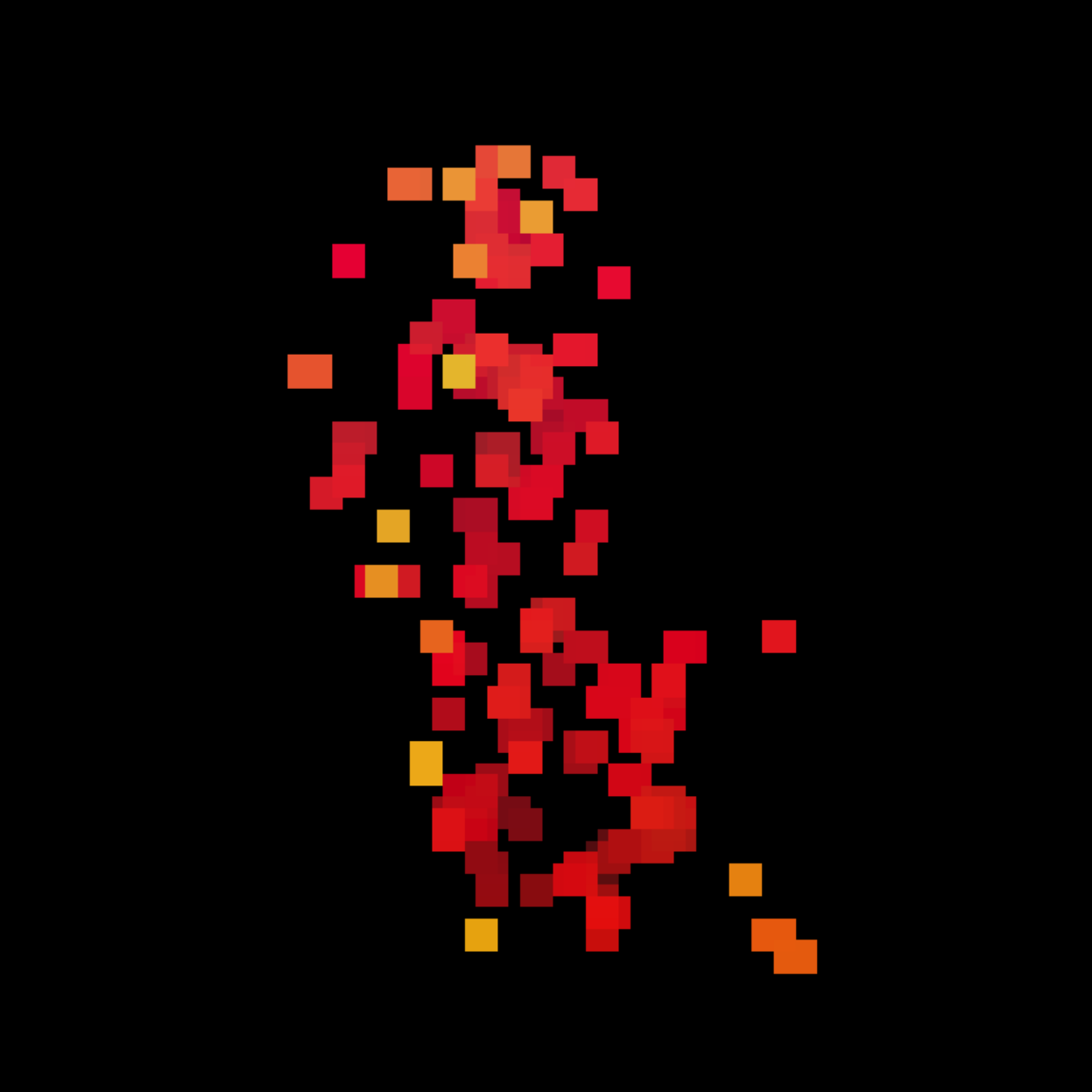};
    
\nextgroupplot[ width=\Plotheight,
        height=\Plotheight,
        major grid style={draw=none},
        minor grid style={draw=none},
        tick style={draw=none},
        axis line style={draw=none},
        xmin=0, xmax=15, ymin=0, ymax=15
        ]
    \addplot graphics[xmin=0, xmax=15, ymin=0, ymax=15]{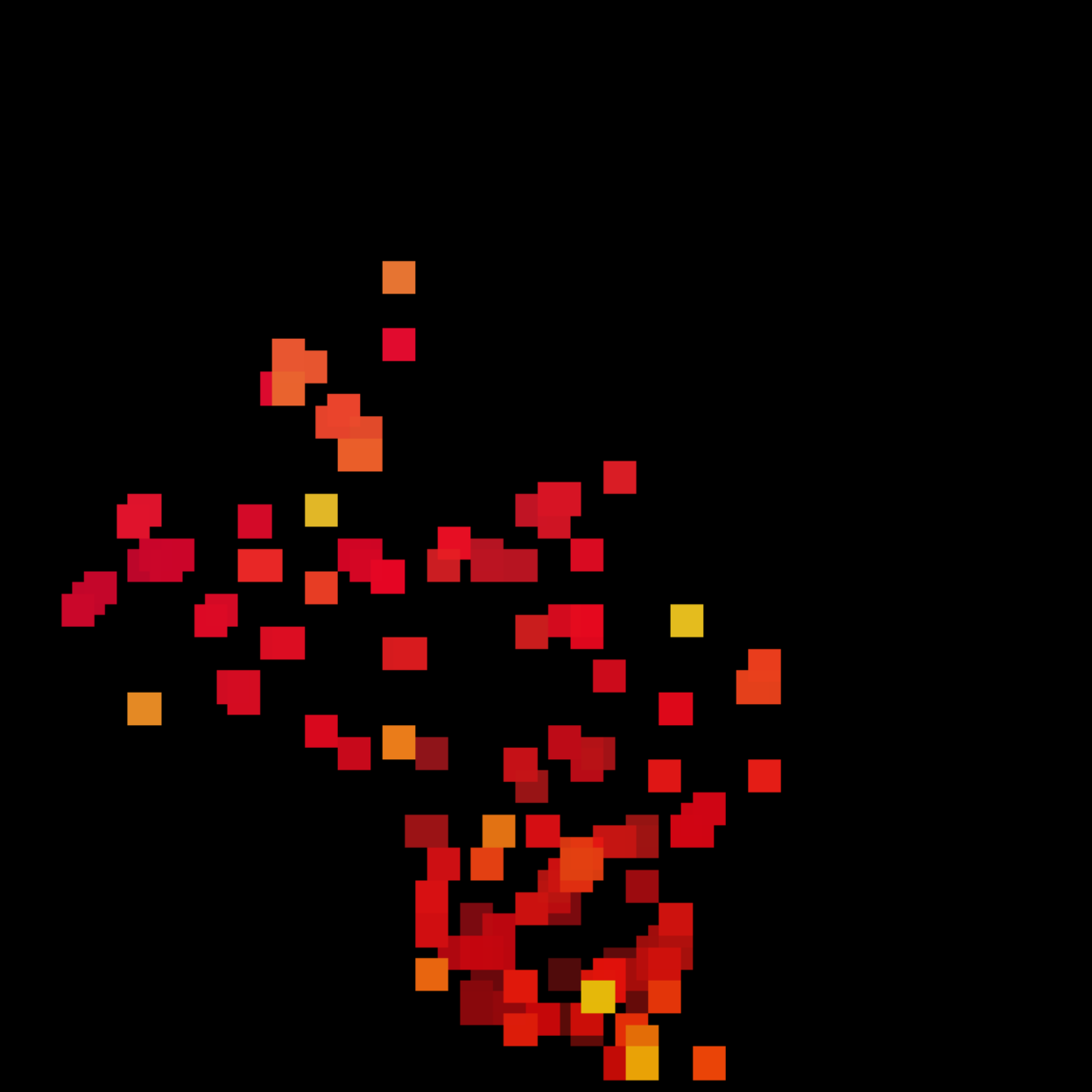};
    
\nextgroupplot[ width=\Plotheight,
        height=\Plotheight,
        major grid style={draw=none},
        minor grid style={draw=none},
        tick style={draw=none},
        axis line style={draw=none},
        xmin=0, xmax=15, ymin=0, ymax=15
        ]
    \addplot graphics[xmin=0, xmax=15, ymin=0, ymax=15]{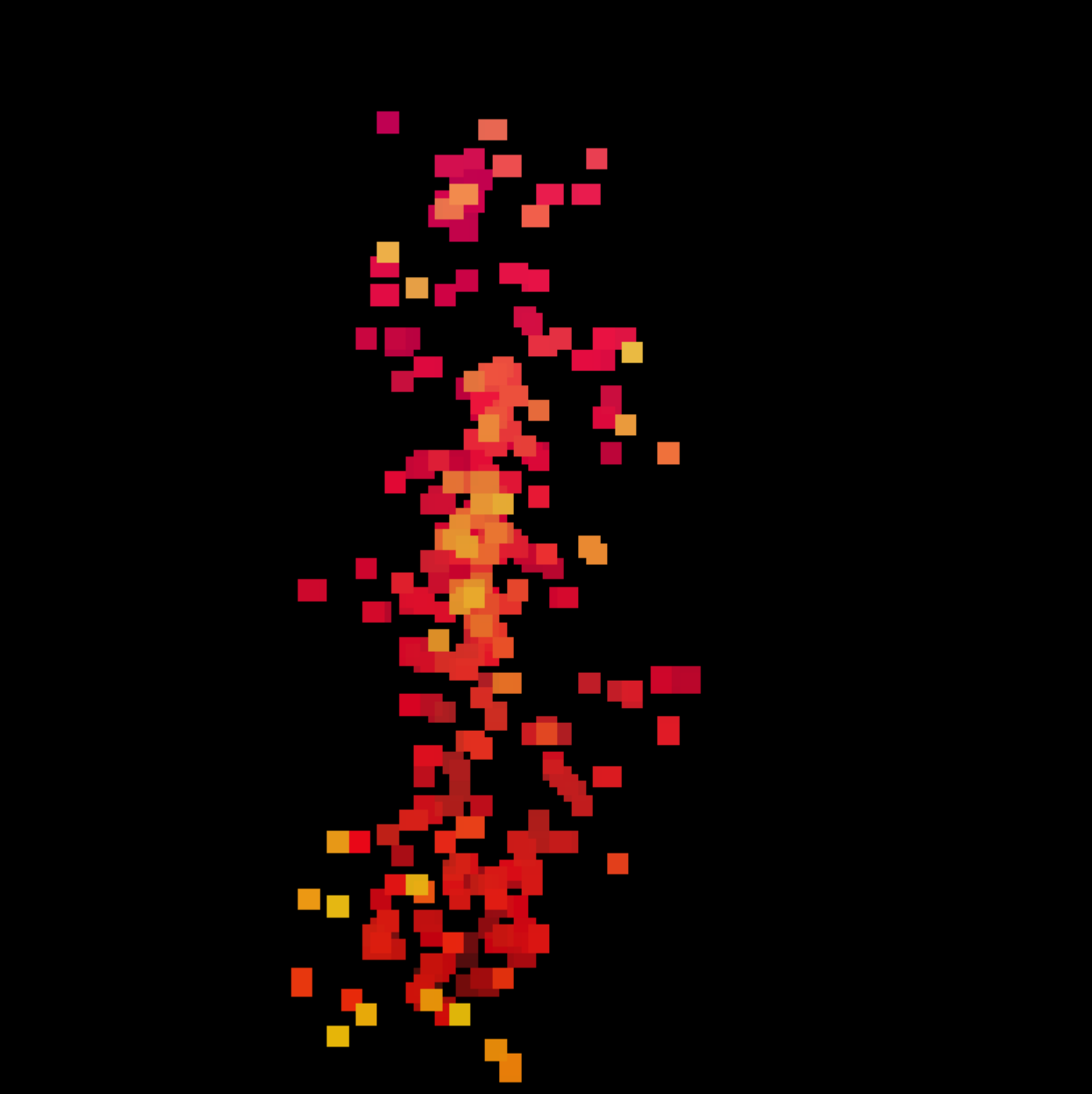};

 \nextgroupplot[width=\Plotheightradar,
        height=\Plotheightradar,
        major grid style={draw=none},
        minor grid style={draw=none},
        tick style={draw=none},
        ylabel={\textbf{Output}},
        axis equal
        ]
    \addplot[
            scatter, 
            only marks,
            scatter, 
            mark=*,
            mark size = \markerSize,
            grid=none,
            point meta=explicit,
            point meta min=\pointMetaMinVal,
            point meta max=\pointMetaMaxVal,
            colormap/jet,] table[x=xcor,y =ycor, point meta={\thisrow{RCS}}]{figures/Results/Dilation/FR_112/RadarLog_112_Dilation_Detections.dat};
    
    \addplot[solid,color=black, thick] table[x=xcor,y expr=\thisrow{ycor}]{figures/Results/Dilation/FR_112/RadarLog_112_Dilation_Pred.dat};
    \addplot[dashed,color=black, thick] table[x=xcor,y expr=\thisrow{ycor}]{figures/Results/Dilation/FR_112/RadarLog_112_Dilation_GT.dat};
 \nextgroupplot[ width=\Plotheightradar,
        height=\Plotheightradar,
        major grid style={draw=none},
        minor grid style={draw=none},
        tick style={draw=none},
        axis equal
        ]
    \addplot[
            scatter, 
            only marks,
            scatter, 
            mark=*,
            mark size = \markerSize,
            point meta=explicit,
            point meta min=\pointMetaMinVal,
            point meta max=\pointMetaMaxVal,
            colormap/jet,] table[x=xcor,y =ycor, point meta={\thisrow{RCS}}]{figures/Results/Dilation/FR_141/RadarLog_141_Dilation_Detections.dat};
    \addplot[solid,color=black, thick] table[x=xcor,y expr=\thisrow{ycor}]{figures/Results/Dilation/FR_141/RadarLog_141_Dilation_Pred.dat};
    \addplot[dashed,color=black, thick] table[x=xcor,y expr=\thisrow{ycor}]{figures/Results/Dilation/FR_141/RadarLog_141_Dilation_GT.dat};        
\nextgroupplot[ width=\Plotheightradar,
        height=\Plotheightradar,
        major grid style={draw=none},
        minor grid style={draw=none},
        tick style={draw=none},
        xmin=-4.5,
        xmax=4.5,
        ymin=2,
        ymax=8.5,
        axis equal,
        ]
    \addplot[
            scatter, 
            only marks,
            scatter, 
            mark=*,
            mark size = \markerSize,
            point meta=explicit,
            point meta min=\pointMetaMinVal,
            point meta max=\pointMetaMaxVal,
            colormap/jet,] table[x=xcor,y =ycor, point meta={\thisrow{RCS}}]{figures/Results/Dilation/FR_1889/RadarLog_1889_Dilation_Detections.dat};  
    \addplot[solid,color=black, thick] table[x=xcor,y expr=\thisrow{ycor}]{figures/Results/Dilation/FR_1889/RadarLog_1889_Dilation_Pred.dat};
    \addplot[dashed,color=black, thick] table[x=xcor,y expr=\thisrow{ycor}]{figures/Results/Dilation/FR_1889/RadarLog_1889_Dilation_GT.dat};   
    \end{groupplot}
\end{tikzpicture}
    \caption{Exemplary radar frames illustrate the impact of the radar-image dilation method for superimposed ghost objects with a higher robustness for near-field object detection. The output includes the native radar point cloud with the predicted object (\textcolor{black}{\rule[0.5ex]{0.5em}{.55pt}}) and ground truth (\textcolor{black}{\rule[0.5ex]{0.5em}{.2pt}\hspace{0.2em}\rule[0.5ex]{0.5em}{.2pt}}).}
    \label{res:Dilation}
\end{figure}
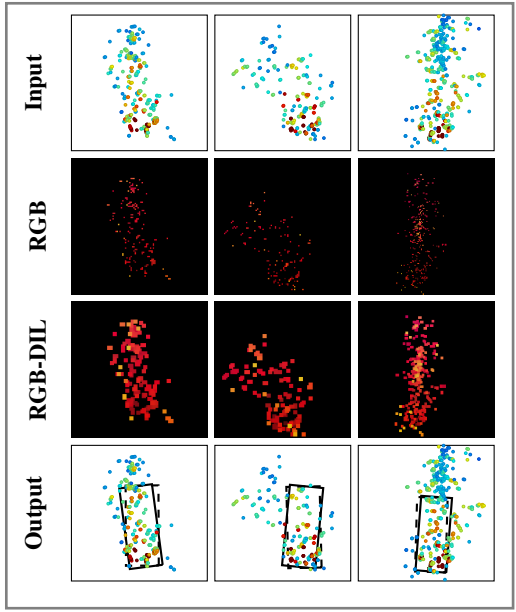

In Figure \ref{res:Dilation}, exemplary input snapshots are presented, displaying intensity-colored radar point clouds of an actual vehicle overlaid with radar multipath reflections. For visualization purposes, radar point clouds are not normalized in the RGB images shown in Figure \ref{res:Dilation}, but normalized before being processed by the \ac{ai} model. However, relying solely on the native RGB image poses a significant challenge for reliable object detection using ComplexYOLO without further pre-processing steps, resulting in increased \ac{fn} rates for pre-crash object detection in the near-field. To address this challenge, a radar-based image dilation method, detailed in Section \ref{subsec:morphological_operations}, is introduced to improve vehicle contour and radar features as depicted in the RGB-DIL image. Figure \ref{res:Dilation} represents the native and the RGB-DIL images. Without the mentioned pre-processing steps, the objects can not be detected by ComplexYOLO. 

Training with radar-image dilation significantly improves the model's object detection performance in near-field scenarios, even when ghost objects and reduced radar features occur due to vehicle-to-vehicle \ac{ol} scenarios. As shown in Figure \ref{res:Dilation}, the ComplexYOLO output layer (continuous black line), after applying dilation, demonstrates the ability to detect objects after applying the radar-based image dilation method. Table \ref{tab:RMSE-Table} highlights the \ac{fn} rates across different distance intervals within vehicle-to-vehicle \ac{ol} scenarios in the RDL-NativeDB, showing a reduction in \ac{fn} occurrences from 29 instances (\SI{2.93}{\percent} of the database) to just six instances (\SI{0.61}{\percent}) when radar-image dilation with a 3x3 \ac{SE} is used. Notably, the method achieves a \SI{12}{\percent} reduction of \ac{fn} at close distances, underscoring the effectiveness of radar-image dilation in reducing detection errors and improving the model's robustness in critical near-field conditions.

\setlength{\dashlinedash}{1pt}
\setlength{\dashlinegap}{2pt}
\begin{table}
    \centering
    \caption{results w/o radar pre-processing and radar-image dilation.}
    \label{tab:RMSE-Table}
    \fontsize{16}{20}\selectfont
    \setlength{\tabcolsep}{2.6pt}
    \resizebox{\columnwidth}{!}{
    \begin{tabular}{|c|ccc|ccc|ccc|ccc|}
        \hline
        \multicolumn{1}{|c|}{\textbf{Scenario}} & \multicolumn{3}{c|}{\textbf{Total}} & \multicolumn{3}{c|}{\textbf{$\mathrm{I_{\text{\fontsize{7}{8}\selectfont \num{1}}}} \geq$ \textbf{\SI{12}{\mathbf{\meter}}}}} & \multicolumn{3}{c|}{\textbf{\textbf{\SI{12}{\mathbf{\meter}}} $> \mathrm{I_{\text{\fontsize{7}{8}\selectfont \num{2}}}} \geq$ \textbf{\SI{5}{\mathbf{\meter}}}}} & \multicolumn{3}{c|}{\textbf{$\mathrm{I_{\text{\fontsize{7}{8}\selectfont \num{3}}}}$ \textless{} \textbf{\SI{5}{\mathbf{\meter}}}}} \\
        
        \hline
        NativeDB-OL & FN &  FN[\si{\percent}] & FR & FN & FN[\si{\percent}] & FR & FN &  FN[\si{\percent}] & FR & FN &  FN[\si{\percent}] & FR \\
        \hline
              
        \multirow{1}{*}{RGB} & \multirow{1}{*}{\num{29}} & \multirow{1}{*}{\num{2.93}} & \multirow{1}{*}{\num{989}} & \multirow{1}{*}{-}& \multirow{1}{*}{-} &\multirow{1}{*}{\num{302}} & \multirow{1}{*}{-} & \multirow{1}{*}{-} & \multirow{1}{*}{\num{479}}& \multirow{1}{*}{\num{29}}& \multirow{1}{*}{\num{13.94}}& \multirow{1}{*}{\num{208}} \\
        \hdashline
        \multirow{1}{*}{RGB-DIL-\num{3}} & \multirow{1}{*}{\num{6}} & \multirow{1}{*}{\num{0.61}} & \multirow{1}{*}{\num{989}} & \multirow{1}{*}{-}& \multirow{1}{*}{-} &\multirow{1}{*}{\num{302}} & \multirow{1}{*}{\num{2}} & \multirow{1}{*}{\num{0.42}} & \multirow{1}{*}{\num{479}}& \multirow{1}{*}{\num{4}}& \multirow{1}{*}{\num{1.92}}& \multirow{1}{*}{\num{208}} \\
        \hline
    \end{tabular}
    }
    \vspace{-5mm}
\end{table}

\subsection{Model Performance on Series Radar Data vs. Tracking Methods} 
\label{subsec:Classic_AI}
Current series radar sensors have a lower native resolution, causing a reduction in the detection distribution along the vehicle's contour. Coupled with their lower velocity resolution compared to research radar sensors, this results in less prominent \ac{md} signatures. Moreover, there is a significant increase in detection fluctuations per radar cycle, which poses a considerable challenge for the model's object detection and regression tasks. Therefore, the Proof of Concept aims to evaluate the model's performance on sparse series radar point clouds and compares it to tracking methods. In order to achieve a more advanced model performance, the individual radar frames from the integrated corner series sensors within the test vehicle are combined to enhance both the sparse radar point cloud and radar features.

\begin{figure}[t!]
  \centering
  \subfigure[Traditional Standard tracking method using Kalman filter for a point target's center of mass.]{\includegraphics[width=0.323\columnwidth]{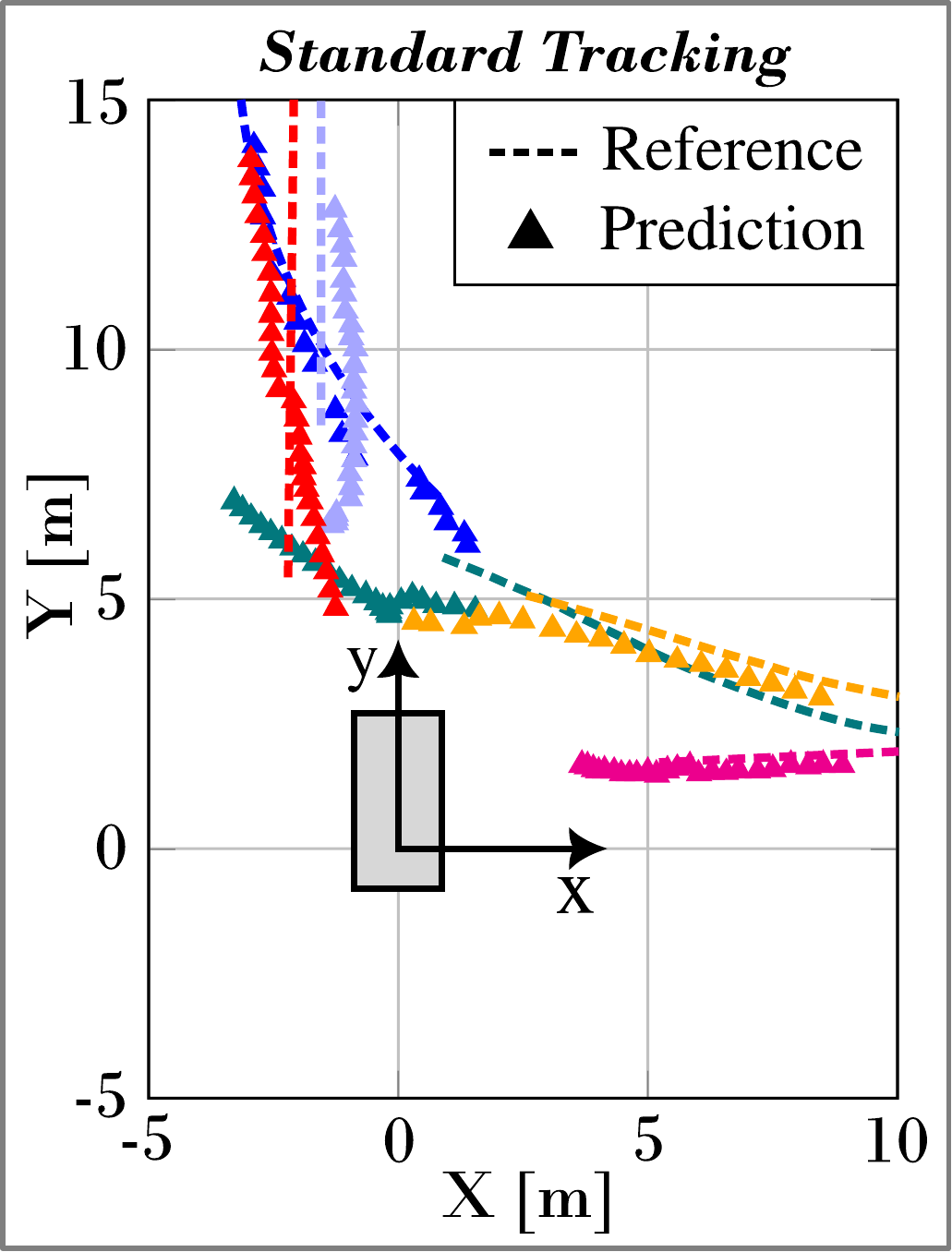}}
  \label{subplot:EM}
  \hfill
  \subfigure[Incorporating radar detections and adaptive motion models in tracking.]{\includegraphics[width=0.323\columnwidth]{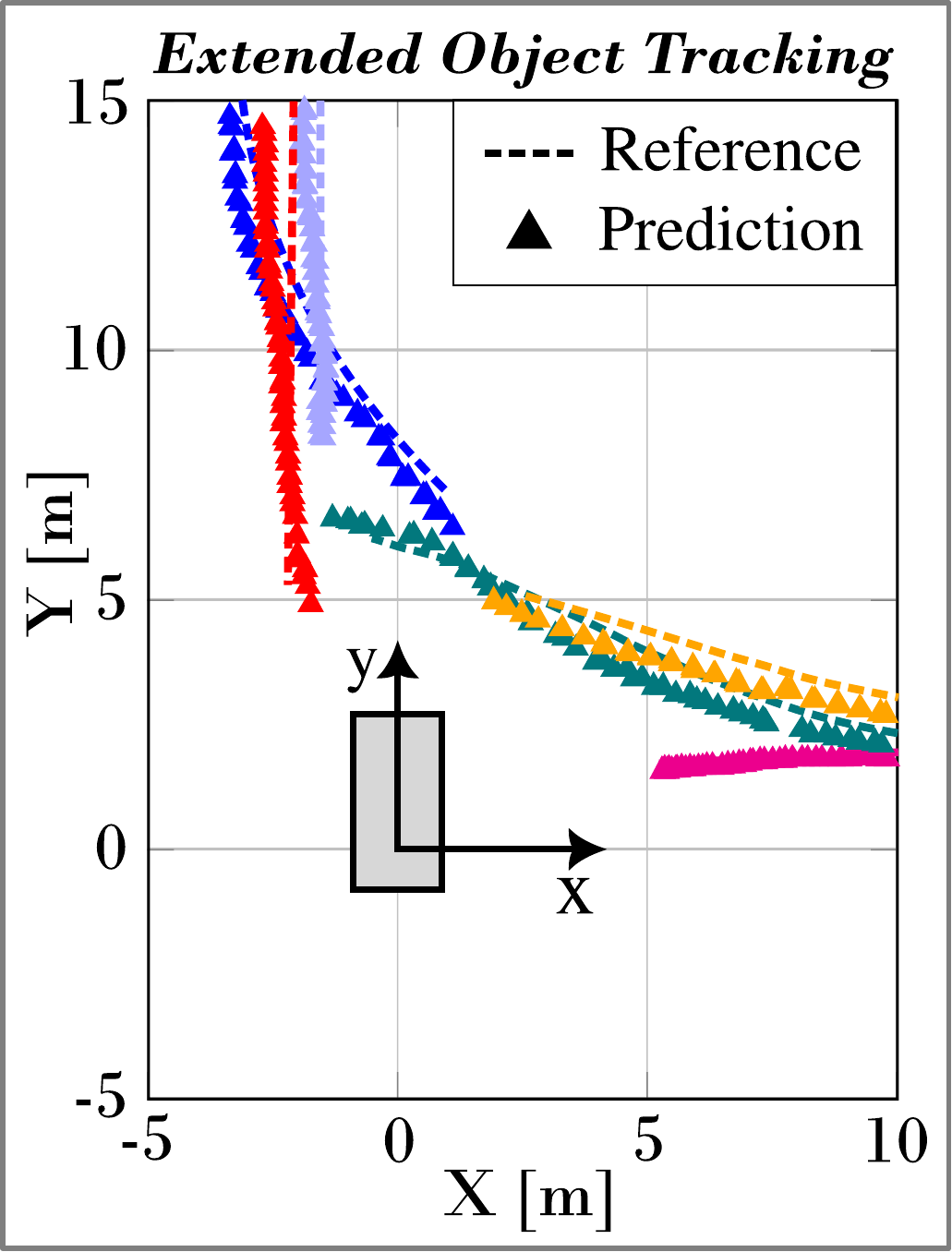}}
  \label{subplot:EOT}
  \hfill
  \subfigure[Single radar frame-based \ac{ai} object state estimation incorporating all radar detections.]{\includegraphics[width=0.323\columnwidth]{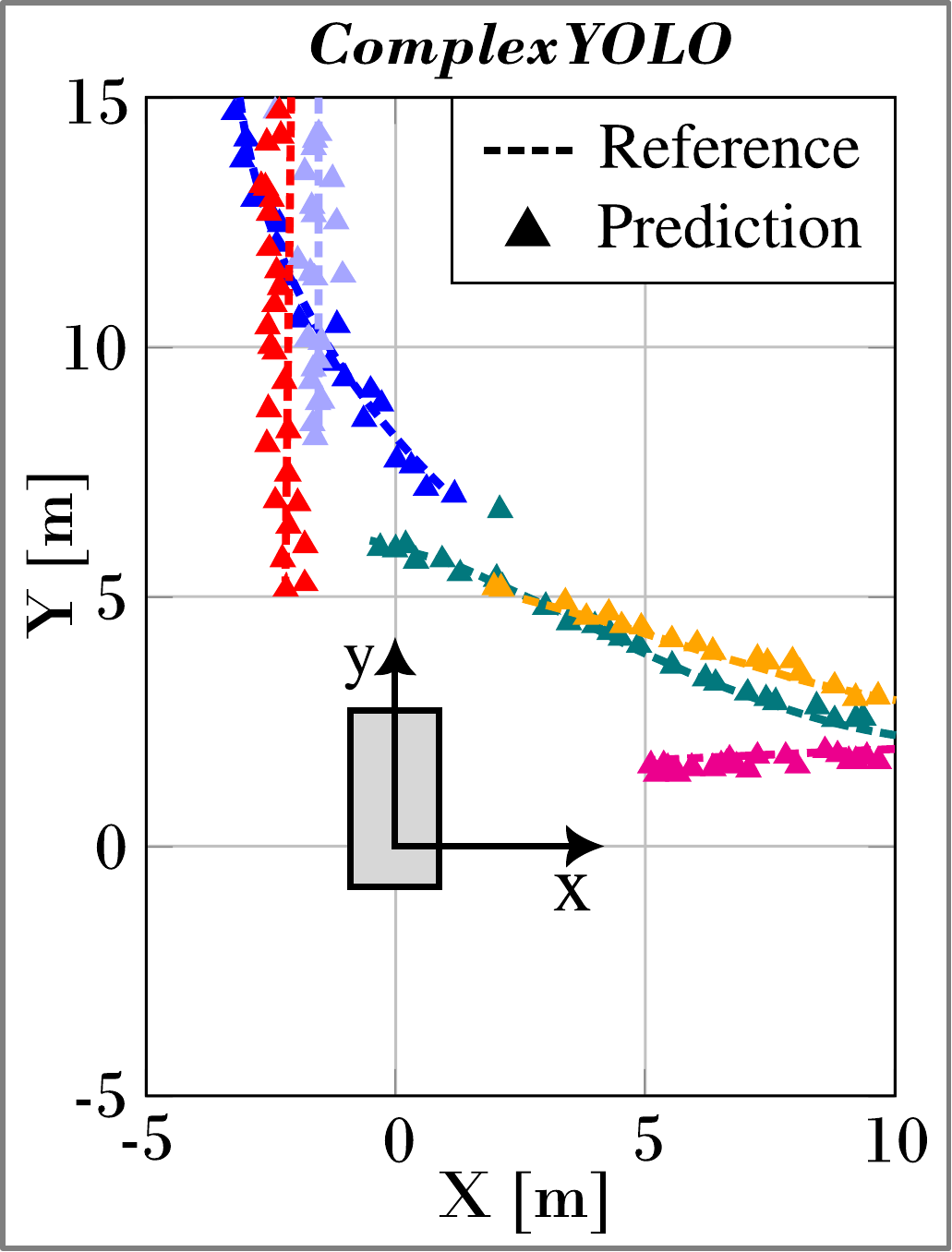}}
  \label{subplot:CY}
  \vspace{-2mm}
  \caption{Exemplary representation of color-coded measurement scenarios showing the reference (\textcolor{black}{\rule[0.5ex]{0.5em}{.55pt}}) and the prediction ($\blacktriangle$) along a comparison between classical tracking methods and the frame-based ComplexYOLO model.}
  \label{fig:evaluationMethods}
  \vspace{-4mm}
\end{figure}

Figure \ref{fig:evaluationMethods} displays color-coded measurement results obtained using tracking methods based on series sensor data cycles, as shown in subplots \ref{fig:evaluationMethods}(a)-(b), contrasted with those produced by ComplexYOLO, depicted in \ref{fig:evaluationMethods}(c). 
The measurements can be categorized into distinct scenarios: linear frontal (highlighted in red and light blue), frontal curve avoidance (depicted in dark blue), linear side (shown in magenta), and side curve avoidance scenarios (represented by green and orange). In the visual representation, the reference is depicted as a dashed line. At the same time, the measurement results are denoted by triangles, symbolizing the predicted center of mass (${x_{\zeta}}$, ${y_{\zeta}}$) for each respective method.
The object's dimensions, such as length and width, are not taken into account in ComplexYOLO's training and testing methodology, which focuses solely on single-object scenarios. As illustrated in Figure \ref{fig:evaluationMethods}, an exemplary measurement demonstrates the detection performance of all three algorithms. The tracking algorithms utilize prior data to improve their predictions, effectively capturing object movement patterns and improving future parameter estimations. As ComplexYOLO operates as a frame-by-frame algorithm, there is no need for initialization or transitional phases, and each radar frame is independently evaluated. This characteristic makes ComplexYOLO highly effective for detecting suddenly occurring objects in the crucial final meters leading up to a collision.

The conventional \ac{em} shown in the subplot \ref{fig:evaluationMethods}(a) exhibits notable deviations in nearly every measurement. These deviations can be attributed to a limitation of the physical measurement model, which only considers the \ac{CoG}. This issue predominantly arises when dealing with unevenly distributed point clouds or when detections are concentrated along a single vehicle's contour edge. In such circumstances, accurately estimating the \ac{CoG} becomes challenging, leading to inaccuracies in determining the object parameters, as depicted in various measurement runs in Fig. \ref{fig:evaluationMethods}(a). 

Advanced research tracking methods like the \ac{IMM-EOT} utilize diverse motion models in conjunction with the \ac{HTG} measurement model, significantly improving the accuracy of estimated object parameters. A notable improvement over the \ac{em} is achieved by integrating the \ac{HTG} measurement model with \ac{EOT}, which considers the distribution and location of radar detections along the vehicle contour to determine kinematic object parameters, rather than relying solely on \ac{CoG}. Across various test runs, the advanced \ac{IMM-EOT} consistently exhibits better accuracy in predicting object parameters than the conventional \ac{em}. Specifically, as depicted in \ref{fig:evaluationMethods}(b), the \ac{IMM-EOT} demonstrates superior performance in linear scenarios (red, light blue, magenta) compared to ComplexYOLO. However, the filter initialization of the \ac{IMM-EOT} in the far-field positively affects the accuracy of object parameter estimation, particularly in close range. The abrupt emergence of objects in the near-field negatively affects the results when the filter initialization isn’t fully executed, as long observation phases, like those in the test scenarios, are rarely available. In addition, maneuvers characterized by high dynamics, such as rapid evasive actions that involve significant braking, present formidable hurdles even for advanced tracking methods.

As shown in Figure \ref{fig:evaluationMethods}(c), the frame-based ComplexYOLO model provides a discontinuous and non-steady progression in the prediction of object parameters, especially in straight-line scenarios with occurrences of \ac{fn}. The curve and avoidance scenarios in dark blue, orange, and green in Figure \ref{fig:evaluationMethods} entail considerably more complex motion patterns. 
Unlike linear scenarios, the radar sensor detects not only the front of the target in these scenarios, but also parts of the vehicle's side, resulting in a moderate L-shape. The partly improved point distribution and the occurrence of \ac{md} signatures enhance the ComplexYOLO model's performance, as Subfigure \ref{fig:evaluationMethods}(c) shows.

Table \ref{tab:Overview_Continental_Measurment} provides a detailed summary of the measurement scenarios shown in Figure \ref{fig:evaluationMethods}. Each scenario is analyzed individually, presenting the corresponding \ac{rmse} for the estimated object parameters, denoted by a subscript $e$. Each measurement's lowest \ac{rmse} is highlighted in bold.
The \ac{IMM-EOT} method performs slightly better than the \ac{em} and ComplexYOLO, particularly in linear vehicle-to-vehicle driving scenarios. However, increasing the density and improving the distribution of point clouds in curving and avoidance scenarios enhance \ac{ai} model performance, yielding more accurate positional estimates than \ac{IMM-EOT}, as indicated by the dark blue scenario. Furthermore, when error outliers for ComplexYOLO are excluded in the green lateral avoidance maneuver, the \ac{rmse} values for $x_{\zeta,\mathrm{e}}$ and $y_{\zeta,\mathrm{e}}$ are comparable to or even lower than those achieved by the \ac{IMM-EOT} method.

The results show significant improvements in estimating kinematic object parameters when using the advanced tracking algorithm \ac{IMM-EOT} and the \ac{ai} model compared to the standard series tracking method. Both research approaches show similar deviations and can be effectively applied to enhance object detection and tracking in automotive safety systems.
However, each approach faces challenges with different driving maneuvers, such as dynamic avoidance and linear scenarios. Combining both methods could help overcome these weaknesses, offering a more robust object detection and tracking solution.
\vspace{-1mm}
\setlength{\dashlinedash}{1pt}
\setlength{\dashlinegap}{2pt}
\newcommand{\sclefactor}{1.1}
\begin{table}[h!]
 \renewcommand{\arraystretch}{1.1}
    \centering
    \caption{analysis of exemplary test scenarios based on tracking methods and complexyolo with series radar data.}
    \label{tab:Overview_Continental_Measurment}
    \fontsize{20}{40}\selectfont
    \setlength{\tabcolsep}{3pt}
    \adjustbox{width=\columnwidth}{\begin{tabular}{|c|cccc|cccc|cccc|}
        \hline
        \multicolumn{1}{|c|}{\textbf{\Huge Algorithm}} & \multicolumn{4}{c|}{\textbf{\Huge Standard Tracking}} & \multicolumn{4}{c|}{\textbf{\Huge Extended Object Tracking}} & \multicolumn{4}{c|}{\textbf{\Huge ComplexYOLO}}\\
        \cline{1-13}
        \raisebox{-0.3ex} {\Huge Parameter} & \scalebox{\sclefactor}{$\bm{x_{\zeta,\mathrm{e}}}$} & \scalebox{\sclefactor}{$\bm{y_{\zeta,\mathrm{e}}}$} & \scalebox{\sclefactor}{$\bm{\psi_\mathrm{e}}$} & \scalebox{\sclefactor}{$\rule[-1.3ex]{0pt}{3ex}{\bm{v_{\mathrm{t,e}}}}$} & \scalebox{\sclefactor}{$\bm{x_{\zeta,\mathrm{e}}}$} & \scalebox{\sclefactor}{$\bm{y_{\zeta,\mathrm{e}}}$} & \scalebox{\sclefactor}{$\bm{\psi_\mathrm{e}}$} & \scalebox{\sclefactor}{$\rule[-1.3ex]{0pt}{3ex}{\bm{v_{\mathrm{_t,e}}}}$} & \scalebox{\sclefactor}{$\bm{x_{\zeta,\mathrm{e}}}$} & \scalebox{\sclefactor}{$\bm{y_{\zeta,\mathrm{e}}}$} & \scalebox{\sclefactor}{$\bm{\psi_\mathrm{e}}$} & \scalebox{\sclefactor}{$\rule[-1.3ex]{0pt}{3ex}{\bm{v_{\mathrm{_t,e}}}}$}\\
        \hdashline  
        \Huge Unit & $\rule[-1.ex]{0pt}{-200ex}$ [\si{\meter}] & [\si{\meter}] & [\si{\degree}] & [\SI[per-mode=fraction]{}{\kilo\meter\per\hour}] & [\si{\meter}] & [\si{\meter}] & [\si{\degree}] & [\SI[per-mode=fraction]{}{\kilo\meter\per\hour}] & [\si{\meter}] & [\si{\meter}] & [\si{\degree}] & [\SI[per-mode=fraction]{}{\kilo\meter\per\hour}]\\
        \hline

         \multirow{1}{*}{\Huge Red} & \hspace{4pt}\num{0.55}~ & \hspace{4pt}\num{1.37}~ & \hspace{4pt}\num{8.46}~ & \hspace{4pt}\num{1.42}~  & \hspace{4pt}\num{0.37}~ & \hspace{4pt}\textbf{\num{0.23}}~ & \hspace{4pt}\textbf{\num{4.45}}~ & \hspace{4pt}\textbf{\num{0.83}}~ & \hspace{4pt}\textbf{\num{0.33}}~ & \hspace{4pt}\num{0.34}~ & \hspace{4pt}\num{4.61}~ & \hspace{4pt}\num{1.44}~ \\
        \cdashline{1-13}
        
        \multirow{1}{*}{\Huge Light Blue} & \hspace{4pt}\num{0.53}~ & \hspace{4pt}\num{2.80}~ & \hspace{4pt}\num{6.48}~ & \hspace{4pt}\num{4.08}~ & \hspace{4pt}\textbf{\num{0.20}}~ & \hspace{4pt}\textbf{\num{0.17}}~ & \hspace{4pt}\textbf{\num{3.34}}~ & \hspace{4pt}\textbf{\num{1.65}}~ &  \hspace{4pt}\num{0.27}~ & \hspace{4pt}\num{0.26}~ & \hspace{4pt}\num{5.60}~ & \hspace{4pt}\num{3.87}~ \\ 
        \cdashline{1-13}
        
        \multirow{1}{*}{\Huge Dark Blue} & \hspace{4pt}\num{0.63}~ & \hspace{4pt}\num{1.44}~ & \hspace{4pt}\textbf{\num{5.06}}~ & \hspace{4pt}\textbf{\num{1.50}}~ & \hspace{4pt}\num{0.37}~ & \hspace{4pt}\num{0.34}~ & \hspace{4pt}\num{5.28}~ & \hspace{4pt}\num{3.05}~ & \hspace{4pt}\textbf{\num{0.23}}~ & \hspace{4pt}\textbf{\num{0.20}}~ & \hspace{4pt}\num{11.07}~ & \hspace{4pt}\num{4.94}~\\ 
        \cdashline{1-13}

        \multirow{1}{*}{\Huge Magenta} & \hspace{4pt}\num{1.38}~ & \hspace{4pt}\num{0.25}~ & \hspace{4pt}\num{3.73}~ & \hspace{4pt}\textbf{\num{1.07}}~ & \hspace{4pt}\num{0.48}~ & \hspace{4pt}\textbf{\num{0.10}}~ & \hspace{4pt}\textbf{\num{2.51}}~ & \hspace{4pt}\num{2.15}~ & \hspace{4pt}\textbf{\num{0.20}}~ & \hspace{4pt}\num{0.17}~ & \hspace{4pt}\num{3.34}~ & \hspace{4pt}\num{4.21}~\\ 
        \cdashline{1-13}

        \multirow{1}{*}{\Huge Orange} & \hspace{4pt}\num{2.05}~ & \hspace{4pt}\num{0.22}~ & \hspace{4pt}\textbf{\num{3.10}}~ & \hspace{4pt}\textbf{\num{0.59}}~ & \hspace{4pt}\num{0.52}~ & \hspace{4pt}\num{0.34}~ & \hspace{4pt}\num{3.81}~ & \hspace{4pt}\num{1.53}~ & \hspace{4pt}\textbf{\num{0.34}}~ & \hspace{4pt}\textbf{\num{0.18}}~ & \hspace{4pt}\num{5.85}~ & \hspace{4pt}\num{4.30}~\\
        \cdashline{1-13}
        
        \multirow{1}{*}{\Huge Green} & \hspace{4pt}\num{6.96}~ & \hspace{4pt}\num{1.73}~ & \hspace{4pt}\num{10.62}~ & \hspace{4pt}\num{9.02}~ & \hspace{4pt}\num{0.67}~ & \hspace{4pt}\textbf{\num{0.33}}~ & \hspace{4pt}\textbf{\num{5.23}}~ & \hspace{4pt}\textbf{\num{2.41}}~ & \hspace{4pt}\textbf{\num{0.22}}~ & \hspace{4pt}\num{0.35}~ & \hspace{4pt}\num{18.34}~ & \hspace{4pt}\num{5.20}~\\
        \hline
    \end{tabular}
    }
    \vspace{-1mm}
\end{table}
\section{Discussion}
\label{Discussion}
Effectively utilizing AI for radar-based pre-crash object detection presents a significant challenge due to the complex distribution of radar point clouds that vary significantly in density across radar frames, particularly in dynamic near-field scenarios. Minor changes in distance, especially approaching vehicles, can cause a significant phase shift in radar signal processing, leading to fluctuations in detections. These fluctuations manifest as detections that appear and disappear based on radar features such as \ac{RCS} variations or Doppler values.
An innovative approach focuses on capturing unique radar characteristics by leveraging high-resolution \ac{md} signatures generated by the vehicle's rotating wheels. Incorporating these signatures and additional wheel detections significantly improves object detection and regression accuracy, particularly in critical near-field scenarios using ComplexYOLO. Analysis reveals that \ac{md} signatures can be reliably detected up to approximately 14 meters using high-resolution short-range radar, making this approach well suited for pre-crash applications.

Systematically extracting \ac{md} signatures from the native high-resolution radar database allows for a comprehensive analysis of how radar features impact the model's performance. Detailed analyses underscored the significance of high-resolution \ac{md} signatures in improving the model's reliability and accuracy. Particularly in critical pre-crash overlap scenarios, where radar detection density and point cloud distribution are significantly reduced, the importance of high-resolution features for a more accurate estimation of the object's kinematic parameters is emphasized.
In addition, utilizing \ac{md} signatures can significantly reduce the number of \ac{fn} due to radar multipath reflections and superimposed radar point clouds from ghost objects in pre-crash object detection. However, it is essential to note that they may not eliminate the occurrence of all \ac{fn}. Furthermore, this article focuses on the analysis of \ac{md} signatures and its impact on the performance of the model. Integrating the innovative radar-image dilation method that operates on feature input channels enables the identification and augmentation of high-resolution data, such as the \ac{md} signatures on radar detection level. The anchor-based AI model achieves improved performance in object detection by following this principle, even in situations with superimposed ghost targets and multipath reflection in pre-crash situations. Additionally, modifying the shape and size of the \ac{SE} has the potential to further improvements.

The evaluation of ComplexYOLO's performance includes processing data from series radar sensors and comparing the results with advanced tracking methods, showing its capability to extend beyond high-resolution radar data. However, a limitation of series radar data lies in the inherently lower resolution, leading to a notable decrease in the detection density. Consequently, there is a less distinct distribution of detection points along the vehicle's contour, significantly diminishing the prominence of \ac{md} signatures per radar cycle. 
The analyzes validate that ComplexYOLO has deviations similar to the tracking methods, yet it offers the advantage of frame-based prediction.
The initialization of the series and research tracking algorithms takes place in the far-field, leading to long observation times and improving object prediction accuracy over time. In real-world scenarios, relevant collision objects are not always observable in the far-field due to, e.g. occlusion. However, they can suddenly emerge in the critical near-field, resulting in limited reaction times. However, a thorough initialization of the tracking algorithms requires several sensor cycles, which might not be possible in the last few meters before the collision. Especially in these situations, ComplexYOLO provides the advantage of predicting the kinematic object parameters based on a single sensor measurement cycle and offers the possibility of activating vehicle safety systems. The current limitations of series radar sensors influence the performance of ComplexYOLO. However, the results of high-resolution radars indicate that significant performance improvements are possible, particularly with improved \ac{md} signatures.
\section{Conclusion}
\label{Conclusion}
This paper presents an innovative anchor-based \ac{ai} model for radar short-range vehicle pre-crash object detection and regression. It uses radar point clouds along with supplementary unique high-resolution \ac{md} signatures evaluated from radar raw data.
Comprehensive analyses have investigated the impact of \ac{md} signatures on ComplexYOLO's model performance within specific test conditions and varying distance intervals. Although \ac{md} velocities exhibit a mediocre influence on the accuracy of kinematic object parameters, they notably affect \ac{fn}, particularly in crucial near-field \ac{ol} scenarios. Including all wheel-related \ac{md} signatures alters the estimation of kinematic object parameters and amplifies \ac{fn} across the entire distance spectrum, with a pronounced effect in near-field scenarios.
The model's reliance on radar features, particularly in critical pre-crash scenarios, underscores its capacity to learn, generalize, and effectively utilize high-resolution radar data features. This proficiency is evident in the accuracy of object parameters and the prevalence of \ac{fn}s at varying distances. However, in near-field situations, despite advanced radar sensor technologies and well-defined radar features, there are instances where the \ac{ai} model fails to identify approaching vehicles consistently. Addressing this issue, an innovative integrated radar-image dilation technique on the feature input channels substantially enhances the model's efficacy. It diminishes \ac{fn}s, bolstering overall model robustness and enhancing system dependability, remarkably in pre-crash scenarios, where overlapping point clouds from ghost objects and multipath radar reflections occur.

Furthermore, a series radar database has been established utilizing automotive series sensors, demonstrating the effectiveness of this concept not only on high-resolution research radar data, but also on sensors with drastically lower detection density, less pronounced point cloud distribution, and less prominent \ac{md} signatures.
The model performs better than series tracking algorithms, particularly in critical avoidance situations. In scenarios with a decrease in point cloud resolution or reduced radial velocity information at close ranges, the \ac{ai} model may struggle to identify objects accurately.

Additional robustness tests with different vehicle types are necessary for further development. In addition, exploring potential application areas and investigating combinations with traditional tracking methods are promising tasks to pursue.
\bibliographystyle{bibliography/IEEEtran}
\bibliography{bibliography/Reference}        

@online{Reference:WHO2023,
    author  = {{World Health Organization}},
    title   = {Global status report on road safety 2023},
    year    = {2023},
    url     = {https://www.who.int/teams/social-determinants-of-health/safety-and-mobility/global-status-report-on-road-safety-2023},
    note    = {Accessed: Dez. 15, 2023},
}

@online{Reference:EuroNCAP,
    author  = {Euro NCAP},
    title   = {Vision 2030: a Safer Future for Mobility},
    year    = {2022},
    url     = {https://cdn.euroncap.com/media/74468/euro-ncap-roadmap-vision-2030.pdf},
    note    = {Accessed: March. 29, 2024},
}

@article{Innenraumkonzepte-La,
    author  = {Laakmann, Frank and Zink, Lothar and Seyffert, Martin},
    year    = {2019},
    month   = {04},
    pages   = {54-59},
    title   = {Neue {Innenraumkonzepte} für den {Insassenschutz} in hochautomatisierten {Fahrzeugen}},
    volume  = {121},
    journal = {ATZ - Automobiltechnische Zeitschrift},
    doi     = {10.1007/s35148-019-0017-z}
}

@article{Innenraum3,
    author  = {K. Golowko and P. Mugele and D. Zimmer},
    title   = {Neue {Möglichkeiten} der {Innenraumgestaltung}},
    journal = {ATZ Extra 22 (Suppl 3)},
    year    = {2017},
    pages   = {42-45},
    doi     = {10.1007/s35778-017-0030-3}
}

@article{Insassenbewegung-Ol,
    author  = {Olders, Stefan and Weinkopf, Andreas},
    year    = {2019},
    month   = {01},
    pages   = {26-31},
    title   = {Insassenbewegung in {Pre}-{Crash}-{Manövern}},
    volume  = {121},
    journal = {ATZ - Automobiltechnische Zeitschrift},
    doi     = {10.1007/s35148-018-0191-4}
}

@article{Ganzheitliche-Ko,
    author  = {Kompass, Klaus and Helmer, Thomas and Blaschke, Christoph and Kates, Ronald},
    year    = {2014},
    month   = {10},
    pages   = {10–15},
    title   = {Ganzheitliche und integrale Fahrzeugsicherheit},
    volume  = {116},
    journal = {ATZ - Automobiltechnische Zeitschrift},
    doi     = {10.1007/s35148-014-0489-9}
}

@ARTICLE{Engels_Advances_in_Automotive_Radar_A_framework_on_computationally_efficient_high_resolution_frequency_estimation,
    author  = {Engels, Florian and Heidenreich, Philipp and Zoubir, Abdelhak M. and Jondral, Friedrich K and Wintermantel, Markus},
    journal = {IEEE Signal Processing Magazine}, 
    title   = {Advances in Automotive Radar: A framework on computationally efficient high-resolution frequency estimation}, 
    year    = {2017},
    volume  = {34},
    number  = {2},
    pages   = {36-46},
    doi     = {10.1109/MSP.2016.2637700}
}

@ARTICLE{microdoppler-Ch,
    author  = {Chen, V.C. and Li, F. and Ho, S.-S. and Wechsler, H.},
    journal = {IEEE Transactions on Aerospace and Electronic Systems}, 
    title   = {Micro-Doppler effect in radar: phenomenon, model, and simulation study}, 
    year    = {2006},
    volume  = {42},
    number  = {1},
    pages   = {2-21},
    doi     = {10.1109/TAES.2006.1603402}
}

@ARTICLE{Kamann_Extended_Object_Tracking_Using_Spatially_Resolved_Micro-Doppler_Signatures,
    author  = {Kamann, Alexander and Steinhauser, Dagmar and Gruson, Frank and Brandmeier, Thomas and Schwarz, Ulrich T.},
    journal = {IEEE Transactions on Intelligent Vehicles}, 
    title   = {Extended Object Tracking Using Spatially Resolved Micro-Doppler Signatures}, 
    year    = {2021},
    volume  = {6},
    number  = {3},
    pages   = {440-449},
    doi     = {10.1109/TIV.2020.3035433}
}

@ARTICLE{Held_A_Novel_Approach_for_Model_Based_Pedestrian_Tracking_Using_Automotive_Radar,
    author  = {Held, Patrick and Steinhauser, Dagmar and Koch, Andreas and Brandmeier, Thomas and Schwarz, Ulrich T.},
    journal = {IEEE Transactions on Intelligent Transportation Systems}, 
    title   = {A Novel Approach for Model-Based Pedestrian Tracking Using Automotive Radar}, 
    year    = {2022},
    volume  = {23},
    number  = {7},
    pages   = {7082-7095},
    doi     = {10.1109/TITS.2021.3066680}
}

@INPROCEEDINGS{geiger_kitti,
    author  = {Geiger, Andreas and Lenz, Philip and Urtasun, Raquel},
    booktitle={2012 IEEE Conference on Computer Vision and Pattern Recognition}, 
    title   = {Are we ready for autonomous driving? The KITTI vision benchmark suite}, 
    year    = {2012},
    volume  = {},
    number  = {},
    pages   = {3354-3361},
    doi     = {10.1109/CVPR.2012.6248074}
}

@INPROCEEDINGS{Caesar_nuScenes,
    author  = {Caesar, Holger and Bankiti, Varun and Lang, Alex H. and Vora, Sourabh and Liong, Venice Erin and Xu, Qiang and Krishnan, Anush and Pan, Yu and Baldan, Giancarlo and Beijbom, Oscar},
    booktitle= {2020 IEEE/CVF Conference on Computer Vision and Pattern Recognition (CVPR)}, 
    title   = {nuScenes: A Multimodal Dataset for Autonomous Driving}, 
    year    = {2020},
    volume  = {},
    number  = {},
    pages   = {11618-11628},
    doi     = {10.1109/CVPR42600.2020.01164}
}

@Article{Reference:Nobis_Kernel_Point_Convolution_LSTM_Networks_for_Radar_Point_Cloud_Segmentation,
    AUTHOR  = {Nobis, Felix and Fent, Felix and Betz, Johannes and Lienkamp, Markus},
    TITLE   = {Kernel Point Convolution LSTM Networks for Radar Point Cloud Segmentation},
    JOURNAL = {Applied Sciences},
    VOLUME  = {11},
    YEAR    = {2021},
    NUMBER  = {6},
    ARTICLE-NUMBER = {2599},
    URL     = {https://www.mdpi.com/2076-3417/11/6/2599},
    ISSN    = {2076-3417},
    DOI     = {10.3390/app11062599}
}

@INPROCEEDINGS{Reference:Scheiner_Object_detection_for_automotive_radar_point_clouds_a_comparison,
    author  = {Scheiner, Nicolas and Kraus, Florian and Appenrodt, Nils and Dickmann, Jürgen and Sick,Bernhard},
    booktitle={AI Perspectives}, 
    title   = {Object detection for automotive radar point clouds – a comparison},
    year    = {2021},
    volume  = {3},
    number  = {},
    pages   = {6},
    doi     = {https://doi.org/10.1186/s42467-021-00012-z}
}

@INPROCEEDINGS{early_fusion,
    author  = {Vriesman, Daniel and Junior, Alceu Britto and Zimmer, Alessandro and Brandmeier, Thomas},
    booktitle={2023 26th International Conference on Information Fusion (FUSION)}, 
    title   = {Multimodal Early Fusion of Automotive Sensors based on Autoencoder Network: An anchor-free approach for Vehicle 3D Detection}, 
    year    = {2023},
    volume  = {},
    number  = {},
    pages   = {1-8},
    doi     = {10.23919/FUSION52260.2023.10224140}
}

@INPROCEEDINGS{Reference:Patel_Deep_Learning_based_Object_Classification_on_Automotive_Radar_Spectra,
    author  ={Patel, Kanil and Rambach, Kilian and Visentin, Tristan and Rusev, Daniel and Pfeiffer, Michael and Yang, Bin},
    booktitle={2019 IEEE Radar Conference (RadarConf)}, 
    title   = {Deep Learning-based Object Classification on Automotive Radar Spectra}, 
    year    = {2019},
    volume  = {},
    number  = {},
    pages   = {1-6},
    doi     = {10.1109/RADAR.2019.8835775}
}

@INPROCEEDINGS{Reference:Lombacher_Semantic_radar_grids,
    author  = {Lombacher, Jakob and Laudt, Kilian and Hahn, Markus and Dickmann, Jürgen and Wöhler, Christian},
    booktitle={2017 IEEE Intelligent Vehicles Symposium (IV)}, 
    title   = {Semantic radar grids}, 
    year    = {2017},
    volume  = {},
    number  = {},
    pages   = {1170-1175},
    doi     = {10.1109/IVS.2017.7995871}
}

@INPROCEEDINGS{Reference:Roos_Estimation_of_the_orientation_of_vehicles_in_high_resolution_radar_images,
    author  = {Roos, Fabian and Kellner, Dominik and Klappstein, Jens and Dickmann, Juergen and Dietmayer, Klaus and Muller-Glaser, Klaus D. and Waldschmidt, Christian},
    booktitle={2015 IEEE MTT-S International Conference on Microwaves for Intelligent Mobility (ICMIM)}, 
    title   = {Estimation of the orientation of vehicles in high-resolution radar images}, 
    year    = {2015},
    volume  = {},
    number  = {},
    pages   = {1-4},
    doi     = {10.1109/ICMIM.2015.7117949}
}

@INPROCEEDINGS{Reference:Schumann_Comparison_of_random_forest_and_long_short_term_memory_network_performances_in_classification_tasks_using_radar,
    author  = {Schumann, Ole and Wöhler, Christian and Hahn, Markus and Dickmann, Jürgen},
    booktitle={2017 Sensor Data Fusion: Trends, Solutions, Applications (SDF)}, 
    title   = {Comparison of random forest and long short-term memory network performances in classification tasks using radar}, 
    year    = {2017},
    volume  = {},
    number  = {},
    pages   = {1-6},
    doi     = {10.1109/SDF.2017.8126350}
}

@INPROCEEDINGS{Reference:Ulrich_DeepReflecs_Deep_Learning_for_Automotive_Object_Classification_with_Radar_Reflections,
    author  = {Ulrich, Michael and Gläser, Claudius and Timm, Fabian},
    booktitle={2021 IEEE Radar Conference (RadarConf21)}, 
    title   = {DeepReflecs: Deep Learning for Automotive Object Classification with Radar Reflections}, 
    year    = {2021},
    volume  = {},
    number  = {},
    pages   = {1-6},
    doi     = {10.1109/RadarConf2147009.2021.9455334}
}

@INPROCEEDINGS{Reference:Schumann_Supervised_Clustering_for_Radar_Applications_On_the_Way_to_Radar_Instance_Segmentation,  
    author  = {Schumann, Ole and Hahn, Markus and Dickmann, Jürgen and Wöhler, Christian},
    booktitle={2018 IEEE MTT-S International Conference on Microwaves for Intelligent Mobility (ICMIM)}, 
    title   = {Supervised Clustering for Radar Applications: On the Way to Radar Instance Segmentation}, 
    year    = {2018},
    volume  = {},
    number  = {},
    pages   = {1-4},
    doi     = {10.1109/ICMIM.2018.8443534}
}

@INPROCEEDINGS{Reference:Scheiner_Off-the-shelf_sensor_vs_experimental_radar_How_much_resolution_is_necessary_in_automotive_radar_classification?,  
    author  = {Scheiner, Nicolas and Schumann, Ole and Kraus, Florian and  Appenrodt, Nils and Dickmann, Jürgen and Sick, Bernhard},
    booktitle={2020 IEEE 23rd International Conference on Information Fusion (FUSION)}, 
    title   = {Off-the-shelf sensor vs. experimental radar – How much resolution is necessary in automotive radar classification?}, 
    year    = {2020},
    volume  = {},
    number  = {},
    pages   = {1-8},
    doi     = {10.23919/fusion45008.2020.9190338}
}

@INPROCEEDINGS{Reference:Danzer_2D_Car_Detection_in_Radar_Data_with_PointNets,
    author  = {Danzer, Andreas and Griebel, Thomas and Bach, Martin and Dietmayer, Klaus},
    booktitle={2019 IEEE Intelligent Transportation Systems Conference (ITSC)},
    title   = {2D Car Detection in Radar Data with PointNets},
    year    = {2019},
    volume  = {},
    number  = {},
    pages   = {61-66},
    doi     = {10.1109/ITSC.2019.8917000}
}

@INPROCEEDINGS{Reference:Qi_Frustum_PointNets_for_3D_Object_Detection_from_RGB-D_Data,
    author  = {Qi, Charles R. and Liu, Wei and Wu, Chenxia and Su, Hao and Guibas, Leonidas J.},
    booktitle={2018 IEEE/CVF Conference on Computer Vision and Pattern Recognition}, 
    title   = {Frustum PointNets for 3D Object Detection from RGB-D Data}, 
    year    = {2018},
    volume  = {},
    number  = {},
    pages   = {918-927},
    doi     = {10.1109/CVPR.2018.00102}
}

@ARTICLE{Reference:Palffy_CNN_Based_Road_User_Detection_Using_the_3D_Radar_Cube,
    author  = {Palffy, Andras and Dong, Jiaao and Kooij, Julian F. P. and Gavrila, Dariu M.},
    journal = {IEEE Robotics and Automation Letters}, 
    title   = {CNN Based Road User Detection Using the 3D Radar Cube}, 
    year    = {2020},
    volume  = {5},
    number  = {2},
    pages   = {1263-1270},
    doi     = {10.1109/LRA.2020.2967272}
}

@INPROCEEDINGS{Reference:Meyer_Graph_Convolutional_Networks_for_3D_Object_Detection_on_Radar_Data,
    author  = {Meyer, Michael and Kuschk, Georg and Tomforde, Sven},
    booktitle={2021 IEEE/CVF International Conference on Computer Vision Workshops (ICCVW)}, 
    title   = {Graph Convolutional Networks for 3D Object Detection on Radar Data}, 
    year    = {2021},
    volume  = {},
    number  = {},
    pages   = {3053-3062},
    doi     = {10.1109/ICCVW54120.2021.00340}
}

@INPROCEEDINGS{Reference:Wang_RODNet_Radar_Object_Detection_using_Cross_Modal_Supervision,
    author  = {Wang, Yizhou and Jiang, Zhongyu and Gao, Xiangyu and Hwang, Jenq-Neng and Xing, Guanbin and Liu, Hui},
    booktitle={2021 IEEE Winter Conference on Applications of Computer Vision (WACV)}, 
    title   = {RODNet: Radar Object Detection using Cross-Modal Supervision}, 
    year    = {2021},
    volume  = {},
    number  = {},
    pages   = {504-513},
    doi     = {10.1109/WACV48630.2021.00055}
}

@INPROCEEDINGS{Reference:Brodeski_Deep_Radar_Detector,
    author  = {Brodeski, Daniel and Bilik, Igal and Giryes, Raja},
    booktitle={2019 IEEE Radar Conference (RadarConf)}, 
    title   = {Deep Radar Detector}, 
    year    = {2019},
    volume  = {},
    number  = {},
    pages   = {1-6},
    doi     = {10.1109/RADAR.2019.8835792}
}

@INPROCEEDINGS{Reference:AI_Object_Detection,
    author  = {Dreher, Maria and Erçelik, Emeç and Bänziger, Timo and Knol, Alois},
    booktitle={2020 IEEE 23rd International Conference on Intelligent Transportation Systems (ITSC)}, 
    title   = {Radar-based 2D Car Detection Using Deep Neural Networks}, 
    year    = {2020},
    volume  = {},
    number  = {},
    pages   = {1-8},
    doi     = {10.1109/ITSC45102.2020.9294546}
}

@INPROCEEDINGS{Reference:Meyer_Deep_Learning_Based_3D_Object_Detection_for_Automotive_Radar_and_Camera,
    author  = {Meyer, Michael and Kuschk, Georg},
    booktitle={2019 16th European Radar Conference (EuRAD)}, 
    title   = {Deep Learning Based 3D Object Detection for Automotive Radar and Camera},
    year    = {2019},
    volume  = {},
    number  = {},
    pages   = {133-136},
    doi     = {}
}

@article{Reference:Lee_Deep_Learning_on_Radar_Centric_3D_Object_Detection,
    author  = {Lee, Seungjun},
    title   = {Deep Learning on Radar Centric 3D Object Detection},
    journal = {CoRR},
    volume  = {abs/2003.00851},
    year    = {2020},
    url     = {https://arxiv.org/abs/2003.00851},
    eprinttype= {arXiv},
    eprint  = {2003.00851},
    bibsource= {dblp computer science bibliography, https://dblp.org}
}

@Article{Referemce:Nobis_Radar_Voxel_Fusion_for_3D_Object_Detection,
    AUTHOR  = {Nobis, Felix and Shafiei, Ehsan and Karle, Phillip and Betz, Johannes and Lienkamp, Markus},
    TITLE   = {Radar Voxel Fusion for 3D Object Detection},
    JOURNAL = {Applied Sciences},
    VOLUME  = {11},
    YEAR    = {2021},
    NUMBER  = {12},
    ARTICLE-NUMBER = {5598},
    URL     = {https://www.mdpi.com/2076-3417/11/12/5598},
    ISSN    = {2076-3417},
    DOI     = {10.3390/app11125598}
}

@INPROCEEDINGS{Reference:Xu_RPFA-Net_a_4D_RaDAR_Pillar_Feature_Attention_Network_for_3D_Object_Detection,
    author  = {Xu, Baowei and Zhang, Xinyu and Wang, Li and Hu, Xiaomei and Li, Zhiwei and Pan, Shuyue and Li, Jun and Deng, Yongqiang},
    booktitle={2021 IEEE International Intelligent Transportation Systems Conference (ITSC)},
    title   = {RPFA-Net: A 4D RaDAR Pillar Feature Attention Network for 3D Object Detection}, 
    year    = {2021},
    volume  = {},
    number  = {},
    pages   = {3061-3066},
    doi     = {10.1109/ITSC48978.2021.9564754}
}

@INPROCEEDINGS{Reference:Lang_PointPillars_Fast_Encoders_for_Object_Detection_From_Point_Clouds,
    author  = {Lang, Alex H. and Vora, Sourabh and Caesar, Holger and Zhou, Lubing and Yang, Jiong and Beijbom, Oscar},
    booktitle={2019 IEEE/CVF Conference on Computer Vision and Pattern Recognition (CVPR)},
    title   = {PointPillars: Fast Encoders for Object Detection From Point Clouds}, 
    year    = {2019},
    volume  = {},
    number  = {},
    pages   = {12689-12697},
    doi     = {10.1109/CVPR.2019.01298}
}

@INPROCEEDINGS{Reference:Niederloehner_Self-Supervised_Velocity_Estimation_for_Automotive_Radar_Object_Detection_Networks,
    author  = {Niederlöhner, Daniel and Ulrich, Michael and Braun, Sascha and Köhler, Daniel and Faion, Florian and Gläser, Claudius and Treptow, André and Blume, Holger},
    booktitle={2022 IEEE Intelligent Vehicles Symposium (IV)},
    title   = {Self-Supervised Velocity Estimation for Automotive Radar Object Detection Networks},
    year    = {2022},
    volume  = {},
    number  = {},
    pages   = {352-359},
    doi     = {10.1109/IV51971.2022.9827295}
}

@INPROCEEDINGS{Reference:Koehler_Improved_Multi_Scale_Grid_Rendering_of_Point_Clouds_for_Radar_Object_Detection_Networks,
    author  = {Köhler, Daniel and Quach, Maurice and Ulrich, Michael and Meinl, Frank and Bischoff, Bastian and Blume, Holger},
    booktitle={2023 26th International Conference on Information Fusion (FUSION)},
    title   = {Improved Multi-Scale Grid Rendering of Point Clouds for Radar Object Detection Networks},
    year    = {2023},
    volume  = {},
    number  = {},
    pages   = {1-8},
    doi     = {10.23919/FUSION52260.2023.10224223}
}

@misc{Reference:Ian_Explaining_and_Harnessing_Adversarial_Examples,
    title   = {Explaining and Harnessing Adversarial Examples}, 
    author  = {Ian J. Goodfellow and Jonathon Shlens and Christian Szegedy},
    journal = {CoRR},
    year    ={2015},
    volume={abs/1412.6572},
    eprint  = {1412.6572},
    archivePrefix={arXiv},
    primaryClass={stat.ML},
    url={https://arxiv.org/abs/1412.6572}, 
}

@misc{Reference:Szegedy_Intriguing_properties_of_neural_networks,
    title={Intriguing properties of neural networks}, 
    author={Christian Szegedy and Wojciech Zaremba and Ilya Sutskever and Joan Bruna and Dumitru Erhan and Ian Goodfellow and Rob Fergus},
    year={2014},
    eprint={1312.6199},
    archivePrefix={arXiv},
    primaryClass={cs.CV},
    url={https://arxiv.org/abs/1312.6199}, 
}

@misc{Reference:Hendrycks_Benchmarking_Neural_Network_Robustness_to_Common_Corruptions_and_Perturbations,
    title={Benchmarking Neural Network Robustness to Common Corruptions and Perturbations}, 
    author={Dan Hendrycks and Thomas Dietterich},
    eprint={1903.12261},
    archivePrefix={arXiv},
    primaryClass={cs.LG},
    journal={ArXiv},
    year={2019},
    volume={abs/1903.12261},
    url={https://api.semanticscholar.org/CorpusID:56657912}
}

@misc{Reference:Shuangzhi_Common_Corruption_Robustness_of_Point_Cloud_Detectors_Benchmark_and_Enhancement,
    title={Common Corruption Robustness of Point Cloud Detectors: Benchmark and Enhancement}, 
    author={Shuangzhi Li and Zhijie Wang and Felix Juefei-Xu and Qing Guo and Xingyu Li and Lei Ma},
    year={2022},
    eprint={2210.05896},
    archivePrefix={arXiv},
    primaryClass={cs.CV},
    url={https://arxiv.org/abs/2210.05896}
}

@INPROCEEDINGS{Reference:Infuence_Rain,
    author  = {Hasirlioglu, Sinan and Kamann, Alexander and Doric, Igor and Brandmeier, Thomas},
    booktitle={2016 IEEE 19th International Conference on Intelligent Transportation Systems (ITSC)}, 
    title   = {Test methodology for rain influence on automotive surround sensors}, 
    year    = {2016},
    volume  = {},
    number  = {},
    pages   = {2242-2247},
    doi     = {10.1109/ITSC.2016.7795918}
}

@INPROCEEDINGS{Reference:Multipath,
    author  = {Kamann, Alexander and Held, Patrick and Perras, Florian and Zaumseil, Patrick and Brandmeier, Thomas and Schwarz, Ulrich T.},
    booktitle={2018 21st International Conference on Intelligent Transportation Systems (ITSC)}, 
    title   = {Automotive Radar Multipath Propagation in Uncertain Environments}, 
    year    = {2018},
    volume  = {},
    number  = {},
    pages   = {859-864},
    doi     = {10.1109/ITSC.2018.8570016}
}

@INPROCEEDINGS{Reference:Increasing_the_Robustness_of_Semantic_Segmentation_Models_with_Painting_by_Numbers,
    author  = {Christoph Kamann and Carsten Rother},
    booktitle={Computer Vision ECCV 2020}, 
    title   = {Increasing the Robustness of Semantic Segmentation Models with Painting-by-Numbers},
    publisher={Springer International Publishing},  
    year    = {2020},
    volume  = {},
    number  = {},
    pages   = {369-387},
    doi     = {10.1007/978-3-030-58607-2_22}
}

@INPROCEEDINGS{Reference:Chamseddine_Ghost_Target_Detection_in_3D_Radar_Data_using_Point_Cloud_based_Deep_Neural_Network,
    author  = {Chamseddine, Mahdi and Rambach, Jason and Stricker, Didier and Wasenmuller, Oliver},
    booktitle={2020 25th International Conference on Pattern Recognition (ICPR)}, 
    title   = {Ghost Target Detection in 3D Radar Data using Point Cloud based Deep Neural Network},
    year    = {2021},
    volume  = {},
    number  = {},
    pages   = {10398-10403},
    doi     = {10.1109/ICPR48806.2021.9413247}
}

@article{EOT_Granststroem_17,
    author = {Granström, Karl and Baum, Marcus and Reuter, Stephan},
    year = {2017},
    month = {12},
    pages = {},
    title = {Extended Object Tracking: Introduction, Overview, and Applications},
    volume = {12},
    journal = {Journal of Advances in Information Fusion}
}

@INPROCEEDINGS{Reference:Buhren_SimulationofAutomotiveRadar,
    author  = {Buhren, M. and Bin Yang},
    booktitle={2006 IEEE Intelligent Vehicles Symposium}, 
    title   = {Simulation of Automotive Radar Target Lists using a Novel Approach of Object Representation},
    year    = {2006},
    volume  = {},
    number  = {},
    pages   = {314-319},
    doi     = {10.1109/IVS.2006.1689647}
}

@ARTICLE{Reference:Hammerstrand_AdaptiveRadarSensorModel,
    author  = {Hammarstrand, Lars and Lundgren, Malin and Svensson, Lennart},
    journal = {IEEE Transactions on Aerospace and Electronic Systems}, 
    title   = {Adaptive Radar Sensor Model for Tracking Structured Extended Objects}, 
    year    = {2012},
    volume  = {48},
    number  = {3},
    pages   = {1975-1995},
    doi     = {10.1109/TAES.2012.6237574}
}

@article{Reference:Hammerstrand_Extended_Object_Tracking_using_a_Radar_Resolution_Model,
    author  = {Hammarstrand, Lars and Svensson, Lennart and Sandblom, Fredrik and Sorstedt, Joakim},
    year    = {2012},
    month   = {07},
    pages   = {2371-2386},
    title   = {Extended Object Tracking using a Radar Resolution Model},
    volume  = {48},
    journal = {IEEE Transactions on Aerospace and Electronic Systems - IEEE TRANS AEROSP ELECTRON SY},
    doi     = {10.1109/TAES.2012.6237597}
}

@INPROCEEDINGS{Reference:Gunnarsson_Tracking_vehicles_using_radar_detections,
    author  = {Gunnarsson, J. and Svensson, L. and Danielsson, L. and Bengtsson, F.},
    booktitle={2007 IEEE Intelligent Vehicles Symposium}, 
    title   = {Tracking vehicles using radar detections}, 
    year    = {2007},
    volume  = {},
    number  = {},
    pages   = {296-302},
    doi     = {10.1109/IVS.2007.4290130}
}

@article{Reference:Gilholm_Poisson_models_for_extended_target_and_group_tracking,
    author = {Gilholm, Kevin and Godsill, S.J. and Maskell, Simon and Salmond, David},
    year = {2005},
    month = {08},
    pages = {},
    title = {Poisson models for extended target and group tracking},
    volume = {5913},
    journal = {Proceedings of SPIE - The International Society for Optical Engineering},
    doi = {10.1117/12.618730}
}

@INPROCEEDINGS{HTG-XIA,
    author  = {Xia, Y. and Wang, P. and Berntorp, K. and Koike-Akino, T. and Mansour, H. and Pajovic, M. and Boufounos, P. and Orlik, P. V.},
    booktitle={ICASSP 2020 - 2020 IEEE International Conference on Acoustics, Speech and Signal Processing (ICASSP)}, 
    title   = {Extended Object Tracking Using Hierarchical Truncation Measurement Model with Automotive Radar}, 
    year    = {2020},
    volume  = {},
    number  = {},
    pages   = {4900-4904},
    doi     = {10.1109/ICASSP40776.2020.9054614}
}

@INPROCEEDINGS{Reference:Knill_A_direct_scattering_model_for_tracking_vehicles_with_high-resolution_radars,
    author  = {Knill, Christina and Scheel, Alexander and Dietmayer, Klaus},
    booktitle={2016 IEEE Intelligent Vehicles Symposium (IV)}, 
    title   = {A direct scattering model for tracking vehicles with high-resolution radars},
    year    = {2016},
    volume  = {},
    number  = {},
    pages   = {298-303},
    doi     = {10.1109/IVS.2016.7535401}
}

@article{Reference:Simon_Complex-YOLO_Real_time_3D_Object_Detection_on_Point_Clouds,
    author  = {Martin Simon and Stefan Milz and Karl Amende and Horst{-}Michael Gross},
    title   = {Complex-YOLO: Real-time 3D Object Detection on Point Clouds},
    journal = {CoRR},
    volume  = {abs/1803.06199},
    year    = {2018},
    url     = {http://arxiv.org/abs/1803.06199},
    eprinttype= {arXiv},
    eprint  = {1803.06199},
    bibsource= {dblp computer science bibliography, https://dblp.org}
}

@book{costa1984signal,
    title   = {Signal Processing and Range Spreading in the FM-CW Radar},
    author  = {Costa, E.A. and Chadwick, R.B. and Environmental Research Laboratories (U.S.)},
    series  = {NOAA technical report ERL.: WPL},
    url     ={https://books.google.de/books?id=iR4ZuwEACAAJ},
    year={1984},
    publisher={U.S. Department of Commerce, National Oceanic and Atmospheric Administration, Environmental Research Laboratories}
}

@ARTICLE{Reference:Xu_FMCW_Chirps,
    author  = {Xu, Lifan and Sun, Shunqiao and Mishra, Kumar Vijay and Zhang, Yimin D.},
    journal = {IEEE Transactions on Aerospace and Electronic Systems}, 
    title   = {Automotive FMCW Radar With Difference Co-Chirps}, 
    year    = {2023},
    volume  = {59},
    number  = {6},
    pages   = {8145-8165},
    doi     = {10.1109/TAES.2023.3299259}
}

@article{Stove1992LinearFR,
    title   = {Linear FMCW radar techniques},
    author    = {Andrew Stove},
    journal   = {Proceedings of the IEEE},
    year      = {1992},
    url       = {https://api.semanticscholar.org/CorpusID:57959585}
}

@phdthesis{Diss_Heidenreich,
    author = {P. Heidenreich},
    title = {Antenna array processing: Autocalibration and fast high-resolution methods for automotive radar},
    school = {Technische Univer.Darmstadt},
    year = {2012},
}

@ARTICLE{Reference:Burg,
    author  = {Gupta, I.J. and Beals, M.J. and Moghaddar, A.},
    journal = {IEEE Transactions on Antennas and Propagation}, 
    title   = {Data extrapolation for high resolution radar imaging}, 
    year    = {1994},
    volume  = {42},
    number  = {11},
    pages   = {1540-1545},
    doi     = {10.1109/8.362783}
}

@ARTICLE{Reference:Burg2,
    author  = {de Waele, S. and Broersen, P.M.T.},
    journal = {IEEE Transactions on Signal Processing}, 
    title   = {The Burg algorithm for segments}, 
    year    = {2000},
    volume  = {48},
    number  = {10},
    pages   = {2876-2880},
    doi     = {10.1109/78.869039}
}

@book{signal_Richards_05,
    title   = {Fundamentals of Radar Signal Processing},
    author  = {Richards, M.A.},
    isbn    = {9780071444743},
    lccn    = {2005047891},
    series  = {Professional Engineering},
    url     = {https://books.google.de/books?id=VOvsJ7G1oDEC},
    year    = {2005},
    publisher={Mcgraw-hill}
}

@INPROCEEDINGS{Reference:Zaumsiel_Radar_Signature_of_a_Micro_Doppler_generating-Soft_Target_for_Automotive_Pre_Crash_Systems,
    author  = {Zaumseil, Patrick and Engert, Rainer and Zdetski, Dennis and Steinhauser, Dagmar and Jumar, Ulrich and Brandmeier, Thomas},
    booktitle={2023 20th European Radar Conference (EuRAD)},
    title   = {Radar Signature of a Micro-Doppler generating Soft-Target for Automotive Pre-Crash Systems},
    year    = {2023},
    volume  = {},
    number  = {},
    pages   = {209-212},
    doi     = {10.23919/EuRAD58043.2023.10289490}
}

@article{Ransac_Fischler,
    author  = {Fischler, Martin A. and Bolles, Robert C.},
    title   = {Random Sample Consensus: A Paradigm for Model Fitting with Applications to Image Analysis and Automated Cartography},
    year    = {1981},
    issue_date= {June 1981},
    publisher= {Association for Computing Machinery},
    address = {New York, NY, USA},
    volume  = {24},
    number  = {6},
    issn    = {0001-0782},
    url     = {https://doi.org/10.1145/358669.358692},
    doi     = {10.1145/358669.358692},
    journal = {Commun. ACM},
    month   = {jun},
    pages   = {381–395},
    numpages= {15}
}

@INPROCEEDINGS{Reference:EOT,
    author  = {Hanumegowda, Anusha and Dewangan, Soumya and Bhupala, Srihari and Gruson, Frank and Steinhauser, Dagmar},
    booktitle={2021 18th European Radar Conference (EuRAD)}, 
    title   = {Extended Object Tracking with IMM Filter for Automotive Pre-Crash Safety Applications}, 
    year    = {2022},
    volume  = {},
    number  = {},
    pages   = {177-180},
    doi     = {10.23919/EuRAD50154.2022.9784586}
}

@INPROCEEDINGS{HTG-XIA-WANG,
    author  = {Xia, Yuxuan and Wang, Pu and Berntorp, Karl and Mansour, Hassan and Boufounos, Petros and Orlik, Philip V.},
    booktitle={2020 IEEE 11th Sensor Array and Multichannel Signal Processing Workshop (SAM)}, 
    title   = {Extended Object Tracking Using Hierarchical Truncation Model with Partial-View Measurements}, 
    year    = {2020},
    volume  = {},
    number  = {},
    pages   = {1-5},
    doi     = {10.1109/SAM48682.2020.9104388}
}

@ARTICLE{HTG-XIA-Journal,
    author  = {Xia, Yuxuan and Wang, Pu and Berntorp, Karl and Svensson, Lennart and Granström, Karl and Mansour, Hassan and Boufounos, Petros and Orlik, Philip V.},
    journal = {IEEE Journal of Selected Topics in Signal Processing}, 
    title   = {Learning-Based Extended Object Tracking Using Hierarchical Truncation Measurement Model With Automotive Radar}, 
    year    = {2021},
    volume  = {15},
    number  = {4},
    pages   = {1013-1029},
    doi     = {10.1109/JSTSP.2021.3058062}
}

@article{Ellipse-Li,
    title   = {Ellipse fitting based approach for extended object tracking},
    author  = {Li, Borui and Mu, Chundi and Bai, Yongqiang and Bi, Jianquan and Wang, Lei and others},
    journal = {Mathematical Problems in Engineering},
    volume  = {2014},
    year    = {2014},
    publisher={Hindawi}
}

@article{Reference:Alexey_Bochkovskiy_YOLOv4_Optimal_Speed_and_Accuracy_of_Object_Detection,
    author  = {Alexey Bochkovskiy and Chien-Yao Wang and Hong-Yuan Mark Liao},
    title   = {YOLOv4: Optimal Speed and Accuracy of Object Detection},
    year    = {2020},
    journal = {arXiv},
}

@article{Reference:Chen_Multi_View_3D_Object_Detection_Network_for_Autonomous_Driving,
    author  = {Xiaozhi Chen and Huimin Ma and Ji Wan and Bo Li and Tian Xia},
    title   = {Multi-View 3D Object Detection Network for Autonomous Driving},
    journal = {CoRR},
    volume  = {abs/1611.07759},
    year    = {2016},
    url     = {http://arxiv.org/abs/1611.07759},
    eprinttype= {arXiv},
    eprint  = {1611.07759},
    bibsource= {dblp computer science bibliography, https://dblp.org}
}

@book{Computer-Ha,
    title   = {Computer and Robot Vision},
    author  = {Haralick, R. M. and Shapiro, L. G.},
    year    = {1993},
    number  = {Bd. 2},
    publisher={Addison-Wesley},
    isbn    = {9780201569438},
}

@book{Morphological-So,
    author  = {Soille, Pierre},
    title   = {Morphological Image Analysis},
    year    = {2002},
    publisher= {Springer},
    address = {Berlin},
    isbn    = {978-3-540-42988-3},
    doi     = {https://doi.org/10.1007/978-3-662-05088-0},
    edition = {2nd},
    series  = {Computational Imaging and Vision},
    volume  = {18},
    pages   = {389},
}

@Inbook{cam_img_Gao2021,
    author  = "Gao, Xiang and Zhang, Tao",
    title   = "Cameras and Images",
    bookTitle="Introduction to Visual SLAM: From Theory to Practice",
    year    = "2021",
    publisher="Springer Singapore",
    address = "Singapore",
    pages   = "87--108",
    isbn    = "978-981-16-4939-4",
    doi     = "10.1007/978-981-16-4939-4_4",
    url     = "https://doi.org/10.1007/978-981-16-4939-4_4"
}

@INPROCEEDINGS{Reference:Schumann_RadarScenes_A_Real_World_Radar_Point_Cloud-Data_Set_for_Automotive_Applications,
    author  = {Schumann, Ole and Hahn, Markus and Scheiner, Nicolas and Weishaupt, Fabio and Tilly, Julius F. and Dickmann, Jürgen and Wöhler, Christian},
    booktitle={2021 IEEE 24th International Conference on Information Fusion (FUSION)}, 
    title   = {RadarScenes: A Real-World Radar Point Cloud Data Set for Automotive Applications}, 
    year    = {2021},
    volume  = {},
    number  = {},
    pages   = {1-8},
    doi     = {10.23919/FUSION49465.2021.9627037}
}
\end{document}